\documentclass{article}
\usepackage{palatino}
\usepackage[letterpaper,margin=1in]{geometry}
\usepackage{hyperref}

\usepackage{algorithm}
\usepackage{algorithmic}
\usepackage{makecell}
\usepackage{times}
\usepackage{graphicx}
\usepackage[rflt]{floatflt}
\usepackage{epsfig}
\usepackage{amsmath,amssymb,theorem,float,bbm,bm,enumerate,multirow}
\usepackage{rotating}
\usepackage{array}
\usepackage{epstopdf}
\usepackage{times}
\usepackage{color}
\usepackage{url}
\usepackage{chngcntr}
\usepackage{nameref}
\usepackage{zref-xr}
\zxrsetup{toltxlabel}
\usepackage{subcaption}
\DeclareCaptionFont{tiny}{\tiny}

\newcommand{\beq}{\begin{equation}}
\newcommand{\eeq}{\end{equation}}

\newcommand\I{\mathbb{I}}

\newcommand\R{\mathbb{R}}

\newcommand\cA{\mathcal{A}}

\newcommand{\suchthat}{\;\ifnum\currentgrouptype=16 \middle\fi|\;}

\newcommand{\g}{\mathbf{g}}

\newcommand{\cL}{{\cal L}}

\newcommand{\vertiii}[1]{{\left\vert\kern-0.25ex\left\vert\kern-0.25ex\left\vert #1
    \right\vert\kern-0.25ex\right\vert\kern-0.25ex\right\vert}}

\newcommand{\E}{\mathbf{E}}

\newcommand{\myref}[1]{(\ref{#1})}

\DeclareMathOperator{\argmax}{argmax}
\DeclareMathOperator{\argmin}{argmin}

\newcounter{exampleI}
{\theorembodyfont{\rmfamily} \theoremstyle{plain} }

\newcounter{exampleII}
{\theorembodyfont{\rmfamily} \theoremstyle{plain} }

\newcounter{exampleIII}
{\theorembodyfont{\rmfamily} \theoremstyle{plain} }

{\theorembodyfont{\rmfamily} }
{\theorembodyfont{\rmfamily} \newtheorem{defn}{Definition}}
\newtheorem{theo}{Theorem}
\newtheorem{prop}{Proposition}
\newtheorem{lemm}{Lemma}

\newcommand{\proof}{\noindent{\itshape Proof:}\hspace*{1em}}

\newcommand{\qed}{\nolinebreak[1]~~~\hspace*{\fill} \rule{5pt}{5pt}\vspace*{\parskip}\vspace*{1ex}}

\newcommand {\commentout}[1] {}

\def\ints{{{\rm Z} \kern -.35em {\rm Z} }}  
\def\smallints{{{\rm Z} \kern -.3em {\rm Z} }}  
\def\pints{{{\rm I} \kern -.15em {\rm N} }}      
\newcommand{\reals}{\mathbb R}

\def\cplx{{{\rm I} \kern -.45em {\rm C} }}       
\def\l2{\rm {\mathcal L}^{2}(\reals)}            

\newcommand{\abs}[1]{\left|#1\right|}

\newtheorem{nad}{Notation and Definitions}[section]

\newtheorem{corollary}{Corollary}

\newcommand{\be}{\begin{eqnarray}}
\newcommand{\ee}{\end{eqnarray}}
\newcommand{\bea}{\begin{eqnarray}}
\newcommand{\eea}{\end{eqnarray}}
\newcommand{\beaa}{\begin{eqnarray*}}
\newcommand{\eeaa}{\end{eqnarray*}}
\newcommand{\bnad}{\begin{nad}}
\newcommand{\enad}{\end{nad}}

\usepackage[suppress]{color-edits}
\addauthor{sw}{blue}
\addauthor{ab}{magenta}
\addauthor{vs}{red}
\usepackage{authblk}

\title{Structured Linear Contextual Bandits:\\
	A Sharp and Geometric Smoothed Analysis}
\author{Vidyashankar Sivakumar}	
\author{Zhiwei Steven Wu}
\author{Arindam Banerjee}
\affil{Department of Computer Science \& Engineering\\ 
University of Minnesota, Twin Cities\\ 
Minneapolis, MN, USA}
	
\begin{document}

    \date{}
	
	\maketitle

	\begin{abstract}
		Bandit learning algorithms typically involve the balance of exploration and exploitation. However, in many practical applications, worst-case scenarios needing systematic exploration are seldom encountered. In this work, we consider a smoothed setting for structured linear contextual bandits where the adversarial contexts are perturbed by Gaussian noise and the unknown parameter $\theta^*$ has structure, e.g., sparsity, group sparsity, low rank, etc. We propose simple greedy algorithms for both the single- and multi-parameter (i.e., different parameter for each context) settings and provide a unified regret analysis for $\theta^*$ with any assumed structure. The regret bounds are expressed in terms of geometric quantities such as Gaussian widths associated with the structure of $\theta^*$. We also obtain sharper regret bounds compared to earlier work for the unstructured $\theta^*$ setting as a consequence of our improved analysis. We show there is implicit exploration in the smoothed setting where a simple greedy algorithm works.
	\end{abstract}
	
	\section{Introduction}
	\label{sec:main_intro}
	Contextual bandits~\cite{Langford-nips07} is a powerful framework for sequential decision-making, with many applications to clinical trials, web search, and content optimization. In a typical scenario, users arrive over time, and the algorithm chooses among various content (e.g., news articles) to present to each user and observes the outcome (e.g.,~clicks). A popular parametric formulation for this problem is the linear contextual bandit setting~\cite{chu2011contextual,Langford-www10}: in rounds $t = 1,\hdots,T$, the algorithm selects a context $x_{i^t}^t$ 
from $k$ available contexts $x_1^t,\hdots,x_k^t$ and receives a noisy reward $r^t(x^t_{i^t}) = \langle x_{i^t}^t,\theta^* \rangle + \omega^t$ where $\theta^*$, $\omega^t$ are the unknown parameter and noise respectively. The goal of the algorithm is to select arms to maximize rewards over time observing only the \abedit{available} contexts and the reward associated with the selected context in each round.  Such algorithms typically need to balance \emph{exploration}, making potentially sub-optimal decisions for the sake of information acquisition, and \emph{exploitation}, selecting decisions that are optimal based on the estimate \abedit{of $\theta^*$}. In particular, 
greedy algorithm which \abcomment{reads a bit odd}  myopically selects contexts maximizing rewards based on the current parameter estimate $\hat{\theta}$, i.e., choosing $x_{i^t}^t = \underset{x_i^t:1 \leq i \leq k}{\argmax} \langle x_i^t,\hat{\theta} \rangle$ are known to be sub-optimal in the worst case (see \cite{competingBandits-itcs16} for an example). At the same time, the greedy algorithm offers several appealing features, including its simplicity in computation and its best-effort treatments \abcomment{I don't know what that means, maybe cit?} to every user \cite{bbcd16,bastani2017exploiting}.

Given the advantages of the greedy algorithm, there has been recent work that investigates when the greedy algorithms perform well. On the practical side, \cite{practicalCB-arxiv18} shows that there is strong empirical evidence that exploration free algorithms perform well on real data sets. On the theoretical side, a line of work \cite{bastani2017exploiting,kmrw18,externalities-colt18} analyzed conditions under which inherent diversity in the data makes explicit exploration unnecessary.  In particular, the work of \cite{kmrw18,externalities-colt18} provide a \emph{smoothed analysis} on the greedy algorithm under the following setting: in each round the contexts $x_i^t, 1 \leq i \leq k$ are of the form $\mu_i^t + g_i^t, 1 \leq i \leq k$, where the $\mu_i^t \in \R^p$'s are possibly selected adverserially with the constraint $\|\mu_i^t\|_2 \leq 1$ and $g_i^t \sim N(0,\sigma^2 \I_{p \times p})$ are random Gaussian perturbations independent of the $\mu_i^t$'s. The algorithm in each round selects a context $x_{i^t}^t$ and receives noisy reward $r^t = \langle x_{i^t}^t,\theta^*_{i^t} \rangle + \omega^t$ where the parameter $\theta^*_{i^t}$ is unknown and there can be a different parameter corresponding to each context. 

Our work substantially generalizes the smoothed analysis framework for linear contextual bandits considered in \cite{kmrw18,externalities-colt18}. We enrich and refine these prior analyses by explicitly capturing the structure in the unknown parameters, specifically low values according to some atomic norm $R(\cdot)$ (e.g., $\ell_1$ norm, group-sparse norms, nuclear norms, k-support norm, etc. \cite{jaov09,arfs12,yuli06,tibs96,care09}). We consider two variants of the problem: the multi parameter setting when there is a separate parameter corresponding to each context, i.e., $\theta^*_1,\hdots,\theta^*_k$ and the single parameter setting when there is a single unknown parameter, i.e., $\theta^* = \theta^*_1 = \theta^*_2 = \hdots = \theta^*_k$. In any round $t$ the greedy algorithm maintains estimates of the true parameters $\hat{\theta}_1^t,\hdots,\hat{\theta}_k^t$ using the constrained least squares estimator:
\beq
\hat{\theta}_i^t = \underset{\theta \in \R^p}{\argmin} \quad \cL(\theta;Z_i^t,y_i^t) \quad \text{s.t.} \quad R(\theta) \leq R(\theta^*_i) ~,
\label{eq:ch6_cons_ls_intro}
\eeq
where $\cL(\theta;Z_i^t,y_i^t)$ is the least squares loss, $Z_i^t$ is the design matrix in round $t$ whose rows are contexts chosen in the rounds prior to $t$ and $y_i^t$ is a vector with the corresponding rewards for context $i$.
The greedy algorithm then selects the arm corresponding to the highest reward w.r.t.~to the current parameter estimate, i.e., $x_{i^t}^t = \underset{x_i^t:1 \leq i \leq k}{\argmax} \langle x_i^t,\hat{\theta}_i^t \rangle$. We analyze the performance of the greedy algorithm w.r.t.~the regret which compares the performance with a clairvoyant learner having knowledge of the optimal parameter $\theta^*_i$,
\beq
\text{Reg}(T) = \sum_{t=1}^T \left ( \max_i \langle x_i^t,\theta^*_i \rangle - \langle x_{i^t}^t,\theta^*_{i^t} \rangle \right ) ~,
\label{eq:reg_defn}
\eeq
where in the single parameter setting $\theta^*_i = \theta^*_{i^t} = \theta^*$.

In our main results we derive worst case regret bounds for the single and multi parameter settings. Consider first the single parameter problem setting. In any round $t$, denote the error vector $\Delta^t = \hat{\theta}^t - \theta^*$. It is evident from equation \myref{eq:ch6_cons_ls_intro} that the error vector lies in the error set
$E_c = \{ \Delta ~ \suchthat ~ R(\theta^* + \Delta) \leq R(\theta^*) \}$.
Now consider the set $A = \text{cone}(E_c) \cap S^{p-1}$ \cite{birt09,nrwy12} and define by $w(A)$ the Gaussian width of set $A$ \cite{tala05,tala14,gord85}. The Gaussian width is a metric for the complexity/size of a set \cite{tala05,tala14,gord85} widely used in literature on analysis of high-dimensional statistical models \cite{bcfs14,chba15,crpw12,sibp15}. For example, Gaussian width of the error set for $R(\cdot) = \|\cdot\|_1$ and $s$-sparse $\theta^*$ is $\Theta(s \log p)$. \abdelete{Note that the notations $y = \Theta(x)$ (respectively $y = O(x)$, $y = \Omega(x)$) implies there exists absolute constants $c_1,c_2,c_3,c_4$ such that $c_1 \cdot x \leq y \leq c_2 \cdot x$ (respectively $y \leq c_3 \cdot x$, $y \geq c_4 \cdot x$) and $\tilde{\Theta}(\cdot)$, $\tilde{\Omega}(\cdot)$ and $\tilde{O}(\cdot)$ notations hide the dependence on logarithm terms and noise variance.} We show that the single parameter setting requires a warm start phase of $t_{\min} = \tilde{\Theta}(w^2(A))$ rounds when the contexts are chosen randomly or in a round robin fashion. 
After the first $t_{\min }$ rounds where the algorithm accrues linear regret, we obtain worst case regret bounds of the form: 
\beq
\text{Reg}(T) = \tilde{O} \left (\frac{w(A) \sqrt{T}}{\sigma} \right ) ~,
\eeq
where $\sigma^2$ is the variance of the Gaussian perturbations on the contexts.
We make the following observations comparing our results to prior work. 
\begin{enumerate}
\item For the unconstrained problem $w(A) = \Theta(\sqrt{p})$ and $\text{Reg}(T) = \tilde{O} \left ( \frac{\sqrt{pT}}{\sigma} \right )$. When $\sigma^2 = O \left ( \frac{1}{p} \right )$ as considered in \cite{kmrw18}, ignoring logarithmic factors, the regret bounds are sharper compared to the results in \cite{kmrw18} by a factor $\sqrt{p}$. Moreover when $\sigma^2 = O \left ( \frac{1}{p} \right )$, the regret upper bound is of the same order as the regret upper bounds obtained for UCB-style algorithms in \cite{dahk08,abps11} for stochastic linear bandits and better than the regret upper bounds for Thompson sampling \cite{aggo13}. With more smoothing when $\sigma^2 > \frac{1}{p}$ the greedy algorithm performs better giving lower regret whereas less smoothing has the reverse effect.
\item For $R(\cdot) = \|\cdot\|_1$ and $s$-sparse $\theta^*$, $w(A) = \Theta(\sqrt{s \log p})$ leading to the regret bounds, $\text{Reg}(T) = \tilde{O} \left ( \frac{\sqrt{ s \log p \cdot T}}{\sigma} \right )$. Again when $O \left ( \frac{1}{p} \right )$, the regret upper bounds are of the same order as \cite{abps12} where a UCB-style algorithm was proposed for the $\ell_1$ regularized stochastic linear bandits problem. Note that the algorithm proposed in \cite{abps12} is computationally involved and difficult to optimize. 
\item 
Our analysis can handle any atomic norm $R(\cdot)$ and captures the geometry of the problem obtaining results in terms of easily computable geometric quantities like the Gaussian width \cite{tala05,tala14,gord85,chba15}

\abcomment{please add a point -- our algorithm works for any structure, like group sparse, low rank, etc.}
\end{enumerate}
The multi parameter setting requires a warm start phase of $\tilde{\Theta} \left ( \frac{k w^2(A)}{\sigma^4} \right )$, where $k$ is the number of contexts. When $\sigma^2 = O \left (\frac{1}{p} \right)$, 
in the worst case, we require the length of the warm start phase to be $\tilde{\Theta}(k \cdot p^2 \cdot w^2(A) )$.
In the un-structured setting, $w^2(A) = p$ which translates to $\tilde{\Theta}(kp^3)$ rounds in the warm start phase which improves over the $\tilde{\Theta}(kp^{6})$ rounds in \cite{kmrw18} (see Theorem 4.2). The algorithm achieves $\tilde{O} \left ( \frac{w(A) \sqrt{T k}}{\sigma} \right )$ regret after the warm start rounds which is $\sqrt{k}$ times worse compared to the single parameter setting.

We briefly summarize the organization and notations used throughout the paper. We concisely present the main ideas and technical results in Section \ref{sec:main_overview} of the paper. 
Results for the single parameter and multi parameter settings are presented in Section \ref{sec:single_parameter} and \ref{sec:multi_parameter} respectively before concluding in Section \ref{sec:main_conc}. All proofs are pushed to the supplementary section. 

{\bf Notation.} Throughout the paper we use constants like $c,c_1,c_2,\hdots$ whose definition may change from one line to the next. In certain places we use the terms contexts and arms interchangeably. The notations $y = \Theta(x)$ (respectively $y = O(x)$, $y = \Omega(x)$) implies there exists absolute constants $c_1,c_2,c_3,c_4$ such that $c_1 \cdot x \leq y \leq c_2 \cdot x$ (respectively $y \leq c_3 \cdot x$, $y \geq c_4 \cdot x$) and $\tilde{\Theta}(\cdot)$, $\tilde{\Omega}(\cdot)$ and $\tilde{O}(\cdot)$ notations hide the dependence on logarithm terms and noise variance.
	
	\section{Overview of Main Technical Results}
	\label{sec:main_overview}
	 We summarize the major ideas and results in this paper.
 
 {\bf Episodic algorithm.} The algorithm we analyze has an episodic theme \cite{jana18} due to its computational efficiency and simplicity. Let $T$ denote the total number of rounds. In the single parameter setting, denote the episode number by $e$ and let $T_e$ denote the total number of rounds in episode $e$. The number of rounds in each episode increases geometrically with time, i.e., $T_1 = 2T_0$, $T_2 = 2T_1$ and so on. The total number of rounds $T = \sum_{e} T_e$. The number of episodes scales as $\log T$. The regression parameter is estimated at the beginning of episode $e+1$ using only the contexts and rewards observed in the $T_{e}$ rounds in the immediately preceding episode using the following constrained least squares estimator:
\beq
    \hat{\theta}^{(e+1)} = \underset{\theta \in \R^p}{\argmin} \frac{1}{2T_e} \|y^{(e)} - Z^{(e)}\theta\|_2^2 \quad \text{s.t.} \quad R(\theta) \leq R(\theta^*) ~,
    \label{eq:ch6_cons_ls_overview}
\eeq
where $Z^{(e)} \in \R^{T_e \times p}$ is the design matrix constructed with rows as contexts observed in episode $e$ and $y^{(e)} \in \R^{T_e}$ the corresponding observed rewards. 
In the multi parameter setting, the only difference to the single parameter setting is that we maintain separate design matrices, rewards, parameter estimates and episodes for each context.

{\bf Estimation error.} The regret in both the single and multi parameter settings depends on the estimation error for the parameter estimated using the constrained least squares estimator at the beginning of each episode. Consider parameter estimation in episode $e+1$. Let $Z^{(e)} \in \R^{T_e \times p}$ be the design matrix constructed with rows as contexts observed in episode $e$ and $y^{(e)} \in \R^{T_e}$ the corresponding observed rewards. We precondition the data before parameter estimation using the Puffer transformation \cite{jiro15}. The Puffer transformation computes the SVD of the design matrix as $\frac{1}{\sqrt{T_e}} Z^{(e)} = U^{(e)} D^{(e)} (V^{(e)})^{\intercal}$ followed by transforming the data as $\tilde{Z}^{(e)} = F^{(e)} Z^{(e)},~ \tilde{y}^{(e)} = F^{(e)} y^{(e)}$ where $F^{(e)} = U^{(e)} (D^{(e)})^{-1} (U^{(e)})^{\intercal}$. The parameter at the beginning of episode $e+1$ is then estimated using the following least squares constrained estimator:
 \beq
     \hat{\theta}^{(e+1)} = \underset{\theta \in \R^p}{\argmin} \frac{1}{T_e} \|\tilde{y}^{(e)} - \tilde{Z}^{(e)} \theta \|_2^2 \quad s.t. \quad R(\theta) \leq R(\theta^*) ~.
     \label{eq:intro_ls_cons}
 \eeq
We derive upper bounds on the parameter estimation error using the Puffer transformed data. In the worst case Puffer transformed data gives better estimation bounds compared to the bounds obtained using raw data \cite{crpw12,nrwy12,bcfs14}. Our analysis borrows tools and techniques from the existing vast literature on high-dimensional estimation \cite{wain19,vers18}. Specifically, following the analysis framework in \cite{bcfs14}, we need three main results. First, note that to satisfy the constraint in \myref{eq:intro_ls_cons} the error vector $\Delta$ with $\hat{\theta}^{(e+1)} = \theta^* + \Delta$ lies in the following set,
 \beq
     E_c = \{ \Delta ~ \suchthat ~ R(\theta^* + \Delta) \leq R(\theta^*) \} ~.
 \eeq 
 Second, for consistent estimation we show the design matrix satisfies the following restricted eigenvalue (RE) condition on the error set $A = \text{cone}(E_c) \cap S^{p-1}$ \cite{birt09,nrwy12} with high probability across all episodes once $T > t_{\min} = \tilde{\Theta}(w^2(A))$,
 \beq
     \inf\limits_{u \in A} \frac{1}{T_e} \|\tilde{Z}^{(e)} u \|_2^2 = \tilde{\Omega}(\sigma^2) ~.
 \eeq
 Existing results on the RE condition \cite{mept07,bcfs14,nrwy12} with i.i.d. rows cannot be directly applied since the rows in the design matrix depend on previously selected contexts and rewards. We make use of recent novel results in \cite{bgsw19} on bounds for sum of random quadratic quantities with dependence. 
 Third, for rounds $T > t_{\min}$ we obtain high probability upper bounds on the estimation error with the Puffer transformed data across all episodes.
 \beq
     \max_e \|\hat{\theta}^{(e+1)} - \theta^*\|_2 \leq \tilde{O} \left (  \frac{(w(A)}{\sigma \sqrt{T_{e}}} \right ) ~.
     \label{eq:intro_sp_est_err_bnds}
 \eeq
 The non-asymptotic bounds on the estimation error are novel, both due to dependence of data observed in each round to contexts and rewards observed in previous rounds as also the use of the Puffer transformation for which no results exist for estimation error to the best of our knowledge. The results on parameter estimation errors also holds in the multi parameter setting except we maintain separate parameter estimates for each context.
 
 {\bf Regret.} For both the single and multi parameter settings we show the regret depends on the $\ell_2$ norm of the estimation error for the parameter estimated at the beginning of each episode after an initial warm start phase when the algorithm accrues linear regret. In the single parameter setting the length of the warm start phase is $t_{\min} = \tilde{\Theta} \left (w^2(A) \right )$ rounds while in the multi parameter setting it is $t_{\min} =  \tilde{\Theta} \left ( \frac{  k w^2(A)}{\sigma^4} \right )$ rounds. The dependence of $t_{\min}$ on $\sigma$ for the multi parameter setting implies a large warm start phase when $\sigma$ is small. For example, if $\sigma^2 = O \left ( \frac{1}{p} \right )$ as assumed in \cite{kmrw18}, then $t_{\min}$ scales as $p^2$ which maybe prohibitive in many high-dimensional applications. After the warm start phase we show the regret in the single parameter setting is upper bounded as follows:
  \beq
     \text{Reg}(T) = \tilde{O} \left ( \frac{ w(A) \sqrt{T}}{\sigma}  \right ) ~,
\eeq
The upper bound on the regret in the multi parameter setting after the warm start phase is worse compared to the single parameter setting by a factor of $\sqrt{k}$:
 \beq
      \text{Reg}(T) =  \tilde{O} \left ( \frac{ w(A) \sqrt{kT}}{\sigma}  \right ) ~.
 \eeq

	\section{Single Parameter Regret Analysis}
	\label{sec:single_parameter}
	We present results for the single parameter setting in this section. 
The greedy algorithm proceeds in multiple episodes with the length of each episode increasing geometrically with time \cite{jana18}. We index episode numbers by $e$, time steps by $t$ and arms by $i$. We denote by $T$ the total number of rounds and by $T_e$ the number of rounds in episode $e$. 
In each round, the algorithm observes contexts $x_i^t,1 \leq i \leq k$ and greedily selects the optimal arm based on the current parameter estimate, i.e., $z^t = \underset{x_i^t:1 \leq i \leq k}{\argmax} \langle x_i^t,\hat{\theta}^{(e)} \rangle$ and receives noisy reward $y^t = \langle z^t,\theta^* \rangle + \omega^t$ with $\omega^t$ denoting the noise at time $t$. 
The parameter is estimated at the beginning of each episode using the contexts and rewards observed in the previous episode 
using the constrained least squares estimator with the Puffer transformed design matrix and response \cite{jiro15}. Note that the design matrix is rank deficient in the first $e = \lceil \log t_{\min} \rceil$ rounds with $t_{\min} = \tilde{\Theta}(w^2(A))$ when the contexts will be chosen uniformly at random. 

\begin{algorithm}[t]
	\begin{algorithmic}[1]
		\STATE Initialize empty design matrix and reward vector $Z^{(0)} = [],y^{(0)} = []$
		\FOR{$e = 1,2,3,\hdots,\lfloor \log_2 T \rfloor$}
		\STATE Compute SVD as $\frac{1}{\sqrt{T_{e-1}}} Z^{(e-1)} = U^{(e-1)} D^{(e-1)} (V^{(e-1)})^{\intercal}$
		\STATE Compute the Puffer transformation $F^{(e-1)} = U^{(e-1)} (D^{(e-1)})^{-1} (U^{(e-1)})^{\intercal}$ and define $\tilde{Z}^{(e-1)} = F^{(e-1)}Z^{(e-1)}$ and $\tilde{y}^{(e-1)} = F^{(e-1)} y^{(e-1)}$
		\STATE Estimate parameter using constrained least squares estimator breaking ties arbitrarily when necessary
		\begin{align}
		\hat{\theta}^{(e)} &= \underset{\theta \in \R^p}{\argmin} \frac{1}{2T_{e-1}} \|\tilde{y}^{(e-1)} - \tilde{Z}^{(e-1)} \theta\|_2^2 \nonumber  \\ 
		&\quad \quad \quad \quad \quad \quad \quad \text{s.t.} \quad R(\theta) \leq R(\theta^*) ~,
		\label{eq:single_param_est}
		\end{align}
		where $T_{e-1}$ is the number of observations in the previous episode.
		\STATE Initialize empty design matrix and reward vector $Z^{(e)} = [], y^{(e)} = []$. Set $T_e = 2^{e-1}$ 
		\FOR{$t = 2^{(e-1)} + 1$ to $2^{e}$}
		\STATE Observe contexts $x_1^t,\hdots,x_k^t \in \R^p$
		\STATE Choose arm $z^t = \underset{x_i^t:1 \leq i \leq k}{\argmax} \langle x_i^t,\hat{\theta}^{(e)} \rangle$ and observe reward $y^t = \langle z^t,\theta^* \rangle + \omega^t$ where $\omega^t$ is zero mean $\kappa_{\omega}$-sub-Gaussian noise
		\STATE Append observations $(z^t,y^t)$ to $(Z^{(e)},y^{(e)})$
		\ENDFOR
		\ENDFOR
	\end{algorithmic}
	\caption{Structured Greedy (single parameter)}
	\label{algo:greedy_single_param}
\end{algorithm}

Lemma \ref{lemm:single_param_regret} gives an upper bound for the regret for Algorithm \ref{algo:greedy_single_param}. The greedy algorithm accrues linear regret in the first $t_{\min}$ rounds when the design matrix is rank deficient for parameter estimation, i.e., it does not satisfy the restricted eigenvalue condition. Subsequent rounds are played in an episodic fashion with the regret in any round depending on the accuracy of parameter estimation at the beginning of the episode.

\begin{lemm}{\bf (Single Parameter Regret Bounds)}
	Denote by $\beta = \underset{\substack{{1 \leq i \leq k, 1 \leq t \leq T} \\  {v \in A}}}{\max} \langle x_i^t,v \rangle$, where $A = \text{cone}(E_c) \cap S^{p-1}$ is the error set..
	Assume $T > t_{\min}$, where $t_{\min}$ depends on properties of the true parameter $\theta^*$ and the regularizer $R(\cdot)$. Then,
	\beq 
	\text{Reg}(T) \leq 4 \beta t_{\min} + \sum_{e = \lceil \log t_{\min} \rceil}^{\lfloor \log T \rfloor}  2\beta T_e \|\hat{\theta}^{(e)} - \theta^*\|_2 ~.
	\eeq
	\label{lemm:single_param_regret}	
\end{lemm}

\subsection{Gaussian Contexts}
\label{sec:single_parameter_gaussian}
In order to build intuition, we establish results on performance of the greedy algorithm when the contexts
\abcomment{contexts? maybe say upfront these will be used exchangeably?} 
are completely stochastic, i.e., we derive regret bounds when the contexts are sampled independently from a Gaussian distribution, , $x_i^t \sim N(0,\sigma^2 \I_{p \times p}), 1 \leq i \leq k, t \leq T$ in step 9 of Algorithm \ref{algo:greedy_single_param}. The episodic algorithm ensures independence between data in each round of an episode.
Additionally, the rows of the design matrix are sub-Gaussian and the covariance matrix satisfies the minimum eigenvalue condition. 
\begin{lemm}{\bf (Single Parameter Gaussian Arms Design Matrix Properties)}
		The rows of the design matrix $Z^{(e)} \in \R^{T_e \times p}$ in any episode $e$ satisfy $\kappa_z = \|z^t\|_{\psi_2} \leq c_2 \sigma \sqrt{\log k}$ for $c_2$ some positive constant.  Moreover the minimum eigenvalue of the matrix $E_{z^t}[z^t (z^T)^T]$ satisfies,
	\beq
	\lambda_{\min} (E_{z^t}[z^t (z^t)^\intercal]) \geq c_1 \frac{\sigma^2}{\log k} ~,
	\eeq
	where $c_1$ is some positive constant and the expectation is over the chosen contexts.
	\label{lemm:single_param_design_matrix}
\end{lemm}
The result of Lemma \ref{lemm:single_param_design_matrix} and independence of data in any round to data from another round in any particular episode allows us to use existing results on RE condition and estimation error bounds for design matrices with i.i.d. sub-Gaussian rows. The only deviation from traditional estimation is the use of the Puffer transformation. The Puffer transformation is a preconditioning technique analyzed in \cite{jiro15} and was practically found to have better performance when estimating the sparsity pattern with the Lasso estimator when the design matrix had heavily correlated rows. We obtain the following worst case upper bound on the $\ell_2$ norm of the estimation error with high probability with the Puffer transformed data:
\beq
	\|\hat{\theta}^{(e+1)} - \theta^* \|_2 \leq \tilde{O} \left ( \frac{w(A) }{\sigma \sqrt{T_e}} \right ) ~,
\eeq
where $A$ is the error set. We provide the proof in the appendix which essentially uses the same analysis tools and techniques from \cite{bcfs14}. The regret bounds now follow from a straightforward application of the result of Lemma \ref{lemm:single_param_regret}. When $\sigma = O \left ( \frac{1}{\sqrt{p}} \right )$, as assumed in \cite{kmrw18}, the regret bound is $\tilde{O}(w(A) \sqrt{pT})$.

\begin{theo}{\bf (Gaussian Arms Regret Bounds)}
	Consider Gaussian contexts. Then with probability atleast $1 - \delta$ 
	\beq
	\beta = \underset{\substack{{1 \leq i \leq k, 1 \leq t \leq T} \\  {v \in A}}}{\max} \langle x_i^t,v \rangle \leq  c_1 \sigma (w(A) + \sqrt{\log(1/\delta)}) ~.
	\eeq
 Also with $T \gg t_{\min} \geq c_1 (w(A) + \sqrt{ \log \log T} + \sqrt{\log (1/\delta)})^2\log^2 k$ 
 with probability atleast $1 - 4\delta$
	the following is an upper bound on the regret for the Greedy algorithm,
	\beq
	\text{Reg}(T) \leq O \left (\frac{\gamma \cdot \beta \cdot \log (T) \cdot \sqrt{T}}{\sigma} \right )
	\eeq
	where $\gamma = c\kappa_{\omega} \sqrt{\log k}(w(A) + \sqrt{\log \log T} + \sqrt{\log (1/\delta)})$	
	\label{thm:single_parameter_gaussian_regret_bounds}
\end{theo}


\abcomment{we can make the result more general: instead of the $w(A)$ in the bound, use $(w(A)+\gamma)$, then the probability will be $\exp(-w^2(A) - \gamma^2)$; for example, with $\gamma = \sqrt{\log T}$, the probability will have a $1/T$ term. Also, lets see why we have a $\min(w^2(A),\tau^2)$ term, why $\tau$ does not show up anywhere else, and whether $\gamma$ can also `fix' the second term.}





\subsection{Smoothed Adverserial Contexts}
\label{sec:single_parameter_adverserial}
We now focus on regret bounds when the contexts are $x_i^t = \mu_i^t + g_i^t, 1 \leq i \leq k, \forall 1 \leq t \leq T$.  Remember that an adversary can choose $\mu_i^t, \|\mu_i^t\|_2 = 1, \forall 1 \leq i \leq k$ based on the observed contexts and rewards in the previous rounds.
The primary question is if an adversary can negatively influence the design matrix to affect estimation error, or in other words lower the minimum eigenvalue compared to the completely stochastic setting. 
The answer is in the result of Lemma \ref{lemm:single_param_obl_design_matrix}, where we show that even in the adverserial setting the minimum eigenvalue of the covariance matrix of each row of the design matrix is no worse than the completely stochastic Gaussian setting. In particular, adding small random perturbations to adverserially selected contexts leads to implicit exploration where 
the greedy algorithm works well. 
\begin{lemm}{\bf (Design matrix properties for 
		smoothed adversary)}
	The rows of the design matrix $Z^{(e)} \in \R^{T_e \times p}$ in any episode $e$ are $z^t = \mu^t + g^t$ where $\mu^t,g^t = \underset{\mu_i^t,g_i^t:1 \leq i \leq k}{\argmax} \langle \mu_i^t + g_i^t,\hat{\theta}^{(e-1)} \rangle$, $g_i^t \sim N(0,\sigma^2 \I_{p \times p})$ with the sub-Gaussian norm of $g^t$ satisfying $\|g^t\|_{\psi_2} \leq c_2 \sigma \sqrt{\log k}$ for some constant $c_2$. Moreover we have the following lower bound on the expected minimum eigenvalue for any $\mu_i^t$'s:
	\beq
	\lambda_{\min} (E_{z^t}[z^t (z^t)^\intercal]) \geq c_1 \frac{\sigma^2}{\log k} ~,
	\eeq
	where $c_1$ is some constant.
	\label{lemm:single_param_obl_design_matrix}
\end{lemm}

Due to an adaptive adversary, the selected contexts and noise are no longer independent but 
depend on previously observed contexts and rewards. The dependency introduces additional complexity for analysis of the non-asymptotic estimation error.
To obtain results on the RE condition, we make use of recent novel results from \cite{bgsw19} on lower bounds for sum of quadratics of random variables with dependence. Upper bounds on the noise-design interaction term $\sup\limits_{u \in A} \langle (\tilde{Z}^{(e)})^{\intercal}\tilde{\omega}^{(e)},u \rangle$, where $\tilde{\omega}^{(e)} = F\omega^{(e)}$ is the effective noise due to the Puffer transformation and $A$ is the error set as defined earlier, are also required and obtained using arguments from generic chaining \cite{tala05,tala14}. The analysis leads to an upper bound on the estimation error which is the same if the contexts were completely stochastic Gaussian without any adversary.
\beq
	\|\hat{\theta}^{(e+1)} - \theta^* \|_2 \leq \tilde{O} \left ( \frac{w(A) }{\sigma \sqrt{T_e}} \right ) ~.
\eeq


\abcomment{Should we talk about the constraints on $\mu_i^t$, e.g., when is it selected, constraints on $L_2$ norm, etc.}

\vscomment{Highlighted them in the Introduction. Do you think it will be good to reiterate the greedy algorithm again here? }


High probability regret bounds can now be obtained from the result of Lemma \ref{lemm:single_param_regret}.


\begin{theo}{\bf (Smoothed Adversary Regret Bounds)}
	In the smoothed adversary setting with probability atleast $1 - \delta$ 
	\beq
	\beta = \underset{\substack{{1 \leq i \leq k, 1 \leq t \leq T} \\  {v \in A}}}{\max} \langle x_i^t,v \rangle \leq  (1+ c_1 \sigma (w(A) + \sqrt{\log(1/\delta)})) ~.
	\eeq
     Also with $T \gg t_{\min} \geq c_1( w(A)  + \sqrt{\log \log T} + \sqrt{\log (1/\delta)})^2\log^2 k$ 
     with probability atleast $1 - 4\delta$ the following is an upper bound on the regret,
	\beq
	\text{Reg(T)} \leq O \left ( \frac{ \gamma \cdot \beta \cdot \log (T) \cdot \sqrt{T}}{\sigma} \right ) ~,
	\eeq
	where $\gamma = c\kappa_{\omega} \sqrt{\log k} (w(A) + \sqrt{\log \log T} + \sqrt{\log (1/\delta)})$.
	\label{thm:single_parameter_gaussian_regret_bounds}
\end{theo}





\subsection{Examples}
We instantiate the regret bounds for a few norms under very mild conditions assuming $\sigma = O \left ( \frac{1}{\sqrt{p}} \right )$. Note that for $\ell_2^2$ regularization the setting is similar to \cite{kmrw18}. The worst case regret bounds are better than \cite{kmrw18} by a factor of $\sqrt{p}$. If $\theta^*$ is sparse exploiting structure, e.g. using the $\ell_1$ norm, the regret bounds depend on $\sqrt{s\log p}$ instead of $\sqrt{p}$.

\begin{corollary}
    Consider the smoothed adversary setting. Let $\sigma = O\left (\frac{1}{\sqrt{p}} \right )$. Then with probability atleast $1 - 4\delta$:
    \begin{enumerate}
        \item Let $\theta^*$ be $s$-sparse, $R(\cdot)$ the $\ell_1$ norm. Then when $T \gg \tilde{\Theta}(s \log p)$:
        \beq
            \text{Reg}(T) = \tilde{O} \left ( \sqrt{s \log p} \sqrt{pT} \right ) ~.
        \eeq
        
        \item Let $\theta^* \in \R^{m \times p}$ be a rank $r$ matrix $r \leq \min \{ m,p \}$, $R(\cdot)$ is the nuclear norm. Then when $T \gg \tilde{\Theta}(r(m+p)) $ :
        \beq
            \text{Reg}(T) = \tilde{O} \left (  \sqrt{r(m+p)} p \sqrt{T} \right ) ~.
        \eeq
        
        \item Let $R(\cdot)$ the $\ell_2^2$ norm. Then when $T \gg \tilde{\Theta}(p)$:
        \beq
            \text{Reg}(T) = \tilde{O} \left ( p \sqrt{T} \right ) ~.
        \eeq
    \end{enumerate}
\end{corollary}

	\section{Multi Parameter Regret Analysis}
	\label{sec:multi_parameter}
	We present results for the multi parameter setting in this section. The multi parameter setting has a separate parameter corresponding to each context. The algorithm 
requires a warm start phase of $T_0$ rounds where the contexts are chosen in a round robin fashion before employing the greedy algorithm. As we show later, the length of the warm start phase has dependence on the variance of the Gaussian perturbations and is required to obtain sublinear regret. Similar to the single parameter setting, after the warm start phase the greedy algorithm proceeds in an episodic fashion, except that we now maintain separate episodes for each context. Denote the episode numbers for context $i$ by $e_i$ and the maximum number of episodes for context $i$ after $T$ round as $e_{i,\max}$. 
In episode $e_i$, context $i$ is chosen by the greedy algorithm $T_{i,e_i}$ times. During episode $e_i$, before context $i$ is chosen in $T_{i,e_i}$ rounds by the greedy algorithm, there can also be rounds when context $i$ was optimal but was not chosen by the algorithm, i.e., $x_i^t = \underset{x_j^t:1 \leq j \leq k}{\argmax} \langle x_j^t, \theta^*_j \rangle$ but $x_i^t \neq \underset{x_j^t:1 \leq j \leq k}{\argmax} \langle x_j^t, \hat{\theta}^{(e_j)}_j \rangle$. We denote the number of rounds this happens in episode $e_i$ by $T_{i,e_i}^*$.

Lemma \ref{lemm:multi_param_reg_bnds} below gives an upper bound for the regret for Algorithm \ref{algo:hd_multi_smooth_bandits}.

\begin{lemm}{\bf (Multi Parameter Regret Bounds)}
	The greedy algorithm plays the contexts in an episodic fashion with the maximum episode number for each context $e_i \leq e_{i,\max} \leq \lfloor \log T \rfloor$. Denote by $\beta = \underset{\substack{{1 \leq i \leq k, 1 \leq t \leq T} \\  {v \in A}}}{\max} \langle x_i^t,v \rangle$.
	Let $t_{\min} < T$, where $t_{\min}$ depends on properties of the true parameters $\theta^*_i$, the regularizer $R(\cdot)$, the noise properties, the number of contexts $k$ and the quantity $\beta$. Then,
	\begin{align}
	&\text{Reg}(T) \leq 2 \beta t_{\min} + \beta \sum_{i=1}^k \sum_{e_i=1}^{e_{i,\max}} \left (  T_{i,e_i} \|\theta^*_{i} - \hat{\theta}_{i}^{(e_i)}\|_2 + T_{i,e_i}^*  \|\theta^*_i - \hat{\theta}_i^{(e_i)}\|_2 \right ) 
	\end{align}
	\label{lemm:multi_param_reg_bnds}
\end{lemm}

The regret thus depends on the following: a) the accuracy of estimating $\theta^*_i$ in each episode for all contexts; b) the number of rounds when any context $i$ is optimal but not chosen,i.e., the quantities $T_{i,e_i}^*$, and c) the number of episodes in each context, i.e., the quantities $e_{i,\max}$. A major difference compared to the single parameter setting is the quantity $T_{i,e_i}^*$ and the relation of the regret with $T_{i,e_i}^*$. Note that the estimate of any context parameter improves with the number of times the particular context is chosen. The quantities $T_{i,e_i}^*$, while contributing to the regret, represent rounds when the context is not chosen and hence do not contribute to improvement of the parameter estimate. In contrast in the single parameter setting, since there is only one parameter, any chosen context contributes towards better parameter estimation rates. We need the warm start to ensure the greedy algorithm chooses contexts with constant probability when they are optimal to limit the quantities $T_{i,e_i}^*$. 

\abcomment{ok, the para above is confusing, but the algorithm is simple--lets explain the algorithm more clearly; discussing the different counters, how they are incremented, and how phase transition for each arm happens may help}

We focus on regret bounds when the contexts are  $x_i^t = \mu_i^t + g_i^t, 1 \leq i \leq k, 1 \leq t \leq T$, where $\mu_i^t$'s are adverserially chosen and $g_i^t$'s are the Gaussian perturbations. We begin with a characterization of the number of rounds required in the warm start phase. Remember, the goal of the warm start phase is to ensure that there is a constant probability the algorithm chooses the optimal arm. 
This is the essence of the margin condition in Lemma \ref{lemm:multi_param_margin_condition}. Propositions \ref{prop:multi_param_prop1} and \ref{prop:multi_param_prop2} build towards the result in Lemma \ref{lemm:multi_param_margin_condition}.  Proposition \ref{prop:multi_param_prop1} is a straightforward observation on the relationship between the first and second optimal contexts where we introduce the quantity $r$. To summarize, Proposition \ref{prop:multi_param_prop1} makes the observation that the dot product between the Gaussian perturbation and parameter of the optimal context exceeds the quantity $r$.
\begin{prop}
    Consider any round $t$ when the episode numbers of the $k$ contexts are $e_1,\hdots,e_k$. Let $i^*$ denote the context with the maximum reward, i.e., $i^* = \underset{l:1 \leq l \leq k}{\argmax} \langle \mu_l^t + g_l^t,\theta^*_l \rangle$. Let $j$ denote the context having the second largest reward, i.e., $j = \underset{l:1 \leq l \leq k;l \neq i^*}{\argmax} \langle \mu_l^t + g_l^t,\theta^*_l \rangle$. Define $r = \langle \mu_j^t + g_j^t,\theta^*_j \rangle - \langle \mu_{i^*}^t,\theta^*_{i^*} \rangle$. Then the following condition is satisfied,
	\beq
	\langle g_{i^*}^t,\theta^*_{i^*} \rangle \geq r ~.
	\eeq
	\label{prop:multi_param_prop1}
\end{prop}

\begin{algorithm}[!t]
	\begin{algorithmic}[1]
		\STATE Set $e_1 = \hdots = e_k = 0$.
		Initialize empty design matrices and rewards $Z^{(0)}_1,\hdots,Z^{(0)}_k = []$,$y^{(0)}_1,\hdots,y^{(0)}_k = []$
		\FOR{$t = 1$ to $T_0$} 
		\STATE Observe contexts $x_1^t,\hdots,x_k^t \in \R^p$
		\STATE Pick context $i^t$ from $\{1,\hdots,k \}$ in round robin fashion and observe reward $r_{i^t}^t = \langle x_{i^t}^t,\theta^*_{i^t} \rangle + \omega^t$ where $\omega^t$ is zero mean $\kappa_{\omega}$-sub-Gaussian noise
		\STATE Append observations $(x_{i^t}^t,r_{i^t}^t)$ to $(Z^{(0)}_{i^t},y^{(0)}_{i^t})$
		\ENDFOR
		\STATE Compute SVD of $\frac{1}{\sqrt{T_{i,0}}} Z_{i}^{(0)} = U_i^{(0)}D_i^{(0)}(V_i^{(0)})^{\intercal}$
		\STATE Define the Puffer transformation $F_i^{(0)} = U_i^{(0)}(D_i^{(0)})^{-1}(U_i^{(0)})^{\intercal}$ and compute $\tilde{y}_i^{(0)} = F_i^{(0)}y_i^{(0)}$ and $\tilde{Z}_i^{(0)} = F_i^{(0)}Z_i^{(0)}$ 
		\STATE Estimate parameters using constrained least squares estimator for each context with $T_{1,0} = \hdots = T_{i,0} = \hdots = T_{k,0} = T_0/k$
		\beq
		\hat{\theta}^{(1)}_i = \underset{\theta \in \R^p}{\argmin} \frac{1}{2T_{i,0}} \|\tilde{y}^{(0)}_i - \tilde{Z}^{(0)}_i \theta\|_2^2 \quad \text{s.t.} \quad
		R(\theta) \leq R(\theta^*_i) ~,
		\eeq
		\STATE Increment all $e_i = e_i + 1, 1 \leq i \leq k$. Initialize empty design matrices and rewards $Z^{(e_1)}_1,\hdots,Z^{(e_k)}_k = []$,$y^{(e_1)}_1,\hdots,y^{(e_2)}_k = []$. Also initialize $t_1 = \hdots = t_k = 0$.
		\FOR{$t = T_0$ to $T$}
		\STATE Observe contexts $x_1^t,\hdots,x_k^t \in \R^p$
		\STATE Pick context $i^t$ such that $i^t = \underset{1 \leq i \leq k}{\argmax} \langle x_i^t, \hat{\theta}_{i}^{(e_i)} \rangle$, receive reward $r^t_{i^t} = \langle x^t_{i^t},\theta^*_{i^t} \rangle + \omega^t$ and increment $t_{i^t} = t_{i^t} + 1$
		\STATE Append observations $(x_{i^t}^t,r_{i^t}^t)$ to $(Z^{(e_{i^t})}_{i^t},y^{(e_{i^t})}_{i^t})$
		\IF{$t_{i^t} = 2 T_{i^t,e_{i^t}-1} = T_{i^t,e_{i^t}}$}
		\STATE Compute SVD of $\frac{1}{\sqrt{T_{i,e_{i^t}}}} Z_{i^t}^{(e_{i^t})} = U_{i^t}^{(e_{i^t})} D_{i^t}^{(e_{i^t})} (V_{i^t}^{(e_{i^t})})^{\intercal}$
		\STATE Compute the Puffer transformation $F_{i^t}^{(e_{i^t})} = U_{i^t}^{(e_{i^t})} (D_{i^t}^{(e_{i^t})})^{-1} (U_{i^t}^{(e_{i^t})})^{\intercal}$ and compute $\tilde{Z}_{i^t}^{(e_{i^t})} = F_{i^t}^{(e_{i^t})}Z_{i^t}^{(e_{i^t})}$ and $\tilde{y}_{i^t}^{(e_{i^t})} = F_{i^t}^{(e_{i^t})} y_{i^t}^{(e_{i^t})}$
		\STATE Estimate parameter using constrained least squares estimator
		\begin{align}
		\hat{\theta}^{(e_{i^t} + 1)}_{i^t} = \underset{\theta \in \R^p}{\argmin} \frac{1}{2T_{i^t,e_{i^t}}} \|\tilde{y}^{(e_{i^t})}_{i^t} - \tilde{Z}^{(e_{i^t})}_{i^t} \theta\|_2^2 \nonumber \\
		\text{s.t.} \quad R(\theta) \leq R(\theta^*_{i^t}) ~,
		\end{align}
		where $T_{i^t,e_{i^t}} = 2 T_{i^t,e_{i^t}-1}$. 
		\STATE Increment $e_{i^t} = e_{i^t} + 1$. Initialize empty design matrix $Z^{(e_{i^t})}_{i^t} = []$ and reward $y^{(e_{i^t})}_{i^t} = []$. Initialize $t_{i^t} = 0$.
		\ENDIF
		\ENDFOR
	\end{algorithmic}
	\caption{High-dimensional Greedy (multi parameter)}
	\label{algo:hd_multi_smooth_bandits}
\end{algorithm}

Proposition \ref{prop:multi_param_prop2} states conditions when the greedy algorithm chooses the optimal context. Due to parameter estimation errors, for the greedy algorithm to perceive the context to be optimal the dot product between the optimal parameter vector and Gaussian perturbation should now exceed $r$ by a quantity which depends on the estimation error. 
\begin{prop}
    Assume context $j'$ such that $j' = \underset{l:1 \leq l \leq k,l \neq i^*}{\argmax} \langle \mu_{l}^t + g_{l}^t, \hat{\theta}_{l}^{(e_{l})} \rangle$, i.e., the context other than $i^*$ which has the highest estimated reward. Also assume the parameter estimate for context $i^*$ to be $\hat{\theta}_{i^*}^{(e_{i^*})} = \theta^*_{i^*} + \Delta_{i^*}^{(e_{i^*})}$ and for context $j'$, $\hat{\theta}_{j'}^{(e_{j'})} = \theta^*_{j'} + \Delta_{j'}^{(e_{j'})}$. Then the greedy algorithm selects context $i^*$  if the following condition is satisfied,
	\beq
	\langle g_{i^*}^t,\theta^*_{i^*} \rangle \geq r + \langle \mu_{j'}^t + g_{j'}^t,\Delta_{j'}^{(e_{j'})} \rangle - \langle \mu_{i^*}^t + g_{i^*}^t, \Delta_{i^*}^{(e_{i^*})} \rangle ~.
	\label{eq:multi_param_margin_cond_eq1}
	\eeq
	\label{prop:multi_param_prop2}
\end{prop}

The greedy algorithm always picks the optimal context if the condition in equation \myref{eq:multi_param_margin_cond_eq1} is satisfied. Let us now fix the quantity $r$. Let the estimation errors after the warm start phase be such that $\abs{g_{j'}^t,\Delta_{j'}^{(e_{j'})} \rangle - \langle \mu_{i^*}^t + g_{i^*}^t, \Delta_{i^*}^{(e_{i^*})} \rangle} \leq \frac{\sigma^2}{r}$. Then the probability that there is a match between the optimal context and the context chosen by the greedy algorithm is precisely the quantity on the l.h.s. in equation \myref{eq:multi_param_margin_cond}. Now what are values of $r$ when equation \myref{eq:multi_param_margin_cond} is satisfied? In the proof provided in the appendix, we will prove that the probability in equation \myref{eq:multi_param_margin_cond} decreases with increasing $r$. Therefore to obtain lower bounds we assume an upper bound on $r$ which we will show to hold with high probability over choices of contexts, $\mu_k^t,g_k^t$, in all rounds. 


\begin{lemm}{\bf (Margin Condition)}
	Consider good events as when $r \leq c_3 \sigma \sqrt{\log (Tk)}$ and consider errors $\Delta_{i^*}^{(e_{i^*})}$ and $\Delta_{j'}^{(e_{j'})}$ to be small enough such that $\langle \mu_{j'}^t + g_{j'}^t,\Delta_{j'}^{(e_{j'})} \rangle - \langle \mu_{i^*}^t + g_{i^*}^t, \Delta_{i^*}^{(e_{i^*})} \rangle \leq \frac{\sigma^2}{r}$. Then the following holds,
	\beq
	P \left ( \langle g_{i^*}^t,\theta^*_{i^*} \rangle \geq r + \frac{\sigma^2}{r} \suchthat \langle g_{i^*}^t,\theta^*_{i^*} \rangle \geq r \right ) \geq \frac{1}{20} ~,
	\label{eq:multi_param_margin_cond}
	\eeq
	for all $r \leq c_3 \sigma \sqrt{\log (Tk)}$.
	\label{lemm:multi_param_margin_condition}
\end{lemm}
The length of the warm start phase is now influenced by the condition that $\|\Delta_{j'}^{(e_{j'})}\|_2$ and $\|\Delta_{i^*}^{(e_{i^*})}\|_2$ are small enough so that $\langle \mu_{j'}^t + g_{j'}^t,\Delta_{j'}^{(e_{j'})} \rangle - \langle \mu_{i^*}^t + g_{i^*}^t, \Delta_{i^*}^{(e_{i^*})} \rangle \leq \frac{\sigma^2}{r}$ in Lemma \ref{lemm:multi_param_margin_condition} which translates to the upper bound below: 
\beq
    \|\Delta_i^{(e_i)}\|_2 = \|\hat{\theta}_i^{(e_i)} - \theta^*_{i} \|_2 \leq \tilde{O} \left ( \sigma \right ) ~.
    \label{eq:ch6_multi_param_est_err_req}
\eeq
The estimation error bounds are in turn influenced by the properties of the design matrices after the warm start phase.
\begin{lemm}{\bf (Multi parameter Design Matrix Properties)}
	Consider any context $i$ and a particular episode $e_i$. The rows of the design matrix $Z^{(e_i)}_{i} \in \R^{T_{i,e_i} \times p}$ are $z^t_i = \mu^t_i + g^t_i$ where in round $t$ context $i$ is chosen by the Greedy algorithm, i.e., $i = \underset{1 \leq l \leq k}{\argmax}  \langle x_l^t,\hat{\theta}_l^{(e_l)} \rangle$ where $x^t_l = \mu_l^t + g^t_l, ~ g_l^t \sim N(0,\sigma^2 \I_{p \times p})$. Then under the condition $\langle g_i^t,\theta^*_i \rangle \geq r$ for some $r \leq c_3 \sigma \sqrt{\log (Tk)}$,
	\begin{equation*}
	\lambda_{\min} \left ( E_{z^t} \left [ z^t_i (z^t_i)^{\intercal}  \suchthat z^t_i \text{ satisfies } \zeta \right ] \right ) \geq c_2 \frac{\sigma^2}{\log (Tk)} ~,
	\end{equation*} 
	where $\zeta$ is the condition $z^t_i = \underset{g_l^t:1 \leq l \leq k}{\argmax} \langle x_l^t, \hat{\theta}_l^{(e_l)} \rangle;\langle g_i^t,\theta^*_i \rangle \geq r; r \leq c_3 \sigma \sqrt{\log (Tk)}$.
	\label{lemm:multi_param_prop_design_matrices}
\end{lemm}
The only difference in the properties of the design matrix compared to the single parameter setting are the sub-Gaussian norm and expected minimum eigenvalue of the covariance matrix. Using similar steps to derive estimation error as in the single parameter setting,
we obtain the following upper bound on the maximum estimation error across all contexts and episodes with high probability:
\beq
	\sup\limits_{1 \leq i \leq k} \sup\limits_{e_i \leq e_{i,\max}} \|\hat{\theta}_i^{(e_i + 1)} - \theta^*_i\|_2 \leq \tilde{O} \left ( \frac{w(A)}{\sigma \sqrt{T_{i,e_i}}} \right ) ~.
	\label{eq:ch6_multi_param_est_err_act}
\eeq
Comparing equations \myref{eq:ch6_multi_param_est_err_req} and \myref{eq:ch6_multi_param_est_err_act} it can be easily inferred that $T_{i,e_i} = \tilde{\Theta} \left ( \frac{w^2(A)}{\sigma^4} \right )$ to satisfy the margin condition and since the episode length increases monotonically the length of the warm start phase $T_0 = \tilde{\Theta} \left ( \frac{k w^2(A)}{\sigma^4} \right )$. 

After the warm start phase, the margin condition of Lemma \ref{lemm:multi_param_margin_condition} holds which ensures that the greedy algorithm chooses the optimal context with probability atleast $1/20$.
In other words in expectation $T_{i,e_i}^* \leq 20 T_{i,e_i}$, i.e., in any particular episode for any context the number of rounds when the context is optimal but not perceived to be optimal by the greedy algorithm is upper bounded by $20$ times the length of the episode. With the result on $T_{i,e_i}^*$'s and the upper bound on the parameter estimation errors, the regret in the multi parameter setting can be derived from the result of Lemma \ref{lemm:multi_param_reg_bnds}.  

\begin{theo}{\bf (Multi parameter Smoothed Adversary Regret Bounds)}
	Consider computation of regret for the Greedy algorithm in the multi parameter setting following Lemma 4. Define the following quantities $r \leq c_3 \sigma \sqrt{\log (Tk)}$, $\gamma = \frac{c_{12} \kappa_{\omega} (w(A) + \sqrt{\log \log T} + \sqrt{\log k} + \sqrt{\log(1/\delta)}) \sqrt{\log (Tk)}}{\sigma}$ and $\beta = \underset{\substack{{1 \leq i \leq k, 1 \leq t \leq T} \\  {v \in A}}}{\max} \langle x_i^t,v \rangle$. The margin condition in Lemma 5 is satisfied with probability atleast $1 - 5\delta$ when,
	\beq
	t_{\min} \geq \frac{4k\gamma^2 r^2 \beta^2}{\sigma^4} + 1 + \sqrt{\frac{1}{2}\log(1/\delta)} ~.
	\eeq
	Under the margin condition, the regret is maximized when in each round each context has equal probability to be selected by the Greedy algorithm. The equal probability implies that in expectation $T_1 = T_2 = \hdots = T_k = \frac{T}{k}$. Also the regret is upper bounded as follows,
	\beq
	\text{Reg}(T) \leq 2 \beta t_{\min} + 82 \beta \gamma \sqrt{Tk} \log (T) ~.
	\eeq
	Moreover $\beta = \underset{\substack{{1 \leq i \leq k, 1 \leq t \leq T} \\  {v \in A}}}{\max} \langle x_i^t,v \rangle \leq (1+ c_1 \sigma (w(A) + \sqrt{\log (1/\delta)}))$ with probability atleast $1 - \delta$. 
	Therefore with probability atleast $1 - 6\delta$ when $T \gg t_{\min}$,
	\beq
	\text{Reg}(T) \leq O\left ( \gamma \cdot \beta \cdot \log(T) \cdot \sqrt{Tk} \right )
	\eeq
	\label{thm:multi_parameter_bandits_reg_bnds}
\end{theo}
The regret is $\sqrt{k}$ times worse than the single parameter setting.

\subsection{Examples}
We instantiate the regret bounds for a few norms. When $R(\cdot)$ is $\|\cdot\|_2^2$, the length of the warm start phase is $\tilde{\Theta}(kp^3)$ which improves over the $\tilde{\Theta}(kp^6)$ obtained in \cite{kmrw18}. Ignoring logarithm terms the regret bounds are of the same order as \cite{aggo13} after the warm start phase but the polynomial in $p$ warm start rounds maybe prohibitive in many applications. 

\begin{corollary}
    Let $\sigma = O \left ( \frac{1}{\sqrt{p}} \right )$. Then with probability atleast $1 - 8\delta$:
    \begin{enumerate}
        \item Let $\theta^*$ be $s$-sparse, $R(\cdot)$ the $\ell_1$ norm, then when $T \gg \tilde{\Theta}(kp^2 s \log p)$:
        \beq
            \text{Reg}(T) = \tilde{O} \left ( \sqrt{p} \sqrt{s \log p} \sqrt{Tk} \right ) ~.
        \eeq
        
        \item Let $\theta^* \in \R^{m \times p}$ be a rank $r$ matrix $r \leq \min \{ m,p \}$, $R(\cdot)$ is the nuclear norm, then when $T \gg \tilde{\Theta}(k p^2 r(m+p)) $ :
        \beq
            \text{Reg}(T) = \tilde{O} \left ( p \sqrt{r(m+p)} \sqrt{Tk} \right ) ~.
        \eeq
        
        \item Let $R(\cdot)$ the $\ell_2^2$ norm, then when $T \gg \tilde{\Theta}(kp^3)$:
        \beq
            \text{Reg}(T) = \tilde{O} \left ( p \sqrt{Tk} \right ) ~.
        \eeq
    \end{enumerate}
\end{corollary}
	
	\section{Conclusions}
	\label{sec:main_conc}
	We analyzed the structured linear contextual bandit problem under the smoothed analysis framework. Our analysis significantly improves on the bounds obtained in \cite{kmrw18}. While previous work have found it difficult to extend exploration strategies to the structured setting with simultaneously exploiting the structure in the parameter, our analysis shows that a simple greedy algorithm achieves sublinear regret under the smoothed bandits framework.

\vspace{5mm}
{\bf Acknowledgements:} The research was supported by NSF grants OAC-1934634, IIS-1908104, IIS-1563950, IIS-1447566, IIS-1447574, IIS-1422557, CCF-1451986, FAI-1939606, a Google Faculty Research Award, a J.P. Morgan Faculty Award, and a Mozilla research grant. Part of this work completed while ZSW was visiting the Simons Institute for the Theory of Computing at UC Berkeley.
	
\bibliography{references}

\begin{thebibliography}{10}

\bibitem{abps11}
Yasin Abbasi-Yadkori, D{\'{a}}vid P{\'{a}}l, and Csaba Szepesv{\'{a}}ri.
\newblock {Online Least Squares Estimation with Self-Normalized Processes: An
  Application to Bandit Problems}.
\newblock In {\em Conference on Learning Theory (COLT)}, 2011.

\bibitem{abps12}
Yasin Abbasi-Yadkori, D{\'{a}}vid P{\'{a}}l, and Csaba Szepesv{\'{a}}ri.
\newblock {Online-to-Confidence-Set Conversions and Application to Sparse
  Stochastic Bandits}.
\newblock In {\em International Conference on Artificial Intelligence and
  Statistics (AISTATS)}, 2012.

\bibitem{aggo13}
Shipra Agarwal and Navin Goyal.
\newblock {Thompson Sampling for Contextual Bandits with Linear Payoffs}.
\newblock In {\em International Conference on Machine Learning (ICML)}, 2013.

\bibitem{arfs12}
Andreas Argyriou, Rina Foygel, and Nathan Srebro.
\newblock {Sparse Prediction with the $k$-Support Norm}.
\newblock In {\em Neural Information Processing Systems (NIPS)}, 2012.

\bibitem{bcfs14}
Arindam Banerjee, Sheng Chen, Farideh Fazayeli, and Vidyashankar Sivakumar.
\newblock {Estimation with Norm Regularization}.
\newblock In {\em Neural Information Processing Systems (NIPS)}, 2014.

\bibitem{bgsw19}
Arindam Banerjee, Qilong Gu, Vidyashankar Sivakumar, and Zhiwei~Steven Wu.
\newblock Random quadratic forms with dependence: Applications to restricted
  isometry and beyond.
\newblock In {\em Advances in Neural Information Processing Systems (NIPS)},
  2019.

\bibitem{bastani2017exploiting}
Hamsa Bastani, Mohsen Bayati, and Khashayar Khosravi.
\newblock Mostly exploration-free algorithms for contextual bandits.
\newblock {\em CoRR arXiv:1704.09011}, 2018.
\newblock Working paper.

\bibitem{birt09}
Peter~J. Bickel, Ya'acov Ritov, and Alexandre~B. Tsybakov.
\newblock {Simultaneous analysis of Lasso and Dantzig selector}.
\newblock {\em The Annals of Statistics}, 37(4):1705--1732, 2009.

\bibitem{practicalCB-arxiv18}
Alberto Bietti, Alekh Agarwal, and John Langford.
\newblock Practical evaluation and optimization of contextual bandit
  algorithms.
\newblock {\em CoRR arXiv:1802.04064}, 2018.

\bibitem{bbcd16}
Sarah Bird, Solon Barocas, Kate Crawford, Fernando Diaz, and Hanna Wallach.
\newblock Exploring or exploiting? social and ethical implications of
  automonous experimentation.
\newblock In {\em Workshop on Fairness, Accountability, and Transparency in
  Machine Learning}, 2016.

\bibitem{care09}
Emmanuel~J. Cand{\`{e}}s and Benjamin Recht.
\newblock {Exact Matrix Completion via Convex Optimization}.
\newblock {\em Foundations of Computational Mathematics}, 9(6):717--772, 2009.

\bibitem{crpw12}
Venkat Chandrasekaran, Benjamin Recht, Pablo~A. Parrilo, and Alan~S. Willsky.
\newblock {The Convex Geometry of Linear Inverse Problems}.
\newblock {\em Foundations of Computational Mathematics}, 12(6):805--849, 2012.

\bibitem{chba15}
Sheng Chen and Arindam Banerjee.
\newblock {Structured Estimation with Atomic Norms: General Bounds and
  Applications}.
\newblock In {\em Neural Information Processing Systems (NIPS)}, 2015.

\bibitem{chu2011contextual}
Wei Chu, Lihong Li, Lev Reyzin, and Robert~E. Schapire.
\newblock Contextual bandits with linear payoff functions.
\newblock In {\em International Conference on Artificial Intelligence and
  Statistics (AISTATS)}, 2011.

\bibitem{dahk08}
Varsha Dani, Thomas~P. Hayes, and Sham~M. Kakade.
\newblock {Stochastic Linear Optimization Under Bandit Feedback}.
\newblock In {\em Conference on Learning Theory (COLT)}, 2008.

\bibitem{gord85}
Y.~Gordon.
\newblock Some inequalities for gaussian processes and applications.
\newblock {\em Israel Journal of Mathematics}, 50(4):265--289, 1985.

\bibitem{hand14}
Ramon~van Handel.
\newblock {Probability in High Dimensions}.
\newblock Technical report, Princeton University, 2014.

\bibitem{jaov09}
L.~Jacob, O.~Obozinski, and J.~P. Vert.
\newblock {Group Lasso with Overlap and Graph Lasso}.
\newblock In {\em International Conference on Machine Learning (ICML)}, number
  2009, 2009.

\bibitem{jana18}
Adel Javanmard and Hamid Javadi.
\newblock {Dynamic Pricing in High Dimensions}.
\newblock {\em Accepted in JMLR}, 2018.

\bibitem{jiro15}
Jinzhu Jia and Karl Rohe.
\newblock Preconditioning the lasso for sign consistency.
\newblock {\em Electronic Journal of Statistics}, 9:1150--1172, 2015.

\bibitem{kmrw18}
Sampath Kannan, Jamie Morgenstern, Aaron Roth, Bo~Waggoner, and Zhiwei~Steven
  Wu.
\newblock A smoothed analysis of the greedy algorithm for the linear contextual
  bandit problem.
\newblock {\em CoRR arXiv:1801.04323}, 2018.

\bibitem{Langford-nips07}
John Langford and Tong Zhang.
\newblock {The Epoch-Greedy Algorithm for Contextual Multi-armed Bandits}.
\newblock In {\em Advances in Neural Information Processing Systems (NIPS)},
  2007.

\bibitem{Langford-www10}
Lihong Li, Wei Chu, John Langford, and Robert~E. Schapire.
\newblock {A contextual-bandit approach to personalized news article
  recommendation}.
\newblock In {\em International World Wide Web Conference (WWW)}, 2010.

\bibitem{competingBandits-itcs16}
Yishay Mansour, Aleksandrs Slivkins, and Zhiwei~Steven Wu.
\newblock Competing bandits: Learning under competition.
\newblock In {\em Innovations in Theoretical Computer Science (ITCS)}, 2018.

\bibitem{mept07}
S.~Mendelson, A.~Pajor, and N.~Tomczak-Jaegermann.
\newblock Reconstruction and sub{G}aussian operators in asymptotic geometric
  analysis.
\newblock {\em Geometric and Functional Analysis}, 17:1248--1282, 2007.

\bibitem{nrwy12}
Sahand~N. Negahban, Pradeep Ravikumar, Martin~J. Wainwright, and Bin Yu.
\newblock {A Unified Framework for High-Dimensional Analysis of
  {\$}M{\$}-Estimators with Decomposable Regularizers}.
\newblock {\em Statistical Science}, 27(4):538--557, 2012.

\bibitem{externalities-colt18}
Manish Raghavan, Aleksandrs Slivkins, Jennifer~Wortman Vaughan, and
  Zhiwei~Steven Wu.
\newblock The externalities of exploration and how data diversity helps
  exploitation.
\newblock In {\em Conference on Learning Theory (COLT)}, pages 1724--1738,
  2018.

\bibitem{sibp15}
V.~Sivakumar, A.~Banerjee, and P.~Ravikumar.
\newblock Beyond sub-gaussian measurements: High-dimensional structured
  estimation with sub-exponential designs.
\newblock In {\em Advances in Neural Information Processing Systems (NIPS)},
  2015.

\bibitem{siba17}
Vidyashankar Sivakumar and Arindam Banerjee.
\newblock {High-Dimensional Structured Quantile Regression}.
\newblock In {\em International Conference on Machine Learning (ICML)}, 2017.

\bibitem{tala05}
Michel Talagrand.
\newblock {\em {The Generic Chaining}}.
\newblock Springer Monographs in Mathematics. Springer Berlin, 2005.

\bibitem{tala14}
Michel Talagrand.
\newblock {\em {Upper and Lower Bounds of Stochastic Processes}}.
\newblock Springer, 2014.

\bibitem{tibs96}
Robert Tibshirani.
\newblock {Regression Shrinkage and Selection via the Lasso}.
\newblock {\em Journal of the Royal Statistical Society}, 58(1):267--288, 1996.

\bibitem{vers12}
Roman Vershynin.
\newblock {Introduction to the non-asymptotic analysis of random matrices}.
\newblock In Y~Eldar and G.~Kutyniok, editors, {\em Compressed Sensing}, pages
  210--268. Cambridge University Press, Cambridge, nov 2012.

\bibitem{vers18}
Roman Vershynin.
\newblock {\em High-Dimensional Probability: An Introduction with Applications
  in Data Science}.
\newblock Cambridge Series in Statistical and Probabilistic Mathematics.
  Cambridge University Press, 2018.

\bibitem{wain19}
Martin Wainwright.
\newblock {\em High-Dimensional Statistics: A Non-Asymptotic Viewpoint}.
\newblock Cambridge University Press (To appear), 2019.

\bibitem{yuli06}
Ming Yuan and Yi~Lin.
\newblock {Model Selection and Estimation in Regression With Grouped
  Variables}.
\newblock {\em Journal of the Royal Statistical Society}, 68(1):49--67, 2006.

\end{thebibliography}
\bibliographystyle{plain}

\vspace{5mm}

\appendix

    \section{Background and Preliminaries}
	\label{sec:supp_back}
	We provide definitions of important entities and some well-known results that will be used throughout the proofs. 

\subsection{Random Variables, Vectors and Concentration Inequalities}

We briefly review definitions and properties of random variables and vectors. We borrow from \cite{vers12} which is a more thorough and easily accessible exposition of the below material.

\subsubsection{Sub-Gaussian Random Variables}
We define and state properties of sub-Gaussian random variables.
\begin{defn} {\bf (Sub-Gaussian random variables)}
	A random variable $x$ is sub-Gaussian if it satisfies any of the following properties for positive constants $\kappa_1,\kappa_2,\kappa_3$,
	\begin{enumerate}
		\item Tails: $P(|x| > t) \leq \exp(1 - t^2/\kappa_1^2),\quad t \geq 0$;
		\item Moments: $(E|x|^p)^{1/p} \leq \kappa_2 \sqrt{p}, \quad \forall p \geq 1$;
		\item Super-exponential moment: $E\exp(x^2/\kappa_3^2) \leq e$.
	\end{enumerate}
	Moreover the sub-Gaussian norm of the random variable, denoted as $\|x\|_{\psi_2}$, is the smallest $\kappa_2$ such that,
	\beq
	\|x\|_{\psi_2} = \sup\limits_{p \geq 1} p^{-1/2}(E|x|^p)^{1/p} ~.
	\eeq	
\end{defn}
The tail decay, moment growth and growth of moment generating function in the definition are equivalent with each implying the others with the constants $\kappa_1,\kappa_2,\kappa_3$ differing from each other by at most an absolute constant factor. The zero mean, $\sigma^2$-variance Gaussian distribution $N(0,\sigma^2)$ is a sub-Gaussian distribution with sub-Gaussian norm $c\sigma$ for some constant $c$.

We characterize large deviation properties of sums of sub-Gaussian random variables below.

\begin{lemm} {\bf (Hoeffding-type inequality)}
	Let $x_1,\hdots,x_n$ be independent centered sub-Gaussian random variables. Let $\kappa = \max\limits_{1 \leq i \leq n} \|x_i\|_{\psi_2}$. Then for any $a \in \R^n$ and $t \geq 0$, we have,
	\beq
	P \left ( \left | \sum_{i=1}^n a_i x_i \right | \geq t \right ) \leq e \cdot \exp \left ( - \frac{ct^2}{\kappa^2 \|a\|_2^2} \right ) ~,
	\eeq
	where $c > 0$ is an absolute constant.
	\label{lemm:hoeffding}
\end{lemm}

Sub-Gaussian random variables are rotationally invariant.
\begin{lemm}{\bf (Rotation invariance)}
	Consider a finite number of independent centered sub-gaussian random variables $x_i$. Then $\sum_{i}x_i$ is also a centered sub-gaussian random variable. Moreover,
	\beq
	\| \sum_{i}x_i\|^{2}_{\psi_2} \leq c \sum_{i}\|x_i\|_{\psi_2}^2
	\eeq
	\label{lemm:rotinvsubg}
\end{lemm}

\subsubsection{Random Vectors}
We will work with random vectors $x \in \R^p$ which are samples from a probability distribution in $\R^p$.

\begin{defn} (Isotropic random vectors)
	A random vector $x \in \R^p$ is isotropic if $\Sigma = E[x x^T] = \I_{p \times p}$. Equivalently, $E[\langle x,u \rangle^2] = \|u\|_2^2$ for any $u \in \R^p$.
\end{defn}

An example of an isotropic random vector is the $p$-dimensional Gaussian random vector $x \sim N(0,\sigma^2 \I_{p \times p})$. Let $\Sigma = E[x x^T]$ be an invertible matrix, which is true if the probability distribution, from which $x$ is sampled, is not supported in any proper subspace of $\R^p$. Then $\Sigma^{-1/2}x$ is an isotropic random vector. 

\begin{defn} {\bf (Sub-Gaussian random vectors)}
	A random vector $x \in \R^p$ is sub-Gaussian if the one-dimensional marginals $\langle x,u \rangle$ are sub-Gaussian random variables for all $u \in \R^p$. The sub-Gaussian norm of $x$ is defined as,
	\beq
	\|x\|_{\psi_2} = \sup\limits_{u \in S^{p-1}} \|\langle x,u \rangle \|_{\psi_2} ~.
	\eeq
\end{defn}

A random vector with sub-Gaussian elements is a sub-Gaussian random vector.
\begin{lemm}{\bf (Product of sub-Gaussian distributions)}
	Let $x_1, \ldots, x_p$ be independent centered sub-gaussian random variables. Then $x = (x_1, \ldots, x_p)$ is a centred sub-gaussian random vector in $\R^p$, and
	\beq
	\|x\|_{\psi_2} \leq c \max\limits_{i \leq p} \|x_i\|_{\psi_2}
	\eeq
	where $c$ is an absolute constant.
	\label{lemm:prosubg}
\end{lemm}

Projections of sub-Gaussian random vectors in any direction is a sub-Gaussian random variable.
\begin{lemm}
	Consider a sub-Gaussian random vector $x \in \R^p$ with sub-Gaussian norm $\kappa = \max_{i} \| x_i \|_{\psi_2}$, then, $z = \langle x, a \rangle$ is a sub-Gaussian random variable with sub-Gaussian norm $\|z\|_{\psi_2} \leq c \kappa \| a \|_2$ for some absolute constant $c$.
	\label{lemm:rotsubGau}
\end{lemm}

\subsection{Gaussian Widths}
Informally speaking, widths of sets \cite{tala05,tala14} can be seen as measures for the complexity of sets. The non-asymptotic estimation error bounds for the estimators we consider will be expressed in terms of the Gaussian/exponential widths of sets related to the norm $R(\cdot)$. For example, the Gaussian width of the unit norm ball $\Omega_R = \{ u \in \R^p ~ | ~ R(u) \leq 1 \}$ is a common term which shows up in all results.We provide informal definitions for the width of sets and state a few properties useful in analysis of high-dimensional estimators. While we will only describe aspects relevant to this work, widths and the associated tools like generic chaining are deep topics to which entire books have been devoted \cite{tala05,tala14}. 

\begin{defn} {\bf (Gaussian Width)}
	Consider any set $T \subseteq \R^p$. Let $\{X_t\}_{t \in T} = \langle g,t \rangle$ be a stochastic process indexed by the set $T$, where each element $g_i \sim N(0,1),~ 1 \leq i \leq p$ is i.i.d zero mean variance one Gaussian. The quantity $w_g(T) = E_g \left [ \sup\limits_t X_t \right ]$ is called the Gaussian width of the set $T$.
\end{defn}

More generally any stochastic process $\{X_t\}_{t \in T}$ indexed by the set $T \subseteq \R^p$ satisfying the following Hoeffding-type increment condition for some constant $\kappa$,
\beq
\forall u > 0, ~ P(|X_s - X_t| \geq u) \leq 2\exp \left ( - \frac{u^2}{\kappa^2\|s-t\|_2^2} \right ) ~,
\eeq
satisfies the following for some constant $c$ which depends on $\kappa$ due to the majorizing measures theorem,
\beq
E \left [ \sup\limits_{t \in T} X_t \right ] \leq c \cdot w_g(T) ~.
\eeq

Below we state a couple of useful properties of widths of sets.

\begin{prop}{\bf (Properties of width)}
Let $w(\cdot)$ denote the Gaussian or exponential width. Consider set $A \subseteq \R^p$.
\begin{enumerate}
    \item Widths are invariant under orthogonal and linear transformations, i.e., for some unitary matrix $Q \in \R^{p \times p}$ and vector $b \in \R^p$
    \beq
        w(A) = w(QA) \quad w(A+b) = w(A) ~,
    \eeq
    where $QA = \{Qu ~ \suchthat ~ u \in A \}$ and $A + b = \{u + b ~ \suchthat ~ u \in A \}$.
    
    \item Width is invariant under taking the convex hull.
    \beq
        w(\text{conv}(A)) = w(A) ~.
    \eeq

\end{enumerate}

\end{prop}

\subsection{Atomic Norms}
We will consider the class of atomic norms for the regularizer. Consider a set $\cA \subseteq \R^p$ which is a collection of atoms that is compact, centrally symmetric about the origin (that is, $a \in \cA \implies -a \in \cA$). Let $\|\theta\|_{\cA}$ denote the gauge of $\cA$. Then the atomic norm regularizer is defined as follows,
\begin{align}
R(\theta) = \|\theta\|_{\cA} &= \inf \{ t > 0 : \theta \in t ~ \text{conv}(\cA) \} \\
&= \inf \left\{ \sum_{a \in \cA} c_a ~ : ~ \theta = \sum_{a \in \cA} c_a a, ~ c_a \geq 0, ~\forall a \in \cA  \right\} ~.
\end{align}
For example when $\cA = \{ \pm e_i \}_{i=1}^p$ yields $R(\theta) = \|\theta\|_{\cA} = \|\theta\|_1$.

Although the atomic set $\cA$ may contain uncountably many elements, for many popular vector norms $\cA$ can be expressed as a union of $q$-dimensional subspaces, $\cA = \cA_1 \cup \cA_2 \cup \hdots \cup \cA_m$ \cite{chba15}. We will consider a few such regularizers as examples throughout the paper including the $\ell_1$-norm, $k$-support norm and nuclear norm. More details on the atomic norms considered in this work can be found in \cite{siba17}.

\subsection{Hoeffding-type Bound for Dependent Variables}

Finally, we present a variant of the Hoeffding bound where the coefficients can depend on the randomness of prior random variables.
\begin{lemm}
Let $\{Z_t\}$ be a sub-Gaussian martingale difference sequence (MDS) and let $z_{1:t}$ denote a realization of $Z_{1:t}$. Let $\{a_t\}$ be a 
sequence of random variables such that $a_t = f_t(z_{1:(t-1)})$ for some sequence function $f_t$ with $|a_t| \leq \alpha_t$ a.s.~for suitable constants $\alpha_1,\ldots,\alpha_T$. Then, for any $\tau > 0$, we have
\beq
P \left( \left|\sum_{t=1}^T a_t z_t \right| \geq \tau \right) \leq 2 \exp \left\{ - \frac{\tau^2}{4c\kappa^2 \sum_{t=1}^T \alpha_t^2}  \right\}~,
\label{eq:ab1}
\eeq
for absolute constants $c > 0$ and where $\kappa$ is the $\psi_2$-norm of the conditional subGaussian random variables.
\label{lemm:ch6_hoeffding_martingale}
\end{lemm}
\proof For any realization $z_{1:(t-1)}$ since $Z_t|z_{1:(t-1)}$ is a sub-Gaussian random variable with zero mean, then the conditional moment-generating function (MGF) satisfies: for all $s > 0$
\beq
E[\exp(sZ_t) \mid z_{1:(t-1)}] \leq \exp(c s^2 \kappa^2)~,
\label{eq:mgf1}
\eeq
where $\kappa$ is $\psi_2$-norm of $Z_t$ conditioned on any realization $z_{1:(t-1)}$ and $c >0 $ is an absolute constant. Further, for $a_t = f(z_{1:(t-1)})$ with $|a_t| \leq \alpha_t$, we have
\begin{equation}
E[\exp(sa_t Z_t) \mid z_{1:(t-1)}] \leq \exp(c a_t^2 s^2 \kappa^2) \leq \exp(c \alpha_t^2 s^2 \kappa^2)~,
\label{eq:mgf2}
\end{equation}
where the last inequality holds for all realiztions $z_{1:(t-1)}$. 

For any $s > 0$, note that
\begin{align}
P \left( \sum_{t=1}^T a_t Z_t \geq \tau \right) &= P\left( \exp\left( s\sum_{t=1}^T a_t Z_t \right) \geq \exp(s\tau) \right) \nonumber  \\ 
&\leq \exp(-s \tau) E\left[ \exp\left( s\sum_{t=1}^T a_t Z_t \right) \right]~.
\label{eq:bnd1}
\end{align}
Now, using~\eqref{eq:mgf1}, we have 
\begin{align*}
E\left[ \exp\left( s\sum_{t=1}^T a_t Z_t \right) \right] &= E_{(Z_1,\ldots,Z_T)}\left[ \prod_{t=1}^T \exp (s a_t Z_t) \right] ~\\
&= E_{(Z_1,\ldots,Z_{T-1})}\left[ E_{Z_T|Z_1,\ldots,Z_{T-1}}\left[ \exp( s a_T Z_T )\right] \prod_{t=1}^{T-1} \exp (s a_t Z_t) \right]~\\
& \leq \exp( c s^2 \alpha_T^2 \kappa^2) E_{(Z_1,\ldots,Z_{T-1})}\left[ \prod_{t=1}^{T-1} \exp (s a_t Z_t) \right]\\
& \leq  \exp( c s^2 \alpha_T^2 \kappa^2) \exp(c s^2 \alpha_{T-1}^2 \kappa^2) E_{(Z_1,\ldots,Z_{T-2})}\left[ \prod_{t=1}^{T-2} \exp (s a_t Z_t) \right] \\
& \leq \exp\left( c s^2 \kappa^2 \sum_{t=1}^T \alpha_t^2 \right)~.
\end{align*}
Plugging this back to \myref{eq:bnd1}, we have
\begin{align}
P & \left( \sum_{t=1}^T a_t Z_t \geq \tau \right) \leq \exp \left(-s \tau + c s^2 \kappa^2 \sum_{t=1}^T \alpha_t^2 \right)~.
\end{align}
Choosing $s = \frac{\tau}{2c \kappa^2 \sum_{t=1}^T \alpha_t^2}$, we obtain 
\beq
P \left( \sum_{t=1}^T a_t Z_t \geq \tau \right) \leq \exp \left\{ -  \frac{\tau^2}{4c\kappa^2 \sum_{t=1}^T \alpha_t^2} \right\}~.
\eeq
Repeating the same argument with $-Z_t$ instead of $X_t$, we obtain the same bound for $P( - \sum_t a_t Z_t \geq \tau)$. Combining the two results gives us \myref{eq:ab1}. \qed
	
	\section{Results on Gaussian Random Variables}
	\label{sec:supp_gauss_results}
	\begin{lemm}
	Consider $k$ Gaussians $g_1,\hdots,g_k$ sampled from a $N(0,\sigma^2)$ distribution. Let $g_{(1)} = \underset{g_i:1 \leq i \leq k}{\max} g_i$. Then for some constant $c_2$,
	\beq
	P(g_{(1)} \leq \sqrt{2} \sigma (\sqrt{\log k} + \sqrt{\log(1/\delta)})) \geq 1 - 2\delta
	\eeq
	\label{lemm:general_max_gauss_bnds}
\end{lemm}
\proof We first obtain upper bounds on $g_{(1)}$. We make the following observations,
\begin{align}
\exp(t E[g_{(1)}]) &\leq E[\exp(tg_{(1)})] \nonumber \\
&\leq E[\max \exp(t g_i)] \nonumber \\
&\leq \sum_{i=1}^k E[\exp(t g_i)] \nonumber \\
&\leq n \exp(t^2 \sigma^2/2) ~,
\end{align}
where the first line is due to Jensen's inequality, the second is the union bound, and the final line follows from the definition of the moment generating function. Taking logarithm of both sides of the inequality, we get
\beq
E[g_{(1)}] \leq \frac{\log k}{t} + \frac{t\sigma^2}{2} ~.
\eeq
This can be minimized by setting $t = \frac{\sqrt{2\log k}}{\sigma}$ to give,
\beq
E[g_{(1)}] \leq \sigma \sqrt{2} \sqrt{\log k} ~.
\label{eq:general_max_gaussian_bnds_eq1}
\eeq
We now use the following result from \cite{tala05} to provide large deviation bounds around $E[g_{(1)}]$.
\begin{lemm}{\bf (Lemma 2.1.3 in \cite{tala05})}
	Consider a Gaussian process $(Z_t)_{t \in U}$, where $U$ is finite and a number $\sigma$ such that $\sigma \geq \sup\limits_{t \in U}(E [Z_t^2])^{1/2}$. Then for $u > 0$ we have,
	\beq
	P \left ( \left | \sup\limits_{t \in U} Z_t - E \sup\limits_{t \in U} Z_t \right | \geq u \right ) \leq 2 \exp\left ( - \frac{u^2}{2\sigma^2} \right ) ~.
	\eeq
\end{lemm}

In the context of our setting we have $g_i = Z_i$, $g_{(1)} = \max\limits_{t \in U} Z_t$ and $\max\limits_{t \in U}(E [Z_t^2])^{1/2} = E[g_i^2]^{1/2} = \sigma$. Therefore using $u = \sigma \sqrt{2} \sqrt{\log(1/\delta)}$ we get,
\beq
P(g_{(1)} \geq E[g_{(1)}] + \sigma \sqrt{2} \sqrt{\log (1/\delta)}) \leq 2\delta ~,
\label{eq:general_max_gaussian_bnds_eq2}
\eeq
Now the stated result can be derived from \myref{eq:general_max_gaussian_bnds_eq1} and \myref{eq:general_max_gaussian_bnds_eq2}. \qed

\begin{lemm}
	Let $x_1,\hdots,x_k$ be $k$ independent Gaussian random variables with variance $\sigma^2$ and let $z = \underset{1 \leq i \leq k}{\argmax} \ x_i$. Then,
	\beq
	\text{Var}(z) \geq c_1 \frac{\sigma^2}{\log k} ~,
	\eeq
	where $c_1$ is some positive constant.
	\label{lemm:max_gauss_variance}
\end{lemm}

\abcomment{is this result related to Sudakov minorization? not necessary for the current work, but if we can establish such a connecion, it will simplify future work}

\abcomment{declare math operator for Var}

\proof For each $i$, let $e_i$ denote the event of $i = \underset{i'}{\argmax} ~ x_{i'}$. Then the variance of $z$ can be written as
\begin{align}
\text{Var}(z) &\geq \sum_i P[e_i]\, \text{Var}[z \mid e_i]\\
&= \frac{1}{k} \sum_i \text{Var}[z \mid e_i] = \text{Var}[z \mid e_1]
\end{align}
where the last two steps follow from the fact that the distributions among arms are identical.  Furthermore, 
\begin{align*}
\;\;\text{Var}[z \mid e_1] 
&= \text{Var}\left[x_1 \mid e_1\right]\\
&> P\left[ x_1 \geq \sqrt{\log(k)} \wedge \max_{i>1} x_i < \sqrt{\log(k)} \mid e_1\right]  \text{Var}\left[x_1 \mid x_1 \geq \sqrt{\log(k)} \wedge \max_{i>1} x_i < \sqrt{\log(k)} \right] \\
&> \frac{P\left[ x_1 \geq \sqrt{\log(k)} \wedge \max_{i>1} x_i < \sqrt{\log(k)} \right]}{P[e_1]} \text{Var}\left[x_1 \mid x_1 \geq \sqrt{\log(k)} \right] \\
&= k P\left[ x_1 \geq \sqrt{\log(k)} \right] P\left[\max_{i>1} x_i < \sqrt{\log(k)} \right] \text{Var}\left[x_1 \mid x_1 \geq \sqrt{\log(k)} \right ]
\end{align*}
where the last step follows from $P[e_1] = 1/k$ and that $x_1$ is independent from all other draws.

We use the following known result of the Gaussian distribution \cite{hand14}.

\[P[x_1 \geq \sqrt{\log(k)}] > 1/k.\]
Furthermore, 
\begin{align}
P\left[\max_{i>1} x_i < \sqrt{\log(k)} \right] &= \prod_{i=2}^k P\left[x_i < \sqrt{\log(k)} \right]\\
&\geq (1 - C / k)^k \geq 1/e^C
\end{align}
where $C$ is an absolute constant. Finally, by \cite{kmrw18}, 
we have $\text{Var}[x_1 \mid x_1 \geq \sqrt{\log(k)}] \geq \Omega(1/\log(k))$

Putting everything together, we have
\begin{align*}
\text{Var}[z] &\geq \text{Var}[z\mid e_1] \\
              &\geq k (1/k) (1/e^C) \text{Var}[x_1 \mid x_1\geq \sqrt{\log(k)}] \\
              &\geq 1/\log(k)
\end{align*} \qed

\begin{lemm}
	Let an adversary pick $\mu_i \in \R, 1 \leq i \leq k$ and then consider $k$ random draws from a Gaussian distribution $g_i \sim N(0,\sigma^2), 1 \leq i \leq k$. Then the following is true for any adversary,
	\beq
	\text{Var}[g ~|~ g = \underset{g_i:1 \leq i \leq k}{\argmax} ~ g_i + \mu_i] \geq \text{Var}[g ~|~ g = \underset{g_i:1 \leq i \leq k}{\argmax}~ g_i] ~.
	\eeq
	\label{lemm:adversary_gaussian_variance}
\end{lemm}
\proof Without loss of generality assume $\mu_1 \geq \mu_2 \geq \hdots \geq \mu_k$. Also let $g_{(1)} \geq g_{(2)} \geq \hdots g_{(k)}$ denote the order statistics of the Gaussian variables. Now any $\mu_i$ can be mapped to any $g_{(j)}$ to give $k!$ possibilities. Lets divide the $k!$ events into $k$ disjoint sets $A_1,\hdots,A_k$ in the following way. Consider one mapping $\{ (\mu_{i_l},g_{(1)}), \hdots, (\mu_i,g_{(j)}),\hdots, (\mu_{i_h},g_{(k)})\}$ where there are indices $i,j = \underset{1 \leq i,j \leq k}{\argmax} ~ \mu_i + g_{(j)}$. If $j = k$ then we put the mapping in the bin $A_k$. Otherwise assuming $j < k$, let $\mu_{i_1},\hdots,\mu_{i_h}$ be mapped to $g_{(j+1)},\hdots,g_{(j + h)}$ such that $1 \leq j \leq j + h \leq k$. We then find an index $i_m$ such that after swapping $\mu_i$ and $\mu_{i_m}$ such that the new mapping is $\{ (\mu_{i_l},g_{(1)}), \hdots, (\mu_{i_m},g_{(j)}),\hdots,(\mu_i,g_{(j+m)}),\hdots (\mu_{i_h},g_{(k)})\}$  we find that $i,j+m = \underset{1 \leq i_m,j+m \leq k}{\argmax} \mu_{i} +g_{(j+m)}$ but when we swap $\mu_i$ with $\mu_{i_{m+1}}$ for the mapping \\ $\{ (\mu_{i_l},g_{(1)}), \hdots, (\mu_{i_{m+1}},g_{(j)}),\hdots,(\mu_i,g_{(j+m+1)}),\hdots (\mu_{i_h},g_{(k)})\}$ then $\mu_i + g_{j+m+1}$ is no longer the maximum. We then put the mapping $\{ (\mu_{i_l},g_{(1)}), \hdots, (\mu_i,g_{(j)}),\hdots, (\mu_{i_h},g_{(k)})\}$ in the bin $A_{j+m}$. Note that the bin $A_{j+m}$ will also have the mappings $\{ (\mu_{i_l},g_{(1)}), \hdots, (\mu_{i_n},g_{(j)}),\hdots,(\mu_i,g_{(j+n)}),\hdots (\mu_{i_h},g_{(k)})\}$ for all $1 \leq n \leq m$ with $\mu_{i_n}$ swapped with $\mu_i$ where $\mu_i + g_{(j+n)}$ is the maximum as also the mappings $\{ (\mu_{i_l},g_{(1)}), \hdots,(\mu_{i_n},g_{(j-n)}), \hdots, (\mu_{i_n},g_{(j)}),\hdots (\mu_{i_h},g_{(k)})\}$, $1 \leq n \leq j-1$ where $\mu_i + g_{(j-n)}$ is the maximum. Since all these mappings are equally probable, bin $A_{j+m}$ is a set of events out of $k!$ where any $g_{(o)}, 1 \leq o \leq j+m$ are equally probably such that $i,o = \underset{1 \leq o,i \leq k}{\argmax} ~ \mu_i + g_{(o)}$ for some $1 \leq i \leq k$. Moreover from the construction we see that the sets $A_i \cap A_j = \phi, 1 \leq i,j \leq k$ are disjoint and $\cup_{1 \leq i \leq k} A_i$ contains all $k!$ events. Therefore with this construction we make the following observations,
\begin{align}
&\text{Var}[g ~ | ~ g = \underset{g_i:1 \leq i \leq k}{\argmax} ~ g_i + \mu_i] =  \nonumber \\
&=\sum_{i=1}^k \text{Var}(g ~ | g \sim \{ g_{(1)},\hdots, g_{(i)} \} ) P(A_i) ~.
\end{align}

We note that the minimum variance is achieved when $P(A_1) = 1$ and $\text{Var}(g ~ | ~ g \sim {g_{(1)}}) = \text{Var}[g ~|~ g = \underset{g_i:1 \leq i \leq k}{\argmax}~ g_i]$ which is the desired result.  \qed
	
	\section{Proof for Single Parameter Setting with Gaussian Contexts}
	\label{sec:supp_single_param_proofs_gaussian}
	We give the proof for Lemma \ref{lemm:single_param_design_matrix} from the main paper.

\textbf{Lemma \ref{lemm:single_param_regret}}
\textit{
	Denote by $\beta = \underset{\substack{{1 \leq i \leq k, 1 \leq t \leq T} \\  {v \in A}}}{\max} \langle x_i^t,v \rangle$, where $A = \text{cone}(E_c) \cap S^{p-1}$ is the error set.
	Assume $T > t_{\min}$, where $t_{\min}$ depends on properties of the true parameter $\theta^*$ and the regularizer $R(\cdot)$. Then,
	\beq
	\text{Reg}(T) \leq 4 \beta t_{\min} + \sum_{e = \lceil \log t_{\min} \rceil}^{\lfloor \log T \rfloor}  2\beta T_e \|\hat{\theta}^{(e)} - \theta^*\|_2 ~.
	\eeq
}

\proof Let the episodes be indexed by $e$, let $T_e$ denote the number of rounds in episode $e$ and let $T$ denote the total number of rounds. Let $S_e$ denote the rounds in episode $e$. If context $i^t$ is selected in round $t$ and $i^*$ denotes the optimal context then the regret can be computed as follows,
\begin{align}
\text{Reg}(T) &= \sum_{t=1}^T  \langle \theta^*,x_{i^*}^t - x_{i^t}^t \rangle \nonumber \\
&= \sum_{e=1}^{\lfloor \log T \rfloor} \sum_{t \in S_e} \langle \theta^*,x_{i^*}^t - x_{i^t}^t \rangle \nonumber \\
&= \sum_{e=1}^{\lceil \log t_{\min} \rceil} \sum_{t \in S_e} \langle \theta^*,x_{i^*}^t - x_{i^t}^t \rangle + \sum_{e = \lceil \log t_{\min} \rceil}^{\lfloor \log T \rfloor} \sum_{t \in S_e} \langle \theta^*,x_{i^*}^t - x_{i^t}^t \rangle ~.
\label{eq:eq_single_param_regret1}
\end{align}
The first term on the r.h.s. of \myref{eq:eq_single_param_regret1} can be upper bounded as follows,
\begin{align}
\sum_{e=1}^{\lceil \log t_{\min} \rceil} \sum_{t \in S_e}  \langle \theta^*,x_{i^*}^t - x_{i^t}^t \rangle &\leq \sum_{e=1}^{\lceil \log t_{\min} \rceil} \sum_{t \in S_e} | \langle \theta^*,x_{i^*}^t \rangle | + | \langle \theta^*,x_{i^t}^t \rangle | \nonumber \\
&\leq \sum_{e=1}^{\lceil \log t_{\min} \rceil} \sum_{t \in S_e} 2 \beta \nonumber \\
&\leq 4 \beta t_{\min} ~,
\label{eq:eq_single_param_regret2}
\end{align}
where in the third line we use from the algorithm $T_e = 2^{e-1}$ and hence $\sum_{e=1}^{\lceil \log t_{\min} \rceil} \sum_{t=1}^{T_e} 1 \leq 2 t_{\min}$.

We make the following observations to bound the second term on the r.h.s. in equation \myref{eq:eq_single_param_regret1},
\begin{align}
\sum_{e = \lceil \log t_{\min} \rceil}^{\lfloor \log T \rfloor} \sum_{t \in S_e} \langle \theta^*,x_{i^*}^t - x_{i^t}^t \rangle
&= \sum_{e = \lceil \log t_{\min} \rceil}^{\lfloor \log T \rfloor} \sum_{t \in S_e} \langle \theta^* - \hat{\theta}^{(e)}, x_{i^*}^t \rangle - \langle \theta^* - \hat{\theta}^{(e)},x_i^t \rangle + \langle \hat{\theta}^{(e)},x_{i^*}^t - x_i^t \rangle \nonumber \\
&\leq \sum_{e = \lceil \log t_{\min} \rceil}^{\lfloor \log T \rfloor} \sum_{t \in S_e} |\langle \theta^* - \hat{\theta}^{(e)}, x_{i^*}^t \rangle | + |\langle \theta^* - \hat{\theta}^{(e)},x_i^t \rangle | \nonumber \\
&\leq \sum_{e = \lceil \log t_{\min} \rceil}^{\lfloor \log T \rfloor} \sum_{t \in S_e} 2 \beta \|\theta^* - \hat{\theta}^{(e)}\|_2 ~,
\label{eq:eq_single_param_regret3}
\end{align}
where in the second line we use $\langle \hat{\theta}^e,x_{i^*}^t \rangle \leq \langle \hat{\theta}^e,x_i^t \rangle$ as $x_{i^t}^t$ was chosen ahead of $x_{i^*}^t$ in round $t$.

The stated result now follows from \myref{eq:eq_single_param_regret1}, \myref{eq:eq_single_param_regret2} and \myref{eq:eq_single_param_regret3}.  \qed

We give the proof for Lemma \ref{lemm:single_param_design_matrix} from the main paper.

\textbf{Lemma \ref{lemm:single_param_design_matrix}}
\textit{
	The rows of the design matrix $Z^{(e)} \in \R^{T_e \times p}$ in any episode $e$ satisfy $\kappa_z = \|z^t\|_{\psi_2} \leq c_2 \sigma \sqrt{\log k}$ for $c_2$ some positive constant.  Moreover the minimum eigenvalue of the matrix $E_{z^t}[z^t (z^T)^T]$ satisfies,
	\beq
	\lambda_{\min} (E_{z^t}[z^t (z^t)^\intercal]) \geq c_1 \frac{\sigma^2}{\log k} ~,
	\eeq
	where $c_1$ is some positive constant and the expectation is over the random draws of contexts.
}

\proof 
The rows of the design matrix satisfy,
\beq
    z^t = \underset{x_i^t:1 \leq i \leq k}{\argmax} \langle x_i^t,\hat{\theta}^{(e)} \rangle ~,
\eeq
where $\hat{\theta^{(e)}}$ is the estimated parametrer in episode $e$. We first prove the result on the sub-Gaussian norm of $z^t$. Let $Q$ be an orthogonal matrix such that $Q \hat{\theta^{(e)}} = (\|\hat{\theta}^{(e)}\|_2,0,\hdots,0)$. Also for any round $t$, let $(x_1^t,\hdots,x_k^t) = (Q^{\intercal}\epsilon_1^t,\hdots,Q^{\intercal}\epsilon_k^t)$. Due to rotational invariance $\epsilon_i^t \sim N(0,\sigma^2 \I_{p \times p}), 1 \leq i \leq k$. Therefore,
\begin{align}
  &  z^t = \underset{x_i^t:1 \leq i \leq k}{\argmax} \langle x_i^t,\hat{\theta^{(e)}} \rangle \nonumber \\
  & Qz^t = \epsilon^t = \underset{\epsilon_i^t:1 \leq i \leq k}{\argmax} \langle \epsilon_i^t,Q\hat{\theta^{(e)}} \rangle 
\end{align}
Therefore $\epsilon^t \in \R^p$ is a $p$-dimensional random vector such that elements $(\epsilon^t)_j, 1 \leq j \leq p$ are random $N(0,\sigma^2)$ elements with $\|(\epsilon^t)_j\|_{\psi_2} \leq c_3 \sigma$ for some constant $c_3$. For the element at the first position,
\beq
    (\epsilon^t)_1 = \underset{1 \leq i \leq k}{\argmax} (\epsilon_i^t)_1 ~,
\eeq
where $(\epsilon_i^t)_1$ are $N(0,\sigma^2)$ elements. The following Lemma bounds the sub-Gaussian norm of $(\epsilon_i^t)_1$:
\begin{lemm}
    Let $g_1,\cdots,\g_k$ be $k$ Gaussian $N(0,\sigma^2)$ elements and let $h = \underset{1 \leq i \leq k}{\argmax} ~ g_i$. Then the sub-Gaussian norm of $h$ satisfies the following:
    \beq
        \|h\|_{\psi_2} \leq c_6 \sigma \sqrt{\log k} ~.
    \eeq
\end{lemm}
\proof The maximum of $k$-Gaussian elements can be expressed as follows with vector $g = [g_1,\hdots,g_k] \in \R^k$:
\beq
    \|g\|_{\infty} = \sup\limits_{u:\|u\|_1 \leq 1} \langle g,u \rangle ~.
\eeq
Therefore,
\beq
    E \left [\sup\limits_{1 \leq i \leq k} g_i \right ] = E \left [ \sup\limits_{u:\|u\|_1 \leq 1} \langle g,u \rangle \right ] \leq c_4 \sigma \sqrt{\log k} ~,
\eeq
where the last inequality is because the Gaussian width of the unit $\ell_1$ norm ball is $\sqrt{\log k}$ \cite{tala05,tala14,chba15} and by the majorizing measure theorem (see Theorem 2.1.1 in \cite{tala05}). Now from the result of Lemma 2.1.3 in \cite{tala05},
\beq
    P(|\sup\limits_{1 \leq i \leq k} \g_i - E \sup\limits_{1 \leq i \leq k} \g_i | \geq u) \leq 2 \exp \left ( -\frac{u^2}{2\sigma^2} \right ) ~.
\eeq
Note that any random variable $\xi$ is a sub-Gaussian random variable with sub-Gaussian norm $c_5K$ is it satisfies the following tail decay \cite{vers12},
\beq
    P(|\xi| \geq u) \leq 2 \exp \left ( - \frac{u^2}{2K^2} \right ) ~.
\eeq
Therefore $\left ( \sup\limits_{1 \leq i \leq k} \g_i - E \sup\limits_{1 \leq i \leq k} \g_i \right )$ is a $c_5\sigma$-sub-Gaussian random variable. Therefore,
\begin{align}
    \|h\|_{\psi_2} &= \|\sup\limits_{1 \leq i \leq k} g_i - E \sup\limits_{1 \leq i \leq k} g_i + E \sup\limits_{1 \leq i \leq k} g_i\|_{\psi_2} \nonumber \\ &\leq \|\sup\limits_{1 \leq i \leq k} g_i - E \sup\limits_{1 \leq i \leq k} g_i\|_{\psi_2} + \|E \sup\limits_{1 \leq i \leq k} g_i\|_{\psi_2} \nonumber \\
    &\leq c_5\sigma + c_4 \sigma \sqrt{\log k} \nonumber \\
    &\leq c_6 \sigma \sqrt{\log k} ~.
    \label{eq:ch6_center_subg_norm}
\end{align}

Therefore by the definition of sub-Gaussian random variables $(\epsilon^t)_1$ is a sub-Gaussian random variable with $\|(\epsilon^t)_1\|_{\psi_2} \leq c_6 \sigma \sqrt{\log k} $ for some constant $c_6$. Therefore $Qz^t$ is a random vector with independent sub-Gaussian random elements. Therefore from the result of Lemma \ref{lemm:rotsubGau} the elements of $z^t = Q^T Q z^t$ are also independent sub-Gaussian random variables with sub-Gaussian norm of each element  $\|(z^t)_i\|_{\psi_2} \leq c_7 \sigma \sqrt{\log k}$. Also from the result of Lemma \ref{lemm:prosubg}, $z^t$ is a sub-Gaussian random vector with $\|z^t\|_{\psi_2} \leq c_2 \sigma \sqrt{\log k}$ for some constant $c_2$ which proves the first result.


In order to prove the minimum eigenvalue condition, let $Q$ be an orthogonal matrix such that $Q \hat{\theta}^{(e)} = (\|\hat{\theta}^{(e)}\|_2,0,\hdots,0)$ as outlined earlier. Again for any round $t$, let $(x_1^t,\hdots,x_k^t) = (Q^{\intercal}\epsilon_1^t,\hdots,Q^{\intercal}\epsilon_k^t)$. Due to rotational invariance $\epsilon_i^t \sim N(0,\sigma^2 \I_{p \times p}), 1 \leq i \leq k$. Now with $z^t = \underset{x_i^t:1 \leq i \leq k}{\argmax} \langle x_i^t,\hat{\theta} \rangle = \underset{x_i^t:1 \leq i \leq k}{\argmax} \langle Qx_i^t,Q\hat{\theta} \rangle$ and let $\epsilon^t = Q z^t$ 
\begin{align}
\lambda_{\min} \left ( E \left[ z^t (z^t)^{\intercal} \suchthat z^t = \underset{x_i^t:1 \leq i \leq k}{\argmax} \langle x_i^t,\hat{\theta}^{(e)} \rangle \right ] \right )
&= \min\limits_{w:\|w\|_2 = 1} w^{\intercal} \left (E \left [z^t(z^t)^{\intercal} \suchthat z^t = \underset{x_i^t:1 \leq i \leq k}{\argmax} \langle x_i^t,\hat{\theta}^{(e)} \rangle \right ] \right ) w \nonumber \\
&= \min\limits_{w:\|w\|_2 = 1} \left (E \left[ w^{\intercal}z^t(z^t)^{\intercal} w \suchthat  z^t = \underset{x_i^t:1 \leq i \leq k}{\argmax} \langle x_i^t,\hat{\theta}^{(e)} \rangle \right ] \right )  \nonumber \\
&= \min\limits_{w:\|w\|_2 = 1} \left (E \left [\langle w,z^t \rangle^2 \suchthat  z^t = \underset{x_i^t:1 \leq i \leq k}{\argmax} \langle x_i^t,\hat{\theta}^{(e)} \rangle \right ] \right )  \nonumber \\
&\geq \min\limits_{w:\|w\|_2 = 1} \text{Var} \left (\langle w, z^t \rangle \suchthat  z^t = \underset{x_i^t:1 \leq i \leq k}{\argmax} \langle x_i^t,\hat{\theta}^{(e)} \rangle \right ) \nonumber \\
&= \min\limits_{w:\|w\|_2 = 1} \text{Var}\left (\langle Qw,Qz^t \rangle \suchthat  z^t = \underset{x_i^t:1 \leq i \leq k}{\argmax} \langle Qx_i^t,Q\hat{\theta}^{(e)} \rangle \right ) \nonumber \\
&= \min\limits_{w:\|w\|_2 = 1} \text{Var} \left (\langle Qw,Qz^t \rangle \suchthat z_t = \underset{x_i^t:1 \leq i \leq k}{\argmax} (Qx^t)_1 \|\hat{\theta}^{(e)}\|_2\right ) \nonumber \\
&= \min\limits_{w:\|w\|_2 = 1} \text{Var}\left ( \langle w,Qz^t \rangle \suchthat z_t = \underset{x_i^t:1 \leq i \leq k}{\argmax} (Qx^t)_1 \|\hat{\theta}^{(e)}\|_2 \right ) ~,
\end{align}
where the last line uses that minimizing over $w$ and over $Qw$ yield the same result. Now 
$\epsilon^t = Qz^t$ is a $N(0,\sigma^2 \I_{p \times p})$ random vector. Therefore,
\begin{align}
\lambda_{\min} \left( E \left[ z^t (z^t)^{\intercal} \suchthat z^t = \underset{x_i^t:1 \leq i \leq k}{\argmax} \langle x_i^t,\hat{\theta}^{(e)} \rangle \right] \right ) &\geq \min\limits_{w:\|w\|_2=1} \left ( \text{Var} \left [\langle w,\epsilon^t \rangle \suchthat \epsilon^t = \underset{\epsilon_i^t:1 \leq i \leq k}{\argmax} (\epsilon_{i}^t)_1 \|\hat{\theta}^{(e)}\| \right ]  \right ) \nonumber \\
&\geq \min\limits_{w:\|w\|_2 = 1} \left ( w_1^2 \text{Var}((\epsilon^t)_1) \suchthat  \epsilon^t = \underset{\epsilon_i^t:1 \leq i \leq k}{\argmax}  (\epsilon_{i}^t)_1 \|\hat{\theta}^{(e)}\|) \right ) \\ 
&+ \left ( \sum_{j=2}^p w_j^2 \text{Var}((\epsilon^t)_j)  \suchthat  \epsilon^t = \underset{\epsilon_i^t:1 \leq i \leq k}{\argmax}  (\epsilon_{i}^t)_1 \|\hat{\theta}^{(e)}\|) \right ) \nonumber \\
&\geq c_1 \frac{\sigma^2}{\log k} ~,
\end{align}
where second line 
follows as the coordinates of $\epsilon^t$ are independent and the third line follows as from the result of Lemma \ref{lemm:max_gauss_variance} where $\text{Var}((\epsilon^t)_1 | \epsilon^t = \underset{\epsilon_i^t:1 \leq i \leq k}{\argmax} (\epsilon_{i}^t)_1 \|\hat{\theta}^{(e)}\| ) \geq c_1 \frac{\sigma^2}{\log k}$ and $\text{Var}((\epsilon^t)_j | \epsilon^t = \underset{\epsilon_i^t:1 \leq i \leq k}{\argmax} (\epsilon_{i}^t)_1 \|\hat{\theta}^{(e)}\| ) = \sigma^2$. \qed



We give the proof for the estimation error in each episode for the Gaussian contexts setting.
\begin{theo}
	Let  $T_e \geq c_7 (w(A) + \sqrt{\log \log T} + \sqrt{\log (1/\delta)})^2\log^2 k$. Then with probability atleast $1 - \delta \exp(-\eta_2 w^2(A)) - \delta$,
	\beq
	\|\hat{\theta}^{(e+1)} - \theta^* \|_2 \leq O \left (\frac{ \gamma }{\sigma \sqrt{T_e}} \right ) ~,
	\eeq
	where $\gamma = c \kappa_{\omega} \sqrt{\log k} (w(A) + \sqrt{\log \log T} + \sqrt{\log (1/\delta)})$, $E_c = \{ \Delta ~ | ~ R(\theta^* + \Delta) \leq R(\theta^*)  \}$, $A = \text{cone}(E_c) \cap S^{p-1}$ is the error set, $w(\cdot)$ denotes the Gaussian width of a set.
	\label{thm:ch6_single_param_gaussian_est_error}
\end{theo}

\proof Consider parameter estimation at the beginning of episode $e+1$. Assume the design matrix has the SVD decomposition $\frac{1}{\sqrt{T_e}}Z^{(e)} = U D V^{\intercal}$ where $U \in \R^{T_e \times d}$, $D \in \R^{d \times d}$ and $V \in \R^{p \times d}$, where $d$ is the rank of $Z^{(e)}$. Also let $\Sigma^{1/2} = V D V^{\intercal}$. Define the Puffer transformation $F = U D^{-1} U^{\intercal}$ \cite{jiro15} and consider the preconditioned design matrix $\tilde{Z}^{(e)} = FZ^{(e)}$ and response $\tilde{y}^{(e)} = Fy^{(e)}$. Since $y^{(e)} = Z^{(e)}\theta^* + \omega^{(e)}$, it follows that $Fy^{(e)} = FZ^{(e)}\theta^* + F\omega^{(e)}$, i.e. $\tilde{y}^{(e)} = \tilde{Z}^{(e)}\theta^* + \tilde{\omega}^{(e)}$ where $\tilde{\omega}^{(e)} = F\omega^{(e)}$   We then compute the constrained regression estimator $\hat{\theta}^{(e)} = \underset{\theta \in \R^p}{\argmin} \frac{1}{2T_e} \|\tilde{y}^{(e)} - \tilde{Z}^{(e)} \theta \|_2^2 \quad \text{s.t.} \quad R(\theta) \leq R(\theta^*)$. Since $\hat{\theta}^{(e)}$ minimizes the loss function the following observation is straightforward,
\beq
\frac{1}{2T_e} \|\tilde{y}^{(e)} - \tilde{Z}^{(e)}\hat{\theta}^{(e)}\|_2^2 - \frac{1}{2T_e} \|\tilde{y}^{(e)} - \tilde{Z}^{(e)}\theta^* \|_2^2 \leq 0
\label{eq:estimation_err_bnds}
\eeq
Let $\hat{\theta}^{(e)} = \theta^* + \Delta^{(e)}$ where $\Delta^{(e)}$ satisfies $R(\theta^* + \Delta^{(e)}) \leq R(\theta^*)$. Substituting it in \myref{eq:estimation_err_bnds} and subsequent simplification using $u = \frac{\Delta^{(e)}}{\|\Delta^{(e)}\|_2}$ yields the following,
\begin{align}
\frac{1}{2T_e} \|\tilde{Z}^{(e)} \Delta^{(e)}\|_2^2 = \frac{1}{2T_e} \|\tilde{Z}^{(e)} u \|_2^2 \|\Delta^{(e)}\|_2^2
&\leq \frac{1}{T_e} \left \langle \tilde{y}^{(e)} - \tilde{Z}^{(e)} \theta^*, \tilde{Z}^{(e)} \Delta^{(e)} \right \rangle \nonumber \\
&\leq \frac{1}{T_e} \left \langle (\tilde{Z}^{(e)})^{
	\intercal} \tilde{\omega}^{(e)},\Delta^{(e)} \right \rangle \nonumber \\
&\leq \frac{1}{\sqrt{T_e}} \left \langle \frac{1}{\sqrt{T_e}} \Sigma^{1/2} (\tilde{Z}^{(e)})^{\intercal} \tilde{\omega}^{(e)}, \Sigma^{-1/2} \Delta^{(e)} \right \rangle \nonumber \\
&\leq \frac{1}{\sqrt{T_e}} \left \langle \frac{1}{\sqrt{T_e}} \Sigma^{1/2} (FZ^{(e)})^{\intercal} F \omega^{(e)}, \Sigma^{-1/2} \Delta^{(e)} \right \rangle \nonumber \\
&\leq \frac{1}{\sqrt{T_e}} \left \langle V D V^{\intercal} V D U^{\intercal} U D^{-1} U^{\intercal} U D^{-1} U^{\intercal} \omega^{(e)}, \Sigma^{-1/2} \Delta^{(e)} \right \rangle \nonumber \\
&\leq \frac{1}{\sqrt{T_e}} \left \langle V U^{\intercal} \omega^{(e)}, \Sigma^{-1/2} \Delta^{(e)} \right \rangle \nonumber \\
&\leq \frac{1}{\sqrt{T_e}} \langle h, \Sigma^{-1/2} \Delta^{(e)} \rangle \nonumber \\
&\leq \frac{1}{\sqrt{T_e}} \langle h, \Sigma^{-1/2}u \rangle \|\Delta^{(e)}\|_2 ~,
\label{eq:single_param_gauss_est_err_eq13}
\end{align}
where in the fourth line we use $\tilde{Z}^{(e)} = FZ^{(e)}$, $\tilde{\omega}^{(e)} = F\omega^{(e)}$; in the fifth line we use that $\Sigma^{1/2} = V D V^{\intercal}$, $\frac{1}{\sqrt{T_e}}Z^{(e)} = U D V^{\intercal}$ and $F^{(e)} = U D^{-1} U^{\intercal}$. In the second last line we observe that $h \in \R^p$ is a sub-Gaussian random vector with $\|h\|_{\psi_2} \leq c_3 \kappa_{\omega}$. This is because applying results from Lemma \ref{lemm:rotsubGau} twice it can be inferred that $U^{\intercal} \omega^{(e)} \in \R^d$ is sub-Gaussian with $\|U^{\intercal} \omega^{(e)}\|_{\psi_2} \leq c_4 \kappa_{\omega}$ and $h = V U^{\intercal} \omega^{(e)} \in \R^p$ is sub-Gaussian with $\|h\|_{\psi_2} \leq c_3 \kappa_{\omega}$.

\vspace{3mm}

{\bf 1. Minimum eigenvalue condition: Lower bounds for $\inf\limits_{u \in A} \frac{1}{T_e} \|Z^{(e)}u\|_2^2$}

We obtain high probability lower bounds on the quantity $\inf\limits_{u \in A} \frac{1}{T_e} \|Z^{(e)}u\|_2^2$. Remember that $Z^{(e)} \in \R^{T_e \times p}$ is the design matrix before the Puffer transformation. We make the following observations:
\begin{align}
    \frac{1}{T_e} \|Z^{(e)}u\|_2^2
    &= \frac{1}{T_e} \sum_{t=1}^{T_e} \langle z^t,u \rangle^2 \nonumber \\
    &= \frac{1}{T_e} \sum_{t=1}^{T_e} \langle z^t - E[z^t] + E[z^t],u \rangle^2 \nonumber \\
    &= \frac{1}{T_e} \sum_{t=1}^{T_e} \langle z^t - E[z^t],u \rangle^2 + \frac{1}{T_e} \sum_{t=1}^{T_e} \langle E[z^t],u \rangle^2
    - \frac{2}{T_e} \sum_{t=1}^{T_e} \langle z^t - E[z^t],u \rangle \langle E[z^t],u \rangle 
    \label{eq:single_param_gauss_est_err_eq101}
\end{align}
We first analyze the quantity $\frac{1}{T_e} \sum_{t=1}^{T_e} \langle z^t - E[z^t],u \rangle^2$. Let $G \in \R^{T_e \times p}$ be the design matrix with rows as $z^t - E[z^t]$. Using the results of Lemma 2 and the episodic algorithm, we make the observation that the rows of the matrix $G$ are i.i.d. $\sigma$-sub-Gaussian. We want lower bounds on the quantity $\frac{1}{T_e} \|Gu\|_2^2$. 
We use the following result \cite{bcfs14,mept07}.

\begin{theo}[Mendelson, Pajor, Tomczak-Jaegermann~\cite{mept07}]
	There exist absolute constants $c_2$, $c_3$, $c_4$ for which the following
	holds. Let $(\Omega,\mu)$ be a probability space, set $F$ be a subset of the unit
	sphere of $L_2(\mu)$, i.e., $F \subseteq S_{L_2} = \{ f : \|f\|_{L_2} = 1\}$, and assume that $ \sup_{f \in F}~\|f\|_{\psi_2} \leq \kappa$. Then, for any $\theta > 0$ and $n \geq 1$ satisfying
	\beq
	c_2 \kappa \gamma_2(F, \|\cdot\|_{\psi_2}) \leq \theta \sqrt{n}~,
	\eeq
	with probability at least $1- \exp(-c_3 \theta^2 n/\kappa^4)$,
	\beq
	\sup_{f \in F}~\left| \frac{1}{n} \sum_{i=1}^n f^2(X_i) - E\left[f^2\right] \right| \leq \theta~.
	\eeq
	Further, if $F$ is symmetric, then
	\beq
	E\left[ \sup_{f \in F} ~\left| \frac{1}{n} \sum_{i=1}^n f^2(X_i) - E\left[f^2\right] \right| \right] \leq c_4 \max \left\{ 2\kappa \frac{\gamma_2(F, \|\cdot\|_{\psi_2})}{\sqrt{n}}, \frac{\gamma_2^2(F, \|\cdot\|_{\psi_2})}{n} \right\}
	\eeq
	\label{thm:subGREmain}
\end{theo}

For convenience let $z_0$ have the same distribution as the rows of the design matrix $G$. Consider the following class of functions:
\beq
F = \{ f_u,u \in A : f_u(\cdot) = \frac{1}{\sqrt{E [\langle \cdot,u \rangle^2 ]} } \langle \cdot,u \rangle \} ~.
\eeq
Then, $f_u(z_0) = \frac{1}{\sqrt{E [\langle z_0,u \rangle^2]}} \langle z_0,u \rangle$ and $F$ is a subset of the unit sphere, i.e., $F \subseteq S_{L_2}$, since $\|f\|_{L_2} = E[f_u^2] = 1$.

Next, we get an upper bound on $\sup\limits_{f_u \in F} \|f_u\|_{\psi_2} = \sup\limits_{u \in A} \left \|\frac{1}{\sqrt{E[\langle z_0,u \rangle^2]}} \langle z_0,u \rangle \right \|_{\psi_2}$. Note that $\kappa_z = \|z_0\|_{\psi_2} = \sup\limits_{v \in S^{p-1}} \|\langle z_0,v \rangle \|_{\psi_2} \leq c_2 \sigma$ (see arguments before equation \myref{eq:ch6_center_subg_norm}). Also from the result of Lemma 2, $E[\langle z_0,u \rangle^2] \geq \frac{\sigma^2}{\log k}$
Therefore,
\begin{align}
\sup\limits_{f_u \in F} \|f_u\|_{\psi_2} &= \sup\limits_{u \in A} \left \|\frac{1}{\sqrt{E[\langle z_0,u \rangle^2]}} \langle z_0,u \rangle \right \|_{\psi_2} \\
&\leq \frac{c_2 \sigma \sqrt{\log k}}{c_3 \sigma} \\
&\leq c_4 \sqrt{\log k} ~.
\end{align}
As a result we have,
\beq
\gamma_2(F \cap S_{L_2}, \|\cdot\|_{\psi_2}) \leq c_4  \gamma_2(F \cap S_{L_2}, \|\cdot\|_{L_2}) \leq c_4 c_5 w(A) \sqrt{\log k} ~,
\eeq
where the last line follows from generic chaining \cite{tala14,tala05}, for some constant $c_5 > 0$. Therefore, in the context of Theorem \ref{thm:subGREmain}, we choose,
\begin{align}
\theta &=  c_4^2 \frac{(c_6 c_5 w(A) + \sqrt{\log (1/\delta)} + \sqrt{\log \log T}) \log k}{\sqrt{T_e}} \nonumber \\
&\geq c_6 c_4 \sqrt{\log k}  \frac{\gamma_2(F \cap S_{L_2}, \|\cdot\|_{\psi_2})}{\sqrt{T_e}} ~,
\end{align}
for some constant $0 < \delta < 1$, so that the condition on $\theta$ is satisfied. With this choice of $\theta$, we have,
\begin{align}
\frac{\theta^2 T_e}{c_4^4 \log^2 k} &\geq c_6^2 c_5^2 w^2(A) + \log \log T + \log(1/\delta) \nonumber \\
&= \eta_2 w^2(A) + \log \log T + \log(1/\delta) ~.
\end{align}
Then, from Theorem \ref{thm:subGREmain} it follows that with probability atleast $1 -  \exp(-\eta_2 w^2(A) - \log \log T - \log (1/\delta) )$ with $z^t, 1 \leq t \leq T_e$ denoting the rows of $Z^e$ , we have,
\begin{align*}
&\sup\limits_{u \in A} \left | \frac{1}{T_e} \frac{1}{E[\langle z^t - E[z^t],u \rangle^2]} \sum_{t \in [T_e]} \langle z^t - E[z^t],u \rangle^2 - 1 \right | \leq c_4^2 \frac{( c_6 c_5 w(A) + \sqrt{\log \log T} + \sqrt{\log(1/\delta)}) \log k}{\sqrt{T_e}} \\
\Rightarrow &\inf\limits_{u \in A} \frac{1}{T_e} \|G u\|_2^2 \geq E[\langle z^t - E[z^t],u \rangle^2] \left ( 1 - c_4^2 \frac{( c_6 c_5 w(A) + \sqrt{\log \log T} + \sqrt{\log(1/\delta)}) \log k}{\sqrt{T_e}} \right ) ~.
\end{align*}
Substituting $T_e \geq c_7 (w(A) + \sqrt{\log \log T} + \sqrt{\log(1/\delta)})^2 \log^2 k$ so that $1 - c_4^2 \frac{(c_6 c_5w(A) + \sqrt{\log \log T} + \sqrt{\log(1/\delta)}) \log k}{\sqrt{T_e}} \geq c_9$ and noting from Lemma 2 that $E[\langle z^t - E[z^t],u \rangle^2] = \text{Var}[\langle z^t,u \rangle]) \geq c_3 \frac{\sigma^2}{\log k}$ it follows with probability atleast $1 -  \exp(-\eta_2 w^2(A) - \log \log T - \sqrt{\log(1/\delta)})$,
\beq
\inf\limits_{u \in A} \frac{1}{T_e} \|G u \|_2^2 \geq c \frac{\sigma^2}{\log k} ~.
\eeq
Now by a union bound argument for all episodes $e \leq \lfloor \log T \rfloor$ with probability atleast $1 -  \exp(-\eta_2 w^2(A) - \log(1/\delta)) = 1 - \delta \exp(-\eta_2 w^2(A))$,
\beq
\inf\limits_{e} \inf\limits_{u \in A} \frac{1}{T_e} \|G u\|_2^2 = \inf\limits_{e} \inf\limits_{u \in A} \frac{1}{T_e} \sum_{t=1}^{T_e} \langle z^t - E[z^t],u \rangle^2 \geq c \frac{\sigma^2}{\log k} ~.
\label{eq:single_param_gauss_est_err_eq102}
\eeq
We now derive upper bounds for the quantity $\frac{1}{\sqrt{T_e}} \sum_{t=1}^{T_e} \langle z^t - E[z^t],u \rangle \langle E[z^t],u \rangle$. Let $\alpha \in \R^{T_e}$ be the vector whose elements $\alpha_i = \frac{1}{\sqrt{T_e}} \langle E[z^i],u \rangle$ and therefore 
$\|\alpha\|_2 = \frac{1}{\sqrt{T_e}} \sqrt{\sum_{t=1}^{T_e} \langle E[z^t],u \rangle^2}$. Note that it follows from Lemma \ref{lemm:rotsubGau} that $\langle z^t - E[z^t],u \rangle$  is a $c_2 \sigma$-sub-Gaussian random variables, i.e., $\|\langle z^t - E[z^t],u \rangle \|_{\psi_2} \leq c_2 \sigma$. Therefore from the Hoeffding inequality of Lemma \ref{lemm:hoeffding}:
\beq
    P \left ( \left | \sum_{t=1}^{T_e} \alpha_t \langle z^t - E[z^t],u \rangle \right | \geq \tau  \right ) \leq 2 \exp \left ( - \frac{\tau^2}{c_3 \sigma^2 \|\alpha\|_2^2} \right ) ~.
\eeq
Now for any $u,v \in A$, $\langle z^t - E[z^t], u - v \rangle$ is a $c_2 \sigma \|u-v\|_2$-sub-Gaussian random variable. Therefore by an application of Lemma \ref{lemm:hoeffding}:
\beq
    P \left ( \left | \sum_{t=1}^{T_e} \alpha_t \langle z^t - E[z^t],u - v \rangle \right | \geq \tau  \right ) \leq 2 \exp \left ( - \frac{\tau^2}{c_3 \sigma^2 \|u-v\|_2^2 \|\alpha\|_2^2} \right ) ~.
\eeq
Therefore substituting $\sigma_1 = \sqrt{c_3}\sigma \|\alpha\|_2$, we get,
\beq
    P \left ( \left | \sum_{t=1}^{T_e} \alpha_t \langle z^t - E[z^t],u - v \rangle \right | \geq \tau  \right ) \leq 2 \exp \left ( - \frac{\tau^2}{\sigma_1^2 \|u-v\|_2^2} \right ) ~.
\eeq
Therefore by the definition of the Gaussian width,
\beq
    E \left [ \sup\limits_{u \in A} \left | \sum_{t=1}^{T_e} \alpha_t \langle z^t - E[z^t],u \rangle \right |  \right ] \leq c_4 \sigma_1 w(A) = c_5 \sigma \|\alpha\|_2 w(A) ~.
    \label{eq:single_param_gaussian_est_err_eq201}
\eeq
Now for the high probability bounds we refer Theorem 2.2.27 in \cite{tala14}. Applying the result of Theorem 2.2.27 \cite{tala14} leads to the following result :
\beq
    P\left (\sup\limits_{u \in A} \left | \sum_{t=1}^{T_e} \alpha^t \langle z^t - E[z^t],u \rangle \right | \geq E \left [\sup\limits_{u \in A} \left | \sum_{t=1}^{T_e} \alpha^t \langle z^t - E[z^t],u \rangle \right | \right ] + c_6 \sigma_1 \tau  \right ) \leq c_7 \cdot \exp ( -\tau^2) ~.
\eeq
Let $\tau = c_8 (\sqrt{\log(1/\delta)} + \sqrt{\log \log T})$ choosing $c_8$ large enough so that $c_7 \cdot \exp(-\tau^2) \geq c_7 \cdot \exp ( -c_8^2 (\log \log T + \log(1/\delta))) \geq \exp (- \log \log T - \log(1/\delta))$. Also substituting the value of $E \left [\sup\limits_{u \in A} \left | \sum_{t=1}^{T_e}  \alpha^t \langle g^t - E[g^t],u \rangle  \right | \right ]$ from equation \myref{eq:single_param_gaussian_est_err_eq201} and choosing constant $c_9$ large enough, we get the following:
\beq
    P \left ( \sup\limits_{u \in A} \left | \sum_{t=1}^{T_e} \alpha^t \langle z^t - E[z^t],u \rangle \right | \geq c_9 \sigma \|\alpha\|_2 (w(A) + \sqrt{\log(1/\delta)} + \sqrt{\log \log T}) \right ) \leq \exp \left ( - \log (1/\delta) - \log \log T \right ) ~.
\eeq
This above is true for any single episode $e$. Taking a union bound over all $\lfloor \log \log T \rfloor$ episodes, we get:
\begin{align}
    P \left ( \sup\limits_{u \in A} \left | \sum_{t=1}^{T_e} \alpha^t \langle z^t - E[z^t],u \rangle \right | \geq c_8 \sigma \|\alpha\|_2 (w(A) + \sqrt{\log(1/\delta)} + \sqrt{\log \log T}) \right ) 
    &\leq \exp \left ( - \log (1/\delta) - \log \log T + \log \log T \right ) \nonumber \\ 
    &= \delta ~.
    \label{eq:single_param_gauss_est_err_eq103}
\end{align}
Now from equations \myref{eq:single_param_gauss_est_err_eq101}, \myref{eq:single_param_gauss_est_err_eq102} and \myref{eq:single_param_gauss_est_err_eq103} we get,
\beq
    \frac{1}{T_e} \|Z^{(e)}u\|_2^2 \geq c_3 \frac{\sigma^2}{\log k} + \|\alpha\|_2^2 -
    \frac{2c_9 \sigma \|\alpha\|_2(w(A) + \sqrt{\log(1/\delta)} + \sqrt{\log \log T})}{\sqrt{T_e}} ~.
    \label{eq:single_param_gauss_est_err_eq104}
\eeq
Equation \myref{eq:single_param_gauss_est_err_eq104} is minimized when $\|\alpha\|_2 = \frac{c_9 \sigma \|\alpha\|_2(w(A) + \sqrt{\log(1/\delta)} + \sqrt{\log \log T})}{\sqrt{T_e}}$. Substituting the minimum value in equation \myref{eq:single_param_gauss_est_err_eq104} and by simple algebraic manipulations we get:
\beq
    \frac{1}{T_e} \|Z^{(e)}u\|_2^2 \geq \frac{\sigma^2}{\log k} \left ( c_3 - \frac{c_9^2 (w(A) + \sqrt{\log(1/\delta)} + \sqrt{\log \log T})^2 \log k}{T_e} \right )
    \label{eq:single_param_gauss_est_err_eq105}
\eeq
Then with $T_e \geq c_1 (w(A)  + \sqrt{\log \log T} + \sqrt{\log (1/\delta)})^2 \log^2 k $ and choosing $c_1$ large enough so that $c = c_3 - \frac{c_{9}^2 (w(A) + \sqrt{\log \log T} + \sqrt{\log(1/\delta)})^2 \log k}{T_e}  > 0$, we get:
\beq
    \frac{1}{T_e} \|Z^{(e)}u\|_2^2 \geq c \frac{\sigma^2}{\log k} ~.
    \label{eq:single_param_gauss_est_err_eq106}
\eeq

\vspace{3mm}

{\bf 2. Upper Bounds for $\frac{1}{\sqrt{T_e}} \langle h, \Sigma^{-1/2} u \rangle$:}

$h$ is a sub-Gaussian random vector with $\|h\|_{\psi_2} \leq c_3 \kappa_{\omega}$. We use the following result from generic chaining \cite{tala05,tala14} (also Theorem 9 in \cite{bcfs14})

\begin{theo}
	Let set $B \subseteq \R^p$. Assuming $h$ is any centered sub-Gaussian random vector with $\|h\|_{\psi_2} \leq \kappa$, then we have for any $\tau_1 > 0$,
	\beq
	P \left ( \sup\limits_{u \in B} \langle h,u \rangle \geq c_6 \kappa w(B) + \tau_1 \right ) \leq \eta_4 \exp \left ( - \left ( \frac{\tau_1}{c_7 \phi \kappa} \right )^2 \right ) ~,
	\eeq
	where $c_6,\eta_4,c_7$ are positive constants and $\phi = \sup\limits_{u \in B} \|u\|_2$.
	\label{thm:single_param_noise_design_gaussian_width_deviation}
\end{theo}

Therefore applying Theorem \ref{thm:single_param_noise_design_gaussian_width_deviation} on the set $B = \{v \in \R^p ~|~ v = \Sigma^{-1/2}u, u \in A \}$ with $A$ denoting the error set, we get the following noting that $w(B) \leq \sqrt{\Lambda_{\max}(\Sigma^{-1}|A)}w(A)$ where $\sqrt{\Lambda_{\max}(\Sigma^{-1}|A)}$ denotes the restricted maximum eigenvalue of the matrix, i.e. $\Lambda_{\max}(\Sigma^{-1}|A) = \sup\limits_{u \in A} u^T \Sigma^{-1} u$ and $\phi = \sqrt{\Lambda_{\max}(\Sigma^{-1}|A)}$,
\beq
P \left ( \sup\limits_{v \in B} \langle h,v \rangle \geq c_6 c_3 \kappa_{\omega} \sqrt{\Lambda_{\max}(\Sigma^{-1}|A)}w(A) + \tau_1 \right ) \leq \eta_3 \exp \left ( - \left( \frac{\tau_1}{c_7 c_3 \sqrt{\Lambda_{\max}(\Sigma^{-1}|A)} \kappa_{\omega}} \right )^2 \right ) 
\eeq
Substituting $\tau_1 =  c_3 \kappa_{\omega} \sqrt{\Lambda_{\max}(\Sigma^{-1}|A)} ( c_7 \sqrt{ \log \log T} + c_8 \sqrt{\log (1/\delta)})$, where we choose $c_8$ such that $\left ( \frac{c_8}{c_7} \right )^2 \log(1/\delta) + \log \eta_3 > \log (1/\delta)$ we get:
\beq
P \left ( \sup\limits_{v \in B} \langle h,u \rangle \geq c_3 \kappa_{\omega} \sqrt{\Lambda_{\max}(\Sigma^{-1}|A)} (c_6 w(A) + c_7  \sqrt{\log \log T} + c_8 \sqrt{\log (1/\delta)} ) \right ) \leq \exp (- \log (1/\delta) - \log \log T) ~.
\label{eq:single_param_gauss_est_err_eq15}
\eeq
Inequality \myref{eq:single_param_gauss_est_err_eq15} is true for any episode $e$, taking a union bound over all $\lfloor \log T \rfloor$ episodes, we get for all episodes,
\beq
P \left ( \sup\limits_{e} \sup\limits_{v \in B} \langle h,u \rangle \geq c_3 \kappa_{\omega} \sqrt{\Lambda_{\max}(\Sigma^{-1}|A)} (c_6 w(A) + c_7  \sqrt{\log \log T} + c_8 \sqrt{\log (1/\delta)} ) \right ) \leq  \exp (- \log(1/\delta)) = \delta ~.
\label{eq:single_param_gauss_est_err_eq16}
\eeq

\vspace{3mm}

{\bf 3. Estimation Error: Putting it all Together}

Now consider the l.h.s of equation \myref{eq:single_param_gauss_est_err_eq13}. Using the result equation \myref{eq:single_param_gauss_est_err_eq106}, it is nonzero with probability atleast $1 - \delta \exp(-\eta_2 w^2(A)) - \delta$ when $T_e \geq c_7 (w(A) + \sqrt{\log \log T} + \sqrt{\log(1/\delta)})^2\log^2 k$. Moreover due to the preconditioning all eigenvalues are unit length and hence,
\beq
\inf\limits_{e} \inf\limits_{u \in A} \frac{1}{2T_e} \|\tilde{Z}^{(e)} u \|_2^2 \geq c_5 \|u\|_2^2 \geq c_5 ~.
\label{eq:single_param_gauss_est_err_eq14}
\eeq
Therefore from equations \myref{eq:single_param_gauss_est_err_eq13},\myref{eq:single_param_gauss_est_err_eq16} and \myref{eq:single_param_gauss_est_err_eq14}, we get that with probability atleast $1 - \delta \exp(-\eta_2 w^2(A)) - 2\delta$
\beq
\sup_e \|\hat{\theta}^{(e)} - \theta^*\|_2 = \|\Delta^{(e)}\| \leq \frac{c_9 \kappa_{\omega} \sqrt{\Lambda_{\max}(\Sigma^{-1}|A)} (c_6 w(A) + c_7 \sqrt{\log \log T} + c_8 \sqrt{\log (1/\delta)} )}{\sqrt{T_e}} ~,
\eeq
where $c_9 = \frac{c_3}{c_5}$. Now from equation \myref{eq:single_param_gauss_est_err_eq106}, $\Lambda_{\max}(\Sigma^{-1}|A) \leq \frac{\log k}{c \sigma^2}$. 
We have thus proved the advertised result.     \qed

The regret bounds stated in Theorem 2 in the main paper can now be obtained using the upper bounds on the estimation error in each episode. 

\textbf{Theorem \ref{thm:single_parameter_gaussian_regret_bounds}}
\textit{
	Consider Gaussian contexts. Then with probability atleast $1 - \delta$
	\beq
	\beta = \underset{\substack{{1 \leq i \leq k, 1 \leq t \leq T} \\  {v \in A}}}{\max} \langle x_i^t,v \rangle \leq  c_1 \sigma (w(A) + \sqrt{\log(1/\delta)}) ~.
	\eeq
	Also with $T \gg t_{\min} = c_7 (w(A) + \sqrt{ \log \log T} + \sqrt{\log(1/\delta)})^2\log^2 k$ 
	with probability atleast $1 - \delta\exp(-\eta_1 w^2(A)) - 3\delta$ the following is an upper bound on the regret for the Greedy algorithm,
	\beq
	\text{Reg}(T) \leq O \left ( \frac{ \gamma \cdot \beta \cdot \log (T) \cdot \sqrt{T}}{\sigma} \right )
	\eeq
	where $\gamma = c \kappa_{\omega} \sqrt{\log k} (w(A) + \sqrt{\log \log T} + \sqrt{\log (1/\delta)} )$ and 
}

\proof From the result of Lemma 1 we have,
\begin{equation}
\text{Reg}(T) \leq 4 \beta  t_{\min} + \sum_{e = \lceil \log t_{\min} \rceil}^{\lfloor \log T \rfloor} \sum_{1}^{T_e} 2 \beta \|\hat{\theta}^{(e)} - \theta^*\|_2 ~.
\label{eq:single_param_gaussian_regret_bounds_eq1}
\end{equation}

From the result in Theorem 1, we need $T_e > t_{\min} = c_7(w(A) + \sqrt{ \log \log  T} + \sqrt{\log (1/\delta)})^2\log^2 k$ for the RE condition to be satisfied. Moreover in each episode $e$ we use the $\hat{\theta}^{(e)}$ estimated using rounds played in the previous episode $e-1$ with $T_{e} = 2T_{e-1}$. 
Therefore substituting from the result of Theorem 3 the value of $\|\hat{\theta}^{(e)} - \theta^*\|_2$ in \myref{eq:single_param_gaussian_regret_bounds_eq1} we get,
\begin{align}
\text{Regret}(T) \leq
&4 \beta  t_{\min} + \sum_{e = \lceil \log t_{\min} \rceil}^{\lfloor \log T \rfloor} \sum_{1}^{T_e} 2 \beta \|\hat{\theta}^{(e)} - \theta^*\|_2 \nonumber \\
&\leq 4 \beta  c_7(w(A) + \sqrt{\log \log T} + \sqrt{\log (1/\delta)})^2 \log^2 k + \sum_{e = \lceil \log t_{\min} \rceil}^{\lfloor \log T \rfloor} \sum_{1}^{T_e} \frac{2 c\beta \gamma}{\sigma\sqrt{T_{e-1}}} \nonumber \\
&\leq 4 \beta  c_7(w(A) + \sqrt{\log \log T} + \sqrt{\log (1/\delta)})^2 \log^2 k + \sum_{e = \lceil \log t_{\min} \rceil}^{\lfloor \log T \rfloor} \frac{4c \beta \gamma T_{e-1}}{\sigma \sqrt{T_{e-1}}} \nonumber \\
&\leq 4 \beta  c_7(w(A) + \sqrt{\log \log T} + \sqrt{\log (1/\delta)})^2 \log^2 k + \frac{4 c\beta  \gamma \sqrt{T} \log T}{\sigma} ~,
\label{eq:single_param_gaussian_regret_bounds_eq2}
\end{align}
where in the second line we use that in the $e$th episode we play with $\hat{\theta}^{(e)}$ estimated using $T_{e-1}$ rounds played in the previous episode, in the third line we use $T_e = 2 T_{e-1}$ and in the last line we use $T > T_e$ for all $e$. 

Also, 
we have by properties of Gaussian width,
\beq
    E \left [ \sup\limits_{v \in A} \left | \langle x_i^t,v \rangle \right | \right ] =  \Theta(\sigma w(A)) ~. 
\eeq
Again by Theorem 2.2.27 in \cite{tala14}, we get
\beq
    P \left (\sup\limits_{v \in A} \left | \langle x_i^t,v \rangle \right | \geq E \left [ \sup\limits_{v \in A} \left | \langle x_i^t,v \rangle \right | \right ] + c_2 \sigma \tau  \right ) \leq \exp \left (-\tau^2 \right ) ~.
\eeq
Choosing $\tau = \sqrt{\log(1/\delta)}$, we get the stated result. \qed

	\section{Proofs for Single Parameter Setting with Smoothed Adversary}
	\label{sec:supp_single_param_proofs_adversary}
	We give proof for Lemma 3.

\textbf{Lemma \ref{lemm:single_param_obl_design_matrix}}
\textit{
	The rows of the design matrix $Z^{(e)} \in \R^{T_e \times p}$ in any episode $e$ are $z^t = \mu^t + g^t$ where $\mu^t,g^t = \underset{\mu_i^t,g_i^t:1 \leq i \leq k}{\argmax} \langle \mu_i^t + g_i^t,\hat{\theta}^{(e-1)} \rangle$, $g_i^t \sim N(0,\sigma^2 \I_{p \times p})$ with the sub-Gaussian norm of $g^t$ satisfying $\|g^t\|_{\psi_2} \leq c_2 \sigma \sqrt{\log k}$ for some constant $c_2$. Moreover we have the following lower bound on the expected minimum eigenvalue for any $\mu_i^t$'s:
	\beq
	\lambda_{\min} (E_{z^t}[z^t (z^t)^\intercal]) \geq c_1 \frac{\sigma^2}{\log k} ~,
	\eeq
	where $c_1$ is some constant.
}

\proof  For convenience we drop the superscript from $\hat{\theta}^{(e-1)}$. To bound the minimum eigenvalue we make the following observation,

\begin{align}
\lambda_{\min} \left ( E \left [z^t (z^t)^{\intercal} \suchthat x^t = \underset{x_i^t:1 \leq i \leq k}{\argmax} \langle x_i^t,\hat{\theta} \rangle \right ] \right )
&= \min\limits_{w:\|w\|_2 = 1} w^T \left (E \left [z^t(z^t)^{\intercal} \suchthat z^t = \underset{x_i^t:1 \leq i \leq k}{\argmax} \langle x_i^t,\hat{\theta} \rangle \right ] \right) w \nonumber \\
&= \min\limits_{w:\|w\|_2 = 1}
\left ( E \left [w^{\intercal}z^t(z^t)^{\intercal} w \suchthat  z^t = \underset{x_i^t:1 \leq i \leq k}{\argmax}  \langle x_i^t,\hat{\theta} \rangle \right] \right)  \nonumber \\
&\geq \min\limits_{w:\|w\|_2 = 1}\text{Var} \left (  \left [\langle w,z^t \rangle \suchthat  z^t = \underset{x_i^t:1 \leq i \leq k}{\argmax} \langle x_i^t,\hat{\theta} \rangle \right ] \right ) \nonumber \\
&\geq \min\limits_{w:\|w\|_2=1} \text{Var} \left ( \left [ \langle w,g^t \rangle \suchthat g^t = \underset{g_i^t:1 \leq i \leq k}{\argmax} \langle \mu_i^t + g_i^t,\hat{\theta} \rangle \right ] \right ) ~,
\label{eq:eq_ref5}
\end{align}
where the last line follows because $\langle w,z^t \rangle = \langle w,\mu^t \rangle + \langle w,g^t \rangle$.

We will now prove that
\beq
\min\limits_{w:\|w\|=1} \text{Var} \left [\langle g^t,w \rangle \suchthat g^t = \underset{g_i^t:1 \leq i \leq k}{\argmax} \langle \mu_i^t + g_i^t, \hat{\theta} \rangle \right ] \geq  \min\limits_{w:\|w\|=1} \text{Var} \left [ \langle g^t,w \rangle \suchthat g^t = \underset{g_i^t:1 \leq i \leq k}{\argmax} \langle g_i^t,\hat{\theta} \rangle \right ]~.
\label{eq:eq_ref6}
\eeq

Therefore the worst any adversary can do is to ensure that the context corresponding to $g^t = \underset{g_i^t:1 \leq i \leq k}{\argmax} \langle g_i^t,\hat{\theta} \rangle$ is chosen in each round. In fact this can be achieved by choosing $\mu_1^t = \mu_2^t = \hdots = \mu_k^t$ in any round.

We make the following observations. Let $Q$ be an orthogonal matrix such that $Q \hat{\theta} = (\|\hat{\theta}\|_2,0,\hdots,0)$. Also let $(g_1^t,\hdots,g_k^t) = (Q^T\epsilon_1^t,\hdots,Q^T\epsilon_k^t)$. Due to rotational invariance $\epsilon_i^t \sim N(0,\sigma^2 \I_{p \times p}), 1 \leq i \leq k$. Therefore,
\begin{align}
\min\limits_{w:\|w\|=1} \text{Var} \left [\langle g^t, w \rangle \suchthat g^t = \underset{g_i^t:1 \leq i \leq k}{\argmax} \langle \mu_i^t + g_i^t,\hat{\theta} \rangle \right ]
&= \min\limits_{w:\|w\|=1} \text{Var} \left [\langle Qg^t, Qw \rangle \suchthat g^t = \underset{g_i^t:1 \leq i \leq k}{\argmax} \langle Q\mu_i^t + Qg_i^t,Q\hat{\theta} \rangle \right] \nonumber \\
&= \min\limits_{w:\|w\|=1} \text{Var}\left [\langle \epsilon^t,Qw \rangle \suchthat \epsilon^t = \underset{\epsilon_i^t:1 \leq i \leq k}{\argmax} \langle Q\mu_i^t + \epsilon_i^t,Q\hat{\theta} \rangle \right] \nonumber \\
&= \min\limits_{w:\|w\|=1} \text{Var} \left [\langle \epsilon^t,w \rangle \suchthat \epsilon^t = \underset{\epsilon_i^t:1 \leq i \leq k}{\argmax} (Q\mu_i^t + \epsilon_i^t)_1 \right ] \nonumber \\
&= \min\limits_{w:\|w\|=1} \left (w_1^2 \text{Var}((\epsilon^t)_1) \suchthat \epsilon^t = \underset{\epsilon_i^t:1 \leq i \leq k}{\argmax} (Q\mu_i^t + \epsilon_i^t)_1 \right ) + \\
& \left ( \sum_{j=2}^p w_j^2 \text{Var}((\epsilon^t)_j) \suchthat \epsilon^t = \underset{\epsilon_i^t:1 \leq i \leq k}{\argmax} (Q\mu_i^t + \epsilon_i^t)_1 \right ) \nonumber \\
&\geq c \frac{\sigma^2}{\log k}
\end{align} 
where the last line is because the coordinates of $\epsilon^t$ are independent and from Lemma \ref{lemm:adversary_gaussian_variance} and \ref{lemm:max_gauss_variance} we have 
\beq
\left ( \text{Var}((\epsilon^t)_1 \suchthat \epsilon^t = \underset{\epsilon_i^t:1 \leq i \leq k}{\argmax} (Q\mu_i^t + \epsilon_i^t)_1) \right ) \geq \left ( \text{Var}((\epsilon^t)_1 \suchthat \epsilon^t = \underset{\epsilon_i^t:1 \leq i \leq k}{\argmax}  (\epsilon_i^t)_1\right ) \geq \frac{\sigma^2}{\log k}
\eeq
and 
$\text{Var}\left ((\epsilon^t)_j \bigm| \epsilon^t = \underset{\epsilon_i^t:1 \leq i \leq k}{\argmax} (Q\mu_i^t + \epsilon_i^t)_1) \right ) = \sigma^2 $. That completes the proof. \qed

Next we obtain estimation error bounds in the smoothed adversary setting.
\begin{theo}
	The design matrix $Z^{(e)} \in \R^{T_e \times p}$ in all episode where $T_e \geq c_1 (w(A)  + \sqrt{\log \log T} + \sqrt{\log (1/\delta)})^2\log^2 k$ satisfies the following minimum eigenvalue condition with probability atleast $1 - \delta \exp(-\eta_1 w^2(A)) - 2\delta$,
	\beq
	\inf\limits_{u \in A} \frac{1}{T_e} \|Z^{(e)} u \|_2^2 \geq c \frac{\sigma^2}{\log k} ~.
	\eeq
	Moreover, for all episodes when $T_e \geq c_1 (w(A)  + \sqrt{\log \log T} + \sqrt{\log (1/\delta)})^2\log^2 k$ 
	with probability atleast $1 - \delta\exp(-\eta_1 w^2(A)) - 3\delta$, 
	\beq
	\|\hat{\theta}^{(e+1)} - \theta^* \|_2 \leq O \left ( \frac{\gamma }{\sigma \sqrt{T_e}} \right ) ~,
	\eeq
	where $\gamma = c\kappa_{\omega} \sqrt{\log k} (w(A) + \sqrt{\log \log T} + \sqrt{\log (1/\delta)})$.
\end{theo}

\proof Using similar arguments as Theorem \ref{thm:ch6_single_param_gaussian_est_error}, we get:
\beq
    \frac{1}{2T_e} \|\tilde{Z}^{(e)} u \|_2^2 \|\Delta^{(e)}\|_2^2 \leq \frac{1}{\sqrt{T_e}} \langle h, \Sigma^{-1/2}u \rangle \|\Delta^{(e)}\|_2 ~.
\eeq
Note that $h = VU^{\intercal} \omega^{(e)}$ is a sub-Gaussian random vector $\|h\|_{\psi_2} \leq c_1 \kappa_{\omega}$ by direct application of Lemma \ref{lemm:ch6_hoeffding_martingale}.  
We obtain lower bounds for $\inf\limits_{u \in A} \frac{1}{T_e} \|Z^{(e)} u \|_2^2$ and upper bounds for $\sup\limits_{u \in A}\frac{1}{\sqrt{T_e}} \langle h, \Sigma^{-1/2}u \rangle$

\vspace{3mm}

{\bf 1. Lower bounds for $\inf\limits_{u \in A} \frac{1}{T_e} \|Z^{(e)} u \|_2^2$}
 
We first prove that $\inf\limits_{u \in A} \frac{1}{T_e} \|Z^{(e)} u \|_2^2 \geq c \frac{\sigma^2}{\log k}$ with high probability when $T_e \geq c_1 (w^2(A) + \sqrt{\log \log T} + \sqrt{\log(1/\delta)})^2 \log^2 k$. We make the following observations for some $u \in A$,
\begin{align}
    \frac{1}{T_e} \|Z^{(e)}u\|_2^2 &= \frac{1}{T_e} \sum_{t = 1}^{T_e} \langle z^t,u \rangle^2 \nonumber \\
    &= \frac{1}{T_e} \sum_{t = 1}^{T_e} \left \langle g^t - E[g^t] + E[g^t] + \mu^t, u \right \rangle^2 \nonumber \\
    &= \frac{1}{T_e} \sum_{t=1}^{T_e} \left \langle g^t - E[g^t], u \right \rangle^2 + \frac{1}{T_e} \sum_{t=1}^{T_e} \left \langle E[g^t] + \mu^t,u \right \rangle^2 + \frac{2}{T_e}\sum_{t=1}^{T_e} \left \langle  g^t - E[g^t],u \right \rangle \left \langle E[g^t] + \mu^t,u \right \rangle \nonumber \\
    &= \frac{1}{T_e} \sum_{t=1}^{T_e} \left \langle g^t - E[g^t], u \right \rangle^2 + \|\alpha\|_2^2 + \frac{2\|\alpha\|_2}{\sqrt{T_e}} \sum_{t=1}^{T_e} \alpha^t \left \langle g^t - E[g^t], u \right \rangle 
    \label{eq:single_param_obl_est_err_eq9}
\end{align}
where we denote $\alpha = \frac{1}{\sqrt{T_e}}[\langle E[g^1] + \mu^1,u \rangle, \hdots, \langle E[g^{T_e}] + \mu^{T_e},u \rangle ] \in \R^{T_e}$.

We will first obtain lower bounds for the quantity $\inf\limits_{u \in A} \frac{1}{T_e} \sum_{t=1}^{T_e} \left \langle g^t - E[g^t], u \right \rangle^2$ where $A$ is the error set. Compared to the Gaussian context setting, the $g^t$'s can no longer be assumed to be independent. The $g^t$'s are adaptively generated based on observing the history of contexts chosen in earlier rounds and the corresponding rewards. We adopt the nomenclature in \cite{bgsw19} to use their Theorem 5. Let $\xi^t = g^t - E[g^t]$ denote the centered random smoothing vector with $\|\xi^t\|_{\psi_2} \leq \sigma$ (see result before equation \myref{eq:ch6_center_subg_norm}) and $\xi = [(\xi^1)^{\intercal} ,\hdots,(\xi^{T_e})^{\intercal}]^{\intercal} \in \R^{T_e p \times 1}$ be a random vector formed  by concatenating the rows of the centered random smoothed component. 
Also let $V \in R^{T_e \times T_e p}$ denote the following matrix indexed by vectors $u \in A$:
\beq
        V(u) = \frac{1}{\sqrt{T_e}} \begin{bmatrix}
            u^T & 0 & \cdots & 0 \\
            0 & u^T & \cdots & 0 \\
            \vdots & \vdots & \ddots & \vdots \\
            0 & 0 & \cdots & u^T
        \end{bmatrix}.
        \label{eq:vtrans}
\eeq
Then by simple algebraic manipulations the following is a straightforward observation with $\Xi \in \R^{T_e \times p}$ denoting the random matrix obtained by stacking the $g^t$ as rows:
\beq
     \frac{1}{T_e} \sum_{t=1}^{T_e} \langle \xi^t,u \rangle^2 =  \frac{1}{T_e} \|\Xi u\|_2^2 = \|V(u) \xi\|_2^2 ~.
\eeq
To obtain lower bounds on $\inf\limits_{u \in A} \|V(u)\xi|_2^2$ we focus on lower bounding $\inf\limits_{u \in A} \left ( \|V(u) \xi\|_2^2 - E\|V(u) \xi\|_2^2 \right )$ which can be obtained using the result of Theorem 5 in \cite{bgsw19}. To apply Theorem 5, we first show that the random quantity satisfies the conditions required to apply the result of Theorem 5. Application of Theorem 5 in \cite{bgsw19} requires the data generated to satisfy conditions (SP-1) and (SP-2) manifested by three graphical models. We first show that the data generation in the contextual bandit problem can be modelled using graphical model GM3 in \cite{bgsw19}. We make the following observations:
\begin{enumerate}
\item Let $\mathcal{H}_{t-1}$ denote historical data observed until time $t-1$. In time step $t-1$ an adaptive adversary $\mathcal{A}_{t-1}$  maps the histories to $k$ contexts $\mu_1^t,\hdots,\mu^t_k$ in $\R^p$ with $\|\mu_i^t\|_2 \leq 1$, i.e., $\mathcal{A}_{t-1}: \mathcal{H}_{t-1} \rightarrow (B_2^p)^k$ where $B_2^p$ represents the unit ball in $p$ dimensions. Nature perturbs the contexts with random Gaussian noise, i.e., $x_i^t = \mu_i^t + g_i^t$ with $g_i^t \sim N(0,\sigma^2 \I_{p \times p})$. Now, in the context of graphical model GM3, $\mathcal{H}_{t-1} \cup \{x_1^t,\hdots,x_k^t \}$ represents $F_{1:t-1}$. 
\item In time step $t$, a learner chooses one among $k$ contexts $\{x_1^t,\hdots,x_k^t \}$ based on historical data $\mathcal{H}_{t-1}$. Let $z^t$ denote the selected context and $g_t$ denote the corresponding Gaussian perturbation. In the context of GM3, we denote the centered Gaussian perturbation $g^t - \E[g^t]$ by $\xi^t$.
The learner receives the noisy reward $y^t = \langle z^t,\theta^* \rangle + \omega^t$ where $\omega^t$ is an unknown sub-Gaussian noise. History at time step $t$ is now augmented with the new data, i.e., $\mathcal{H}_t = \mathcal{H}_{t-1} \cup \{ \{x_1^t,\hdots,x_k^t \}, z^t,y^t \}$.
\item Now similar to step 1, the contexts in time step $t$, $\{x_1^{t+1},\hdots,x_k^{t+1} \}$, are generated by an adversary $\mathcal{A}_t: \mathcal{H}_t \rightarrow (B_2^p)^k$ perturbed with Gaussian noise and $\mathcal{H}_t \cup \{x_1^{t+1},\hdots,x_k^{t+1} \}$ represents $F_{1:t}$.
\end{enumerate}

\begin{lemm}
    Let $G,\xi,V(u),\nu$  be constructed as above. Define the set $\cA = \{ V(u) ~\suchthat ~ u \in A \}$. Then with probability atleast $1 - \exp(-c_9\epsilon^2 T_e)$
    \beq
        \inf\limits_{V(u) \in \cA} \left ( \|V(u)\xi\|_2^2 - E\|V(u)\xi\|_2^2 \right ) \geq \sigma^2 \left ( -c_{10} \frac{w(A)}{T_e} - \epsilon \right ) ~.
    \eeq
    \label{lemm:ch6_subg_martingale_conc}
\end{lemm}
\proof We start with the result of Theorem 5 in \cite{bgsw19}. Let $\xi'$ be a random vector constructed similar to $\xi$ but with 1-sub-Gaussian norm. Therefore $\xi_i = c_4 \sigma \xi'_i$ for some constant $c_4$. Also,
\begin{align}
   & \|V(u)\xi\|_2^2 = c_4^2 \sigma^2 \|V(u)\xi'\|_2^2 \nonumber \\
   & E\|V(u)\xi\|_2^2 = c_4^2 \sigma^2 \|V(u)\xi'\|_2^2 ~.
   \label{eq:ch6_subg_transform}
\end{align}
We now apply Theorem 5 and Corollary 4 to obtain bounds on $\inf\limits_{u \in A} \|V(u) \xi'\|_2^2$. The values of the quantities in Theorem 5 of \cite{bgsw19} are $\|V(u)\|_F = \|u\|_2 = 1$, $d_F(\cA) = 1$, $\|V(u)\|_{2 \rightarrow 2} = \frac{1}{\sqrt{T_e}} \|u\|_2 = \frac{1}{\sqrt{T_e}}$ and $d_{2 \rightarrow 2}(\cA) = \frac{1}{\sqrt{T_e}}$. Also the Gaussian width of the set $\cA$:
\beq
\gamma_2(\cA, \|\cdot\|_{2 \rightarrow 2}) \leq c_5 \frac{w(A)}{\sqrt{T_e}} ~.
\eeq
Therefore we have,
\beq
    M \leq c_6 \left( \frac{w(A)}{T_e} \right ), \quad V = O\left (\frac{1}{\sqrt{T_e}} \right ), \quad U = \frac{1}{T_e} ~.
\eeq
Then by application of result in Theorem 5 in \cite{bgsw19}, with $0 < \epsilon' < 1$ with probability atleast $1 - \exp(-c_7(\epsilon')^2 T_e)$,
\beq
\inf\limits_{V(u) \in \cA}  \left ( \|V(u)\xi'\|_2^2 - E\|V(u)\xi'\|_2^2 \right ) \geq  - c_8 \frac{w(A)}{T_e} - \epsilon' ~. 
\eeq
Now from the relationship \myref{eq:ch6_subg_transform}, we get
with probability atleast $1 - \exp(-c_9\epsilon^2 T_e)$,
\beq
    \inf\limits_{V(u) \in \cA}  \left ( \|V(u)\xi\|_2^2 - E\|V(u)\xi\|_2^2 \right ) \geq \sigma^2 \left ( - c_{10}\frac{w(A)}{T_e} - \epsilon  \right ) ~,
\eeq
where $\epsilon = c_4^2 \epsilon'$. $c_{10} = c_8 c_4^2$ and $c_9 = c_7/c_4^4$. This proves the stated result. \qed

From the result of Lemma \ref{lemm:single_param_obl_design_matrix} we have,
\beq
     \text{Var}[\langle g^t,u \rangle] = E[ \langle g^t - E[g^t],u \rangle^2] \geq c_1\frac{\sigma^2}{\log k} ~.
\eeq 
Therefore by simple algebraic manipulations we get,
\beq
    E\|V(u)\xi\|_2^2 \geq c_{11} \frac{\sigma^2}{\log k} ~.
\eeq
Therefore using the result of Lemma \ref{lemm:ch6_subg_martingale_conc} with probability atleast $1 - \exp(-c_9\epsilon^2 T_e)$, we get
\beq
    \inf\limits_{V(u) \in \cA} \|V(u)\xi\|_2^2 \geq \frac{\sigma^2}{\log k} \left ( c_{11} - c_{10}\frac{w(A)\log k}{T_e} - \epsilon \log k \right )
\eeq
Now choosing $T_e \geq c_1 (w(A) + \sqrt{\log \log T} + \sqrt{\log (1/\delta)})^2 \log^2 k $ with $c_1 > 1$ large enough so that $0 < c_3 \leq \left ( c_{11} - c_{10}\frac{w(A)\log k}{T_e} - \epsilon \log k  \right )$, choosing $\epsilon \leq \frac{\epsilon'}{\log k}$ with $0 < \epsilon' < 1$ and choosing $\eta_1 = c_7 (\epsilon')^2 c_1$, we get with probability atleast $1 - \delta \exp(-\eta_1 w^2(A) - \log \log T)$,
\beq
    \inf\limits_{V(u) \in \cA} \|V(u)\xi\|_2^2 \geq c_3 \frac{\sigma^2}{\log k} ~.
\eeq
This is the bound for estimation in episode $e$. Taking a union bound over all episodes $e < \log \log T$, we get that with probability atleast $1 - \delta\exp(-\eta_1 w^2(A))$ over all rounds,
\beq
 \inf\limits_{u \in A} \sum_{t=1}^{T_e} \frac{1}{T_e} \langle g^t - E[g^t],u \rangle^2 = \inf\limits_{V(u) \in \cA} \|V(u)\xi\|_2^2 \geq c_3 \frac{\sigma^2}{\log k} ~.
 \label{eq:single_param_obl_est_err_eq10}
\eeq

We now obtain upper bounds for $\sup\limits_{u \in A} \left | \sum_{t=1}^{T_e} \alpha^t \langle g^t - E[g^t],u \rangle \right |$.

Note that $g^t - E[g^t]$ is a $c_1 \sigma$-sub-Gaussian random vector and hence $\beta^t = \langle g^t - E[g^t],u \rangle$ is a centered $c_1 \sigma \|u\|_2 = c_1 \sigma$ sub-Gaussian random variable by Lemma \ref{lemm:rotsubGau} in Section \ref{sec:supp_back}. Also $\beta^t$'s are MDS with $E[\beta^i|\beta^1,\hdots,\beta^{i-1}] = 0$ and the coefficients $\alpha^1,\hdots,\alpha^t$ are adaptive, i.e., $\alpha_i = f_i((x_1^1,\hdots,x_k^1,z^t,y^1),\hdots,(x_1^{i-1},\hdots,x_k^{i-1},z^{i-1},y^{i-1}))$ depends on the history of the previously seen contexts and rewards. By an application of Lemma \ref{lemm:ch6_hoeffding_martingale} for some $u \in A$, we get,
\beq
    P\left ( \left | \sum_{t=1}^{T_e} \alpha^t \langle g^t - E[g^t],u \rangle \right | \geq  \tau   \right ) \leq 2 \exp \left ( - \frac{\tau^2}{c_2 \sigma^2 \|\alpha\|_2^2} \right )
\eeq
Now for any $u,v \in A$, $\langle g^t - E[g^t],u - v \rangle$ is a $c_1 \sigma \|u-v\|_2$-sub-Gaussian random variable. Therefore by the application of  Lemma \ref{lemm:ch6_hoeffding_martingale} we get,
\beq
    P\left ( \left | \sum_{t=1}^{T_e} \alpha^t \langle g^t - E[g^t],u-v \rangle \right | \geq  \tau   \right ) \leq
    2 \exp \left ( - \frac{\tau^2}{c_2 \sigma^2 \|u-v\|_2^2 \|\alpha\|_2^2} \right ) ~.
\eeq
Therefore substituting $\sigma_1 = \sqrt{c_2}\sigma \|\alpha\|_2$, we get,
\beq
    P\left ( \left | \sum_{t=1}^{T_e} \alpha^t \langle g^t - E[g^t],u-v \rangle \right | \geq  \tau   \right ) 
    \leq 2 \exp \left ( - \frac{\tau^2}{\sigma_1^2 \|u-v\|_2^2} \right ) ~.
\eeq
Therefore, from the definition of Gaussian width and the majorizing measures theorem \cite{tala05,tala14},
\beq
   E \left [\sup\limits_{u \in A} \left | \sum_{t=1}^{T_e}   \alpha^t \langle g^t - E[g^t],u \rangle  \right | \right ] \leq c_3 \sigma_1 w(A) = c_4 \sigma \|\alpha\|_2 w(A) ~.
    \label{eq:single_param_obl_est_err_eq101}
\eeq
Now for the high probability bounds we refer Theorem 2.2.27 in \cite{tala14}. Applying the result of Theorem 2.2.27 \cite{tala14} leads to the following:
\beq
    P \left ( \sup\limits_{u \in A} \left | \sum_{t=1}^{T_e} \alpha^t \langle g^t - E[g^t],u \rangle \right |  \geq E \left [\sup\limits_{u \in A} \left | \sum_{t=1}^{T_e} \alpha^t \langle g^t - E[g^t],u \rangle \right |  \right ] + c_5 \sigma_1 \tau  \right )
    \leq c_6 \cdot \exp ( -\tau^2) ~.
\eeq
Let $\tau = c_7 (\sqrt{\log(1/\delta)} + \sqrt{\log \log T})$ choosing $c_7$ large enough so that $c_6 \cdot \exp(-\tau^2) \geq c_6 \cdot \exp ( -c_7^2 (\log \log T + \log(1/\delta))) \geq \exp (- \log \log T - \log(1/\delta))$. Also substituting the value of $E \left [\sup\limits_{u \in A} \left | \sum_{t=1}^{T_e}   \alpha^t \langle g^t - E[g^t],u \rangle  \right | \right ]$ from equation \myref{eq:single_param_obl_est_err_eq101} and choosing constant $c_8$ large enough, we get the following:
\beq
    P \left ( \sup\limits_{u \in A} \left | \sum_{t=1}^{T_e} \alpha^t \langle g^t - E[g^t],u \rangle \right | \geq c_8 \sigma \|\alpha\|_2 (w(A) + \sqrt{\log(1/\delta)} + \sqrt{\log \log T}) \right ) \leq \exp \left ( - \log (1/\delta) - \log \log T \right ) ~.
\eeq
This above is true for any episode $e$. Taking a union bound over all $\lfloor \log \log T \rfloor$ episodes, we get,
\begin{align}
    P \left ( \sup\limits_{u \in A} \left | \sum_{t=1}^{T_e} \alpha^t \langle g^t - E[g^t],u \rangle \right | \geq c_8 \sigma \|\alpha\|_2 (w(A) + \sqrt{\log(1/\delta)} + \sqrt{\log \log T}) \right ) &\leq \exp \left ( - \log (1/\delta) - \log \log T + \log \log T \right ) \nonumber \\
    &= \delta ~.
    \label{eq:single_param_obl_est_err_eq102}
\end{align}

From equations \myref{eq:single_param_obl_est_err_eq9}, \myref{eq:single_param_obl_est_err_eq10} and \myref{eq:single_param_obl_est_err_eq102}, we get with probability atleast $1 - \delta \exp(-\eta_1 w^2(A)) - \delta$,
\beq
 \frac{1}{T_e} \|Z^{(e)}u\|_2^2 \geq c_3 \frac{\sigma^2}{\log k} + \|\alpha\|_2^2 - \frac{2c_8 \sigma \|\alpha\|_2}{\sqrt{T_e}} (w(A) + \sqrt{\log(1/\delta)} + \sqrt{\log \log T}) ~.
 \label{eq:single_param_obl_est_err_eq103}
\eeq
Equation \myref{eq:single_param_obl_est_err_eq103} is minimized when $\|\alpha\|_2 = \frac{c_8 \sigma}{\sqrt{T_e}}(w(A) + \sqrt{\log(1/\delta)} + \sqrt{\log \log T})$. Substituting the minimum value in equation \myref{eq:single_param_obl_est_err_eq103} and by simple algebraic manipulations, we get,
\beq
    \frac{1}{T_e} \|Z^{(e)}u\|_2^2  \geq \frac{\sigma^2}{\log k} \left ( c_3 - \frac{c_8^2 (w(A) + \sqrt{\log(1/\delta)} + \sqrt{\log \log T}) \log k}{\sqrt{T_e}}\right )
\eeq
Then with $T_e \geq c_1 (w(A)  + \sqrt{\log \log T} + \sqrt{\log (1/\delta)})^2 \log^2 k $ and choosing $c_1$ large enough so that $c = c_3 - \frac{c_{8}^2 (w(A) + \sqrt{\log \log T} + \sqrt{\log(1/\delta)})^2 \log k}{\sqrt{T_e}}  > 0$, we get the advertised result on the minimum eigenvalue.

\vspace{3mm}

{\bf 2. Upper Bounds for $\frac{1}{\sqrt{T_e}} \langle h, \Sigma^{-1/2} u \rangle$:}

The following upper bound can be obtained using similar arguments as Theorem \ref{thm:ch6_single_param_gaussian_est_error}:
\begin{align}
P \left ( \sup\limits_{e} \sup\limits_{v \in B} \langle h,u \rangle \geq c_3 \kappa_{\omega} \sqrt{\Lambda_{\max}(\Sigma^{-1}|A)} (c_6 w(A) + c_7  \sqrt{\log \log T} + c_8 \sqrt{\log (1/\delta)} ) \right ) \leq  \exp (- \log(1/\delta)) = \delta ~.
\label{eq:single_param_adv_est_err_eq16}
\end{align}

\vspace{3mm}

{\bf 3. Estimation Error: Putting it all Together}
Again by following similar arguments as Theorem \ref{thm:ch6_single_param_gaussian_est_error}, we obtain the following estimation error bounds with probability atleast $1 - \delta \exp(-\eta_2 w^2(A)) - 2\delta$:
\beq
    \sup\limits_{e} \|\hat{\theta}^{(e)} - \theta^*\|_2 = \|\Delta^{(e)}\|_2 \leq \frac{c_9 \kappa_{\omega}(c_6 w(A) + c_7 \sqrt{\log \log T} + c_8 \sqrt{\log (1/\delta)})}{\sigma \sqrt{T_e}} ~.
\eeq        \qed

\textbf{Theorem 4}
\textit{
	In the oblivious smoothed adversary setting with probability atleast $1 - 2\delta$
	\beq
	\beta = \underset{\substack{{1 \leq i \leq k, 1 \leq t \leq T} \\  {v \in A}}}{\max} \langle x_i^t,v \rangle \leq  (1+ c_1 \sigma (w(A) + \sqrt{\log(1/\delta)} )) 
	\eeq
	Also with $T \gg t_{\min} \geq c_1( w(A)  + \sqrt{\log \log T} + \sqrt{\log (1/\delta)})^2\log^2 k$ 
	with probability atleast $1 - \delta\exp(-\eta_1 w^2(A)) - 3\delta$ the following is an upper bound on the regret,
	\beq
	\text{Reg(T)} \leq O \left ( \frac{\gamma \cdot \beta \cdot \log (T) \cdot \sqrt{T} }{\sigma} \right )  ~,
	\eeq
	where $\gamma = c\kappa_{\omega} \sqrt{\log k} (w(A) + \sqrt{\log \log T} + \sqrt{\log (1/\delta)})$.
}

\proof 
We argue similar to Theorem 2 to get bounds,
\beq
\text{Reg}(T) \leq 4 \beta c( w(A) + \sqrt{\log \log T} + \sqrt{\log(1/\delta)})^2\log^2 k + \frac{4 c\beta \gamma \sqrt{T} \log T}{\sigma}  ~.
\eeq
The result on $\beta$ follows from Theorem 1 and noting that $\|\mu_i^t\|_2 \leq 1$. \qed

	\section{Proofs for Multi Parameter Setting}
	\label{sec:supp_multi_param_proofs}
	\textbf{Lemma \ref{lemm:multi_param_reg_bnds}}
\textit{
	The greedy algorithm plays the contexts in an episodic fashion with the maximum episode number for each context $e_i \leq e_{i,\max} \leq \lfloor \log T \rfloor$. Denote by $\beta = \underset{\substack{{1 \leq i \leq k, 1 \leq t \leq T} \\  {v \in A}}}{\max} \langle x_i^t,v \rangle$.
	Let $t_{\min} < T$, where $t_{\min}$ depends on properties of the true parameters $\theta^*_i$, the regularizer $R(\cdot)$, the noise properties, the number of contexts $k$ and the quantity $\beta$. Then,
	\beq
	\text{Reg}(T) \leq 2 \beta t_{\min} + \sum_{i=1}^k \sum_{e_i=1}^{e_{i,\max}} \left ( \sum_{1}^{T_{i,e_i}} \beta \|\theta^*_{i} - \hat{\theta}_{i}^{(e_i)}\|_2 + \sum_{1}^{T_{i,e_i}^*} \beta \|\theta^*_i - \hat{\theta}_i^{(e_i)}\|_2 \right ) 
	\eeq
}

\proof Let $i^{*}(t) = \underset{1 \leq i \leq k}{\argmax} \langle x_{i}^t,\theta^*_i \rangle$ denote the optimal context in any round $t$. Its context, for shorthand, is $\theta^*_{i^*}$ and let $x_{i^*}^t$ denote the context. Let $i^t$ denote the context chosen in round $t$. The regret can be computed as follows,
\begin{align}
\text{Reg}(T) &\leq \sum_{t=1}^T \langle \theta^*_{i^*},x_{i^*}^t \rangle - \langle \theta_{i^t}^*,x_{i^t}^t \rangle \nonumber \\
&\leq \sum_{t=1}^{t_{\min}} \langle \theta^*_{i^*},x_{i^*}^t \rangle - \langle \theta_{i^t}^*,x_{i^t}^t \rangle + \sum_{t = t_{\min} + 1}^T \langle \theta^*_{i^*},x_{i^*}^t \rangle - \langle \theta_{i^t}^*,x_{i^t}^t \rangle
\label{eq:multi_param_regret_eq1}
\end{align}
The first term on the r.h.s. of \myref{eq:multi_param_regret_eq1} can be upper bounded as follows,
\begin{align}
\sum_{t=1}^{t_{\min}} \langle \theta^*_{i^*},x_{i^*}^t \rangle - \langle \theta_{i^t}^*,x_{i^t}^t \rangle &\leq \sum_{t=1}^{t_{\min}} | \langle \theta^*_{i^*},x_{i^*}^t \rangle | + |  \langle \theta_{i^t}^*,x_{i^t}^t \rangle | \nonumber \\
&\leq \sum_{t = 1}^{t_{\min}} 2\beta \\
&\leq 2 \beta t_{\min} ~.
\label{eq:multi_param_regret_eq2}
\end{align}

To bound the second term on the r.h.s. in \myref{eq:multi_param_regret_eq1}, assume in round $t$ let $e_{i^*(t)}$ denote the episode number corresponding to the optimal context $i^*$ and $e_{i^t}$ denote the episode number corresponding to the selected context $i^t$. Again for shorthand we denote $e_{i^*(t)}$ by $e_{i^*}$. Let $T_1,\hdots,T_k$ be the total number of rounds where contexts $1,\hdots,k$ are played respectively. 
\begin{align}
\sum_{t = t_{\min} + 1}^T \langle \theta^*_{i^*},x_{i^*}^t \rangle - \langle \theta_{i^t}^*,x_{i^t}^t \rangle
&= \sum_{t = t_{\min} + 1}^T \langle \theta^*_{i^*} - \hat{\theta}^{(e_{i^*})}_{i^*},x_{i^*}^t \rangle - \langle \theta_{i^t}^* - \hat{\theta}_{i^t}^{(e_{i^t})},x_{i^t}^t \rangle + \langle \hat{\theta}^{(e_{i^*})}_{i^*},x_{i^*}^t \rangle - \langle \hat{\theta}_{i^t}^{(e_{i^t})},x_{i^t}^t \rangle  \nonumber \\
&\leq \sum_{t = t_{\min} + 1}^T |\langle \theta^*_{i^*} - \hat{\theta}^{(e_{i^*})}_{i^*},x_{i^*}^t \rangle| + |\langle \theta_{i^t}^* - \hat{\theta}_{i^t}^{(e_{i^t})},x_{i^t}^t \rangle| \nonumber \\
&\leq \sum_{i=1}^k \sum_{e_i=1}^{e_{i,\max}} \left ( \sum_{1}^{T_{i,e_i}} \beta \|\theta^*_{i} - \hat{\theta}_{i}^{(e_i)}\|_2 + \sum_{1}^{T_{i,e_i}^*} \beta \|\theta^*_i - \hat{\theta}_i^{(e_i)} \|_2 \right ) ~,
\label{eq:multi_param_regret_eq3}
\end{align}
where the second inequality follows because $\langle \hat{\theta}^{(e_{i^*})}_{i^*},x_{i^*}^t \rangle \leq \langle \hat{\theta}^{(e_{i^t})}_{i^t},x_{i^t}^t \rangle$ as context $i^t$ was chosen ahead of $i^*$ in round $t$ and the third inequality directly follows from the definitions of the various quantities.

The stated result now follows from \myref{eq:multi_param_regret_eq1}, \myref{eq:multi_param_regret_eq2} and \myref{eq:multi_param_regret_eq3}. \qed

\textbf{Proposition \ref{prop:multi_param_prop1}} 
\textit{
    Consider any round $t$ when the episode numbers of the $k$ contexts are $e_1,\hdots,e_k$. Let $i^*$ denote the context with the maximum reward, i.e., $i^* = \underset{1 \leq l \leq k}{\argmax} \langle \mu_l^t + g_l^t,\theta^*_l \rangle$. Let $j$ denote the context having the second largest reward, i.e., $j = \underset{1 \leq l \leq k;l \neq i^*}{\argmax} \langle \mu_l^t + g_l^t,\theta^*_l \rangle$. Define $r = \langle \mu_j^t + g_j^t,\theta^*_j \rangle - \langle \mu_{i^*}^t,\theta^*_{i^*} \rangle$. Then the following condition is satisfied,
	\beq
	\langle g_{i^*}^t,\theta^*_{i^*} \rangle \geq r ~.
	\eeq
}
\proof Since context $i^*$ is optimal in round $t$, we have
\begin{align}
&	\langle \mu_{i^*}^t + g_{i^*}^t, \theta^*_{i^*} \rangle \geq \langle \mu_j^t + g_j^t, \theta^*_j \rangle \nonumber \\
\Rightarrow & \langle g_{i^*}^t,\theta^*_{i^*} \rangle \geq \langle \mu_j^t + g_j^t, \theta^*_j \rangle - \langle \mu_{i^*}^t,\theta^*_{i^*} \rangle ~,
\end{align}
which proves the stated result. \qed

\textbf{Proposition \ref{prop:multi_param_prop2}}
\textit{
	Assume context $j'$ such that $j' = \underset{1 \leq l \leq k,l \neq i^*}{\argmax} \langle \mu_{l}^t + g_{l}^t, \hat{\theta}_{l}^{(e_{l})} \rangle =  \underset{1 \leq l \leq k,l \neq i^*}{\argmax} \langle \mu_{l}^t + g_{l}^t, \theta^*_{l} + \Delta_{l}^{(e_{l})} \rangle$, i.e., the context other than $i^*$ which has the highest estimated reward. Also assume the parameter estimate for context $i^*$ to be $\hat{\theta}_{i^*}^{(e_{i^*})} = \theta^*_{i^*} + \Delta_{i^*}^{(e_{i^*})}$. Then the greedy algorithm selects context $i^*$  if the following condition is satisfied,
	\beq
	\langle g_{i^*}^t,\theta^*_{i^*} \rangle \geq r + \langle \mu_{j'}^t + g_{j'}^t,\Delta_{j'}^{(e_{j'})} \rangle - \langle \mu_{i^*}^t + g_{i^*}^t, \Delta_{i^*}^{(e_{i^*})} \rangle ~.
	\eeq	
}

\proof Now for context $i^*$ to be optimal according to the Greedy algorithm the following condition should be satisfied,
\begin{align}
& \langle \mu_{i^*}^t + g_{i^*}^t, \theta^*_{i^*} + \Delta_{i^*}^{(e_{i^*})} \rangle \geq \langle \mu_{j'}^t + g_{j'}^t, \theta^*_{j'} + \Delta_{j'}^{(e_{j'})} \rangle \nonumber \\
\Rightarrow & \langle g_{i^*}^t,\theta^*_{i^*} \rangle \geq \langle \mu_{j'}^t + g_{j'}^t, \theta^*_{j'} \rangle - \langle \mu_{i^*}^t,\theta^*_{i^*} \rangle + \langle \mu_{j'}^t + g_{j'}^t,\Delta_{j'}^{(e_{j'})} \rangle \nonumber \\
& \quad \quad \quad \quad \quad \quad \quad - \langle \mu_{i^*}^t + g_{i^*}^t, \Delta_{i^*}^{(e_{i^*})} \rangle \nonumber \\
\Rightarrow & \langle g_{i^*}^t,\theta^*_{i^*} \rangle \geq r + \langle \mu_{j'}^t + g_{j'}^t,\Delta_{j'}^{(e_{j'})} \rangle - \langle \mu_{i^*}^t + g_{i^*}^t, \Delta_{i^*}^{(e_{i^*})} \rangle ~,
\end{align}
where in the third line we use the assumption that  $j = \underset{1 \leq l \leq k;l \neq i^*}{\argmax} \langle \mu_l^t + g_l^t,\theta^*_l \rangle$ and hence $\langle \mu_{j'}^t + g_{j'}^t, \theta^*_{j'} \rangle - \langle \mu_{i^*}^t,\theta^*_{i^*} \rangle \leq \langle \mu_{j}^t + g_{j}^t, \theta^*_{j} \rangle - \langle \mu_{i^*}^t,\theta^*_{i^*} \rangle = r$.   \qed

\textbf{Lemma \ref{lemm:multi_param_margin_condition} (Margin Condition)}
\textit{
	Consider good events as when $r \leq c_3 \sigma \sqrt{\log (Tk)}$ and consider errors $\Delta_{i^*}^{(e_{i^*})}$ and $\Delta_{j'}^{(e_{j'})}$ to be small enough such that $\langle \mu_{j'}^t + g_{j'}^t,\Delta_{j'}^{(e_{j'})} \rangle - \langle \mu_{i^*}^t + g_{i^*}^t, \Delta_{i^*}^{(e_{i^*})} \rangle \leq \frac{\sigma^2}{r}$. Then the following holds,
	\beq
	P \left ( \langle g_{i^*}^t,\theta^*_{i^*} \rangle \geq r + \frac{\sigma^2}{r} \suchthat \langle g_{i^*}^t,\theta^*_{i^*} \rangle \geq r \right ) \geq \frac{1}{20} ~,
	\eeq
	for all $r \leq c_3 \sigma \sqrt{\log (Tk)}$.
}

\proof We prove that assuming $\langle \mu_{j'}^t + g_{j'}^t,\Delta_{j'}^{(e_{j'})} \rangle - \langle \mu_{i^*}^t + g_{i^*}^t, \Delta_{i^*}^{(e_{i^*})} \rangle \leq \frac{\sigma^2}{r}$, conditioned on context $i^*$ being optimal in round $t$ implies that it will be played by Greedy with some constant non-zero probability, i.e., we prove the following,

\begin{equation*}
P \left ( \langle g_{i^*}^t,\theta^*_{i^*} \rangle \geq r + \frac{\sigma^2}{r} \suchthat \langle g_{i^*}^t,\theta^*_{i^*} \rangle \geq r \right ) \geq \frac{1}{20} ~,
\end{equation*}

We use the result from Lemma 4.11 in \cite{kmrw18} to lower bound $P \left ( \langle g_{i^*}^t,\theta^*_{i^*} \rangle \geq r + \frac{\sigma^2}{r} \suchthat \langle g_{i^*}^t,\theta^*_{i^*} \rangle \geq r \right )$. We reproduce the proof for the sake of completeness. Denote by $\eta = \langle g_{i^*}^t,\theta^*_{i^*} \rangle$ and $\alpha = \frac{\sigma^2}{r}$. Then,
\begin{align}
P \left ( \langle g_{i^*}^t,\theta^*_{i^*} \rangle \geq r + \frac{\sigma^2}{r} \suchthat \langle g_{i^*}^t,\theta^*_{i^*} \rangle \geq r \right )  &= P[\eta \geq r + \alpha \suchthat \eta \geq r] \nonumber \\
&= \frac{P[\eta \geq r + \alpha]}{P[\eta \geq r]} \nonumber \\
&= \frac{1 - \Phi \left ( \frac{r + \alpha}{\sigma} \right )}{1 - \Phi \left ( \frac{r}{\alpha} \right )}
\end{align}

Using Gaussian tail bounds (Lemma A.6 in \cite{kmrw18}),
\begin{equation*}
\frac{\phi(z)}{2z} \leq 1 - \Phi(z) \leq \frac{\phi(z)}{z} ~.
\end{equation*}
This gives,
\begin{align*}
\frac{1 - \Phi \left ( \frac{r + \alpha}{\sigma} \right )}{1 - \Phi \left ( \frac{r}{\alpha} \right )} &\geq \frac{\phi \left ( \frac{r+\alpha}{\sigma} \right )}{\phi \left ( \frac{r}{\alpha} \right )} \frac{r}{r+\alpha} \frac{1}{2} \\
&\geq \exp \left [ - \frac{(r+\alpha)^2 - r^2}{2\sigma^2} \right ] \frac{r}{2(r+\alpha)} \\
&\geq \exp \left [ - \frac{2r\alpha + \alpha^2}{2 \sigma^2} \right ] \frac{r}{2(r+\alpha)} ~.
\end{align*}
Using $\alpha \leq r$ we get,
\begin{align}
\exp \left [ - \frac{2r\alpha + \alpha^2}{2 \sigma^2} \right ] \frac{r}{2(r+\alpha)} &\geq \frac{1}{4} \exp \left [ - \frac{3r\alpha}{2\sigma^2} \right ] \\
&\geq \frac{1}{4} e^{-\frac{3}{2}} \approx 0.05578 ~,
\end{align}
where in the second inequality we use $\alpha = \frac{\sigma^2}{r}$. Therefore we obtain,
\beq
P \left ( \langle g_{i^*}^t,\theta^*_{i^*} \rangle \geq r + \frac{\sigma^2}{r} \suchthat \langle g_{i^*}^t,\theta^*_{i^*} \rangle \geq r \right ) \approx 0.05578 \geq \frac{1}{20} ~,
\eeq
which proves the third result. 

Finally $P \left ( \langle g_{i^*}^t,\theta^*_{i^*} \rangle \geq r' + \frac{\sigma^2}{r} \suchthat \langle g_{i^*}^t,\theta^*_{i^*} \rangle \geq r' \right ) \approx 0.05578 \geq \frac{1}{20}$ holds for all $r' < r$ due to the following result from \cite{kmrw18}.
\begin{lemm}{\bf (Lemma A.10 in \cite{kmrw18})}
	Let $\eta \sim N(0,\sigma^2)$. Then for any $\alpha > 0$, the conditional ``margin probability",
	\beq
	P[\eta \geq b + \alpha ~ | ~ \eta \geq b] ~,
	\eeq
	is decreasing in $b$.
\end{lemm}				
We have thus proved all the stated results.										\qed

\textbf{Lemma \ref{lemm:multi_param_prop_design_matrices} (Properties of Design Matrices)}
\textit{
Consider any context $i$ and a particular episode $e_i$. The rows of the design matrix $Z^{(e_i)}_{i} \in \R^{T_{i,e_i} \times p}$ are $z^t_i = \mu_i^t + g_i^t$ with $t$ indexing the rounds in episode $e_i$ where context $i$ is chosen by the Greedy algorithm, i.e., $z_i^t = \underset{x_l^t:1 \leq l \leq k}{\argmax}  \langle x_l^t,\hat{\theta}_l^{(e_l)} \rangle$ where $x^t_l = \mu_l^t + g^t_l, ~ g_l^t \sim N(0,\sigma^2 \I_{p \times p})$. Then under the condition $\langle g_i^t,\theta^*_i \rangle \geq r$ for some $r \leq c_3 \sigma \sqrt{\log (Tk)}$,
	\begin{equation*}
	\lambda_{\min} \left (E \left [ z^t_i (z^t_i)^{\intercal}  \suchthat z^t_i = \underset{x_l^t:1 \leq l \leq k}{\argmax} \langle x_l^t, \hat{\theta}_l^{(e_l)} \rangle \right ] \right ) \geq c_2 \frac{\sigma^2}{\log (Tk)} ~.
	\end{equation*}
}

\proof  Using similar argument as used in Lemma 3 we get the following,
\begin{align}
\lambda_{\min} \left ( E \left [z^t_i (z^t_i)^{\intercal} \suchthat z^t_i = \underset{x_l^t:1 \leq l \leq k}{\argmax} \langle x_l^t,\hat{\theta}_l^{(e_l)} \rangle \right ] \right ) &= \min\limits_{w:\|w\|_2 = 1} w^T \left (E \left [z^t_i(z^t_i)^{\intercal} \suchthat z^t_i = \underset{x_l^t:1 \leq l \leq k}{\argmax} \langle x_l^t,\hat{\theta}_l^{(e_l)} \rangle \right ] \right) w \nonumber \\
&= \min\limits_{w:\|w\|_2 = 1}
\left ( E \left [w^{\intercal}z^t_i(z^t_i)^{\intercal} w \suchthat  z^t_i = \underset{x_l^t:1 \leq l \leq k}{\argmax}  \langle x_l^t,\hat{\theta}_l^{(e_l)} \rangle \right] \right)  \nonumber \\
&\geq \min\limits_{w:\|w\|_2 = 1} \left ( \text{Var} \left [\langle w,z^t_i \rangle \suchthat  z^t_i = \underset{x_l^t:1 \leq l \leq k}{\argmax} \langle x_l^t,\hat{\theta}_l^{(e_l)} \rangle \right ] \right ) \nonumber \\
&\geq \min\limits_{w:\|w\|_2 = 1} \left ( \text{Var} \left [\langle w,g^t_i \rangle \suchthat  g^t_i = \underset{g_l^t:1 \leq l \leq k}{\argmax} \langle \mu_l^t + g_l^t,\hat{\theta}_l^{(e_l)} \rangle \right ] \right ) ~.
\label{eq:multi_param_design_matrix_prop_eq1}
\end{align}

Let $j = \underset{1 \leq m \leq k;m \neq i}{\argmax} \langle x_m^t,\hat{\theta}_m^{(e_m)} \rangle$ denote the context which has second maximum reward in round $t$ and let $x_j^t = \mu_j^t + g_j^t, ~ g_j^t \sim N(0,\sigma^2 \I_{p \times p})$. Also let $\hat{\theta}^{(e_i)}_i = \theta^*_i + \Delta^{(e_i)}_i$ and $\hat{\theta}^{(e_j)}_j = \theta^*_j + \Delta^{(e_j)}_j$. Since context $i$ is selected over context $j$ in round $t$, we have the following,
\begin{align}
& \langle x_i^t, \hat{\theta}_i^{(e_i)} \rangle \geq \langle x_j^t,\hat{\theta}_j^{(e_j)} \rangle \nonumber \\
\Rightarrow	& \langle x_i^t, \theta^*_i + \Delta_i^{(e_i)} \rangle \geq \langle x_j^t,\theta^*_j + \Delta_j^{(e_j)} \rangle \nonumber \\
\Rightarrow	& \langle \mu_i^t + g_i^t,\theta^*_i \rangle + \langle x_i^t,\Delta_i^{(e_i)} \rangle \geq \langle x_j^t,\theta^*_j + \Delta_j^{(e_j)} \rangle \nonumber \\
\Rightarrow & \langle g_i^t,\theta^*_i \rangle \geq \langle x_j^t,\theta^*_j + \Delta_j^{(e_j)} \rangle - \langle \mu_i^t,\theta^*_i \rangle - \langle x_i^t,\Delta_i^{(e_i)} \rangle ~.
\end{align}
We now characterize the good events by the condition that $\langle x_j^t,\theta^*_j + \Delta_j^{(e_j)} \rangle - \langle \mu_i^t,\theta^*_i \rangle - \langle x_i^t,\Delta_i^{(e_i)} \rangle \leq c_3 \sigma \sqrt{\log (Tk)}$. Note that there is very less probability on the complementary event $\langle x_j^t,\theta^*_j + \Delta_j^{(e_j)} \rangle - \langle \mu_i^t,\theta^*_i \rangle - \langle x_i^t,\Delta_i^{(e_i)} \rangle \geq c_3 \sigma \sqrt{\log (Tk)}$. Therefore,
\begin{align}
\text{Var} \left [\langle g^t_i,w \rangle \suchthat g^t_i = \underset{g_l^t:1 \leq l \leq k}{\argmax} \langle \mu_l^t + g_l^t, \hat{\theta}_l^{(e_l)} \rangle \right ]
&= \text{Var} \left [\langle g^t_i,w \rangle \suchthat \langle g^t_i,\theta^*_i \rangle \geq \langle x_j^t,\theta^*_j + \Delta_j^{(e_j)} \rangle - \langle \mu_i^t,\theta^*_i \rangle - \langle x_i^t,\Delta_i^{(e_i)} \rangle \right ] \nonumber \\
&\geq \text{Var} \left [\langle g^t_i,w \rangle \suchthat \langle g_i^t,\theta^*_i \rangle \geq c_3 \sigma \sqrt{\log (Tk)} \right ] \nonumber \\
&\geq c_2 \frac{\sigma^2}{\log (Tk)} ~,
\end{align}
where in the second line we condition on the good events when $\langle x_j^t,\theta^*_j + \Delta_j^{(e_j)} \rangle - \langle \mu_i^t,\theta^*_i \rangle - \langle x_i^t,\Delta_i^{(e_i)} \rangle \leq c_3 \sigma \sqrt{\log (Tk)}$ and then use the fact $\text{Var} \left [\langle g^t_i,w \rangle \suchthat \langle g_i^t,\theta^*_i \rangle \geq a \right ] \geq \Omega(1/a^2)$ is a decreasing function of $a$ \cite{kmrw18} so we condition on the maximum value of $a = \langle x_j^t,\theta^*_j + \Delta_j^{(e_j)} \rangle - \langle \mu_i^t,\theta^*_i \rangle - \langle x_i^t,\Delta_i^{(e_i)} \rangle  = c_3 \sigma \sqrt{\log (Tk)}$. Again in the third line we use $\text{Var} \left [\langle g^t_i,w \rangle \suchthat \langle g_i^t,\theta^*_i \rangle \geq a \right ] \geq \Omega(1/a^2)$ \cite{kmrw18}. 			\qed

\begin{theo}
	Consider contexts to be indexed by $i$ and the episode numbers to be indexed by $e_i$. Let $S_{i,e_i}$ denote the set of rounds when context $i$ was selected by the Greedy algorithm in episode $e_i$ with $T_{i,e_i} = | S_{i,e_i} |$. Also assume all rounds satisfy the conditions of Lemma 6. 
	Then when $T_{i,e_i} \geq c_9 ( w(A) + \sqrt{\log \log T} + \sqrt{\log k} +  \sqrt{\log (1/\delta)})^2 \log^2 (Tk)$, with probability atleast $1 - \delta \exp(-\eta_2 w^2(A)) - \delta$ the following RE condition holds for all contexts $1 \leq i \leq k$,
	\beq
	\inf\limits_{1 \leq i \leq k} \inf\limits_{e_i \leq e_{i,\max}} \inf\limits_{u \in A} \frac{1}{T_{i,e_i}} \|Z^{(e_i)}_i u\|_2^2 \geq c_{4} \frac{\sigma^2}{\log (Tk)} ~.
	\eeq
	Also consider parameter estimation using the constrained least squares estimator. Define the following quantities $r \leq c_3 \sigma \sqrt{\log (Tk)}$, $\gamma = \frac{c_{12} \kappa_{\omega} (w(A) + \sqrt{\log \log T} + \sqrt{\log k} + \sqrt{\log(1/\delta)}) \sqrt{\log(Tk)}}{\sigma}$ and $\beta = \underset{\substack{{1 \leq i \leq k, 1 \leq t \leq T} \\  {v \in A}}}{\max} \langle x_i^t,v \rangle$. Then if $T_{i,e_i} \geq \frac{4 \gamma^2 r^2 \beta^2}{\sigma^4}$, then with probability atleast $1 - \delta \exp(-\eta_1 w^2(A)) - 2\delta$,
	\beq
	\sup\limits_{1 \leq i \leq k} \sup\limits_{e_i \leq e_{i,\max}} \|\hat{\theta}_i^{(e_i + 1)} - \theta^*_i\|_2 \leq \frac{\sigma^2}{2 \beta r} ~.
	\eeq
\end{theo}

\proof The following result can be proved with probability atleast $1 - \delta \exp(-\eta_1 w^2(A)) - 2\delta$, using same arguments as Theorem D.4.
\beq
\|\hat{\theta}^{(e_i+1)}_i - \theta^*_i\|_2 = \|\Delta^{(e_i)}_i\|_2 \leq \frac{\gamma}{\sigma \sqrt{T_{i,e_i}}} ~,
\label{eq:multi_param_est_err_bnds_eq11}
\eeq
where 
$\gamma = \frac{c_{12} \kappa_{\omega} (w(A) + \sqrt{\log k} + \sqrt{\log (1/\delta)}) \sqrt{\log (Tk)}}{\sigma}$ and $\beta = \underset{\substack{{1 \leq i \leq k, 1 \leq t \leq T} \\  {v \in A}}}{\max} \langle x_i^t,v \rangle$, it follows from \myref{eq:multi_param_est_err_bnds_eq11} that when $T_{i,e_i} \geq \frac{4\gamma^2 r^2 \beta^2}{\sigma^4}$ then,
\beq 
\|\hat{\theta}^{(e_i+1)}_i - \theta^*\|_2 = \|\Delta^{(e_i)}_i\|_2 \leq \frac{\sigma^2}{2\beta r} ~,
\eeq
which is the desired result. 			\qed

\textbf{Theorem 3}
\textit{
	Consider computation of regret for the Greedy algorithm in the multi parameter setting following Lemma 4. Define the following quantities $r \leq c_3 \sigma \sqrt{\log (Tk)}$, $\gamma = \frac{c_{12} \kappa_{\omega} (w(A) + \sqrt{\log T} + \sqrt{\log k}) \sqrt{\log (Tk)}}{\sigma}$ and $\beta = \underset{\substack{{1 \leq i \leq k, 1 \leq t \leq T} \\  {v \in A}}}{\max} \langle x_i^t,v \rangle$. The margin condition in Lemma 5 is satisfied with probability atleast $1 - \delta\exp(-\eta_1 w^2(A)) - 4\delta$ when,
	\beq
	t_{\min} \geq \frac{4k\gamma^2 r^2 \beta^2}{\sigma^4} + \sqrt{\frac{1}{2} \log(1/\delta)} ~.
	\eeq
	Under the margin condition, the regret is maximized when in each round each context has equal probability to be selected by the Greedy algorithm. The equal probability implies that in expectation $T_1 = T_2 = \hdots = T_k = \frac{T}{k}$. Also the regret is upper bounded as follows,
	\beq
	\text{Reg}(T) \leq 2 \beta t_{\min} + 82 \beta \gamma \sqrt{Tk} \log (T) ~.
	\eeq
	Moreover $\beta = \underset{\substack{{1 \leq i \leq k, 1 \leq t \leq T} \\  {v \in A}}}{\max} \langle x_i^t,v \rangle \leq (1+ c_1 \sigma (w(A) + \sqrt{\log (1/\delta)}))$ with probability atleast $1 - \delta$. Therefore with probability atleast $1 - \delta\exp(-\eta_1 w^2(A)) - 5\delta$, 
	\beq
	\text{Reg}(T) \leq O\left (  \gamma \cdot \beta \cdot \log(T) \cdot \sqrt{Tk} \right )
	\eeq	
}

We first derive bounds on the parameter $t_{\min}$ in Lemma 4. The multi-parameter setting requires a warm start of $T_0$ rounds, where $T_0$ is computed as,
\beq
T_0 = \frac{4k\gamma^2 r^2 \beta^2}{\sigma^4} ~.
\label{eq:multi_param_regret_bnds_eq1}
\eeq 
This is required for the margin condition of Lemma 5 to be satisfied with high probability. To see that, when $T_0 = \frac{k\gamma^2 r^2 \beta^2}{\sigma^4}$, $T_{i,e_i} \geq \frac{4\gamma^2 r^2 \beta^2}{\sigma^4}$ for all contexts $1 \leq i \leq k$ and all episodes $e_i, 1 \leq i \leq k$. Then for any context combination $i,j$ for $r = c_3 \sigma \sqrt{\log (Tk)}$, we have the following,
\begin{align}
\langle \mu_i^t + g_i^t, \Delta_{i}^{(e_i)} \rangle - \langle \mu_j^t + g_j^t,\Delta_j^{(e_j)} \rangle &= \langle x_i^t,\Delta_i^{(e_i)} \rangle - \langle x_j^t,\Delta_j^{(e_j)} \rangle \nonumber \\
&\leq \abs{\langle x_i^t,\Delta_i^{(e_i)} \rangle} + \abs{\langle x_j^t,\Delta_j^{(e_j)} \rangle} \nonumber \\
&\leq \beta \|\Delta_i^{
(e_i)}\|_2 + \beta \|\Delta_{j}^{(e_j)}\|_2 \nonumber \\
&\leq \frac{\sigma^2}{r} ~,
\label{eq:multi_param_regret_bnds_eq3}
\end{align}
where in the last line we use that when $T_{i,e_i}$ with high probability $\|\Delta_i^{e_i}\|_2,\|\Delta_j^{e_j}\|_2 \leq \frac{\sigma^2}{2\beta r}$. Let $i = \underset{1 \leq l \leq k}{\argmax} \langle x_l^t,\theta^*_l \rangle = \langle \mu_l^t + g_l^t,\theta^*_l \rangle$ be the optimal context in round $t$. In the margin condition, we also assume that $\langle g_{i}^t, \theta^*_{i} \rangle \leq c_3 \sigma \sqrt{\log (Tk)}$. We show that over $T$ rounds the assumption is not satisfied only for a constant number of rounds. First note that for any context $l$, $\langle g_l^t,\theta^*_l \rangle$ is a $N(0,\sigma)$ Gaussian random variable. Therefore using Gaussian random variable tail bounds, we get,
\beq
P \left ( \abs{\langle g_l^t,\theta^*_l \rangle} \geq c_3 \sigma \sqrt{\log (Tk)}  \right ) \leq \exp \left ( - c_4 \log (Tk) \right ) ~.
\eeq
Now there are a total of $Tk$ realizations of $\langle g_l^t,\theta^*_l \rangle$ with $1 \leq l \leq k, 1 \leq t \leq T$. Consider the binomial random variable $\nu \sim \text{Binomial}(Tk,\exp \left ( - c_4 \log (Tk) \right ))$. Now $E[\nu] = Tk ~ \exp \left ( - c_4 \log (Tk) \right ) = \exp \left ( - c_4 \log (Tk) + \log (Tk) \right ) \leq 1$ where we assume that constants $c_3,c_4$ are chosen such that the expectation is less than $1$. Therefore by a tail bound for binomials,
\beq
P \left ( \nu \geq 1 + \sqrt{\frac{1}{2} \log(1/\delta)} \right ) \leq \delta   ~.
\label{eq:multi_param_regret_bnds_eq2}
\eeq
Therefore combining \myref{eq:multi_param_regret_bnds_eq1} and \myref{eq:multi_param_regret_bnds_eq2} the margin condition is satisfied with probability atleast $1 - \delta\exp(-\eta_1 w^2(A)) - 3\delta$ when,
\begin{align}
t_{\min} &\geq T_0 + 1 + \sqrt{\frac{1}{2} \log(1/\delta)} \nonumber \\
&\geq \frac{4k\gamma^2 r^2 \beta^2}{\sigma^4} + 1 + \sqrt{\frac{1}{2} \log(1/\delta)} ~.
\end{align}

Now to compute the regret, let $i = \underset{1 \leq l \leq k}{\argmax} \langle x_l^t,\theta^*_l \rangle = \langle \mu_l^t + g_l^t,\theta^*_l \rangle$ be the actual optimal context in round $t$ and $j' = \underset{1 \leq l \leq k;l \neq i}{\argmax} \langle x_l^t,\theta^*_l + \Delta_l^{e_l} \rangle = \langle \mu_l^t + g_l^t,\theta^*_l + \Delta_l^{e_l} \rangle$ be the maximum estimated context rewards other than context $i$. Now according to \myref{eq:multi_param_regret_bnds_eq2}, according to the margin condition except for $1 + \sqrt{\frac{1}{2} \log(1/\delta)}$ rounds with high probability we have $\langle g_i^t,\theta^*_i \rangle \geq r$ for some $r \leq c_3 \sigma \sqrt{\log (Tk)}$. Now for context $i$ to be be selected over context $j'$ we have the following condition,
\begin{align}
\langle g_i^t,\theta^*_i \rangle &\geq r + \langle \mu_{j'}^t + g_{j'}^t,\Delta_{j'}^{e_{j'}} - \langle \mu_i^t + g_i^t,\Delta_i^{e_i^t} \rangle \nonumber \\
&\geq r + \frac{\sigma^2}{r} ~,	
\end{align}
where the second inequality is from equation \myref{eq:multi_param_regret_bnds_eq3}. Now from Lemma 5 we have established the following condition,
\beq
P \left ( \langle g_i^t,\theta^*_i \rangle \geq r + \frac{\sigma^2}{r} \suchthat \langle g_i^t,\theta^* \rangle \geq r  \right ) \geq \frac{1}{20} ~,
\eeq
that is, context $i$ is the estimated optimal context in $1$ out of $20$ times when context $i$ is actually the optimal context. Now let $T_{i,e_i}^*$ be the number of times context $i$ is actually optimal in episode $e_i$. Then the number of times context $i$ is estimated to be optimal is a binomial random variable: $\text{Binomial}(T_{i,e_i}^*, 1/20)$. Therefore applying Chernoff bounds for the binomial random variable $\left ( T_{i,e_i}^*, \frac{1}{20} \right )$,
\begin{align}
&	P \left [ T_{i,e_i} \leq \frac{T_{i,e_i}^*}{20} - \frac{T_{i,e_i}^*}{40} \right ] \leq \exp \left ( - \frac{T^*_{i,e_i}}{160} \right ) \nonumber \\
\Rightarrow	& P \left [ T_{i,e_i}^* \geq 40 T_{i,e_i} \right ] \leq \exp \left ( - \frac{T_{i,e_i}}{4} \right )
\end{align} 

This is for any context $i$ and episode $e_i$. Now taking a union bound over all contexts $1 \leq i \leq k$ and episodes $1 \leq e_i \leq \log T$ and using $T_{i,e_i} \geq c_9( w(A) + \sqrt{\log \log T} + \sqrt{\log k} + \sqrt{\log (1/\delta)})^2\log^2 (Tk)$ we get,
\beq
P[T_{i,e_i}^* \geq 40 T_{i,e_i}] \leq \exp \left ( -\frac{c_9 ^2 (w^2(A) + \log(1/\delta)) \log^2 (Tk)}{4}
\right ) \leq \delta   ~.
\label{eq:multi_param_regret_bnds_eq4}
\eeq 

With this result the regret can be upper bounded as follows with probability atleast $1 - \delta \exp(-\eta_1 w^2(A)) - 4\delta$
\begin{align}
\text{Regret}(x^t,i^1,\hdots,x^T,i^T) &\leq 2 \beta  t_{\min} + \sum_{i=1}^k \sum_{e_i=1}^{e_{i,\max}} \left ( \sum_{t=1}^{T_{i,e_i}} \beta \|\theta^*_{i} - \hat{\theta}_{i}^{e_i}\|_2 + \sum_{t=1}^{T_{i,e_i}^*} \beta \|\theta^*_i - \hat{\theta}_i^{e_i}\|_2 \right ) \nonumber \\
&\leq 2 \beta t_{\min} + \sum_{i=1}^k \sum_{e_i=1}^{e_{i,\max}} \left ( \sum_{t=1}^{T_{i,e_i}} 41 \beta \|\theta^*_{i} - \hat{\theta}_{i}^{e_i}\|_2  \right ) \nonumber \\
&\leq 2 \beta  t_{\min} + \sum_{i=1}^k \sum_{e_i=1}^{e_{i,\max}} \left ( \sum_{t=1}^{T_{i,e_i}} 41 \beta \frac{\gamma}{\sqrt{T_{i,e_i-1}}} \ \right ) \nonumber \\
&\leq 2 \beta  t_{\min} + \sum_{i = 1}^k \sum_{e_i=1}^{e_{i,\max}} 82 \beta \gamma \sqrt{T_{i,e_i}} \nonumber \\
&\leq 2 \beta  t_{\min} + \sum_{i=1}^k 82 \beta \gamma \sqrt{T_i} \log T_i \nonumber \\
&\leq 2 \beta t_{\min} +  82  \beta \gamma \sqrt{Tk} \log T ~,
\end{align}
where in the second inequality we have used the result $\myref{eq:multi_param_regret_bnds_eq4}$, in the fourth inequality we have used $T_{i,e_i} = 2T_{i,e_i-1}$, in the fifth inequality we have used $e_{i,\max} \leq \log T_i$ and in the last inequality we have used $T_i = T/k$ gives the maximum regret and $\log T_i \leq \log T$.

Substituting the value of $\gamma$ assumed earlier and noting \\
 $\beta = \underset{\substack{{1 \leq i \leq k, 1 \leq t \leq T} \\  {v \in A}}}{\max} \langle x_i^t,v \rangle \leq \underset{\substack{{1 \leq i \leq k, 1 \leq t \leq T} \\  {v \in A}}}{\max} \left ( \langle \mu_i^t,v \rangle + \langle g_i^t,v \rangle \right ) \leq (1+ c_1 \sigma (w(A) + \sqrt{\log (1/\delta)}))$ with probability atleast $1 - \delta$ following from Lemma \ref{lemm:general_max_gauss_bnds} proves the stated result. \qed

\end{document}